\documentclass[11pt]{article}

\usepackage[final]{acl}

\usepackage{times}
\usepackage{latexsym}

\usepackage[T1]{fontenc}
\usepackage[utf8]{inputenc}

\usepackage{microtype}

\usepackage{inconsolata}

\usepackage{graphicx}
\usepackage{booktabs}       % professional-quality tables
\usepackage{subcaption}
\usepackage{makecell}  % For the \makecell command

\usepackage[dvipsnames]{xcolor}
\usepackage{amssymb}
\usepackage{xcolor}
\usepackage{tikz}
\usepackage{lipsum}
\usepackage{enumitem}
\usepackage{graphicx}
\usepackage{subcaption}
\usepackage{alphalph}

\usepackage{booktabs}
\usepackage{multirow}
\usepackage{longtable}
\usepackage{array}
\usepackage{makecell}

\definecolor{extraction}{HTML}{E41A1C}
\definecolor{symbolization}{HTML}{377EB8}
\definecolor{structural}{HTML}{4DAF4A}
\definecolor{retrieval}{HTML}{984EA3}
\definecolor{deduction}{HTML}{FF7F00}
\definecolor{algebraic}{HTML}{A65628}
\definecolor{arithmetic}{HTML}{F781BF}
\definecolor{finalanswer}{HTML}{333333}

\definecolor{residual}{HTML}{E41A1C}
\definecolor{attn}{HTML}{377EB8}
\definecolor{mlp}{HTML}{4DAF4A}

\definecolor{qwenThree}{HTML}{1B7837}
\definecolor{qwenTwoFive}{HTML}{762A83}
\definecolor{gemmaFour}{HTML}{D6604D}

\usepackage[most]{tcolorbox}
\usepackage{xcolor}

\definecolor{poscolor}{HTML}{4C72B0}
\definecolor{tokcolor}{HTML}{C44E52}

\usepackage{tcolorbox}
\usepackage{fvextra}

\DefineVerbatimEnvironment{promptlisting}{Verbatim}{
  breaklines=true,
  breakanywhere=true,
  fontsize=\small
}

\newtcolorbox{promptbox}{
  colback=black!3,
  colframe=black!40,
  boxrule=0.5pt,
  arc=2mm,
  left=2mm,
  right=2mm,
  top=1mm,
  bottom=1mm
}
\newcolumntype{L}[1]{>{\raggedright\arraybackslash}p{#1}}

\title{Beneath the Surface of Chains-of-Thought: \\
A Mechanistic Interpretation of Reasoning Operations in LLMs}

\author{
    Seogyeong Jeong\textsuperscript{1}\thanks{Work done during an internship at NAVER AI Lab.}
    \quad
    Jaehui Hwang\textsuperscript{2}
    \quad 
    Dongyoon Han\textsuperscript{2}
    \quad
    Geonmo Gu\textsuperscript{2}
\\
\textbf{Alice Oh}\textsuperscript{1}\thanks{Corresponding authors.}
\quad
\textbf{Taekyung Kim}\textsuperscript{2}\footnotemark[2]
\\
\textsuperscript{1}KAIST
\qquad
\textsuperscript{2}NAVER AI Lab
\\
{\small
\href{mailto:sg.jeong28@kaist.ac.kr}{\color{black}{\texttt{sg.jeong28@kaist.ac.kr}}}
\quad
\href{mailto:alice.oh@kaist.edu}{\color{black}{\texttt{alice.oh@kaist.edu}}}
\quad
\href{mailto:taekyung.k@navercorp.com}{\color{black}{\texttt{taekyung.k@navercorp.com}}}
}
}

\begin{document}
\maketitle

\begin{abstract}
Reasoning in large language models unfolds through diverse functional operations, such as problem formulation, goal decomposition, and deduction. 
Although these operations are explicitly distinguished in text, little is known about how they are geometrically organized in representation spaces.
To this end, we investigate whether distinct reasoning operations exhibit corresponding geometric structure in hidden representations. 
We find that operations are separable in held-out representations, with separability peaking in middle layers, and verify that this structure is not explained by lexical or positional confounds.
Across layers, token-wise operation-alignment becomes more distributed over spans, while identical surface tokens are represented differently depending on the operation of its surrounding chunk.
Attention-masking interventions further show that operation-aligned representations at chunk onset depend on preceding reasoning context.
Consequently, our work demonstrates that language models maintain representational correspondence between linguistic reasoning expressions and their internal geometric structures.
Code and project materials are available at \url{https://github.com/naver-ai/beneath-cot}.
\end{abstract}

\begin{figure*}[t!]
    \centering
    \includegraphics[width=\textwidth]{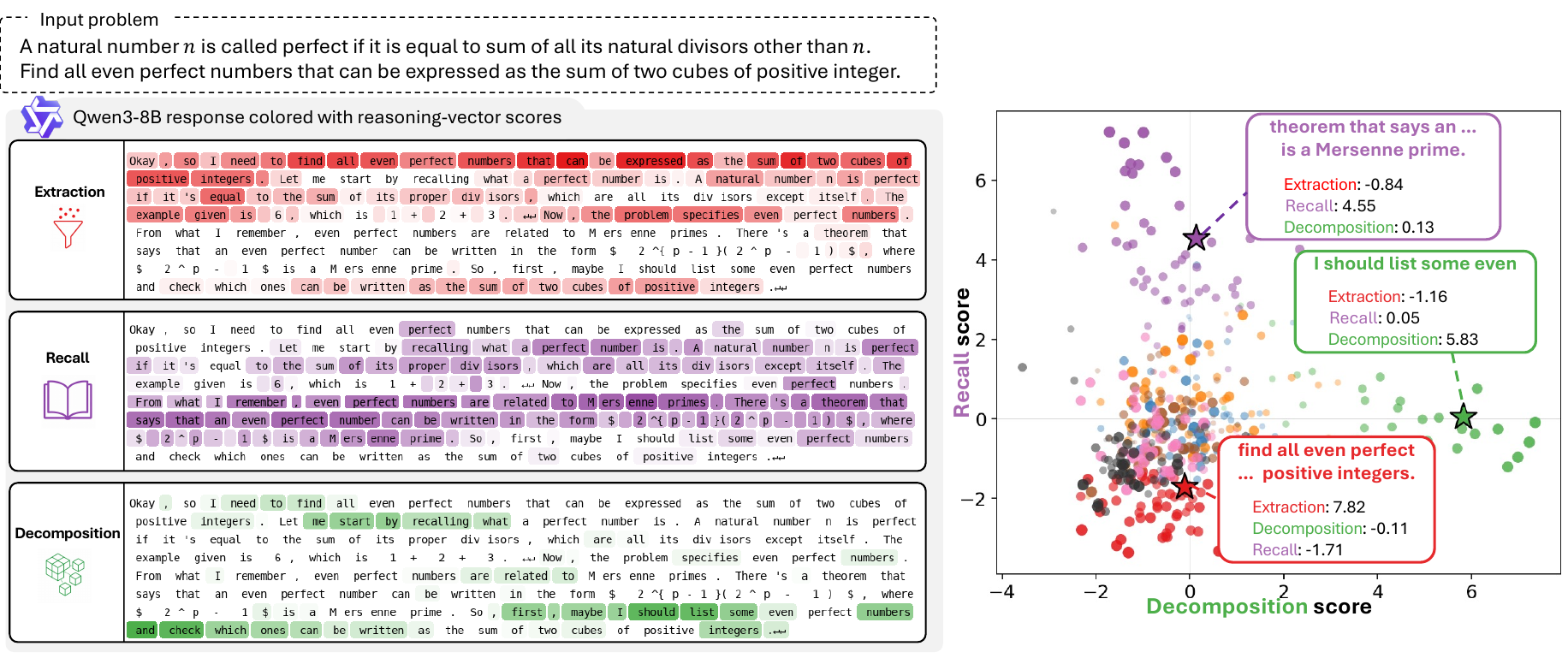}
    \vspace{-2em}
    \caption{\textbf{Overview of reasoning-vector analysis.}
Given an input problem and the model-generated reasoning trace, we probe hidden representations with reasoning operation vectors corresponding to different reasoning operations. 
The left panel shows token-level reasoning operation-alignment scores over the generated text, while the right panel projects span representations onto \textit{Decomposition} and \textit{Recall} reasoning-vector axes. 
Regions with strong reasoning operation-specific signals appear along the corresponding directions.
}
    \label{fig:wide_placeholder}
    \vspace{-1.0em}
\end{figure*}

\section{Introduction}

Recent reasoning-oriented LLMs~\cite{xu2025largereasoningmodelssurvey} have achieved strong performance on multi-step problem-solving tasks. 
Crucially, their training increasingly optimizes not only final-answer correctness but also the reasoning trajectories that produce those answers~\cite{Guo_2025, zhang-etal-2025-lessons}, through methods such as reinforcement learning and search algorithm~\cite{li202512surveyreasoning}. 
As reasoning trajectories become objects of optimization, a central question is what structure these training objectives
are shaping inside the model.

A deeper understanding of LLM reasoning therefore requires examining not only the generated reasoning traces, but also how the reasoning operations expressed in those traces are internally represented.
In this work, we ask whether the hidden representations produced during chain-of-thought (CoT) reasoning~\cite{wei2022chain} encode the reasoning operation being performed, beyond the lexical identity of the current token. 
We further ask whether instances of the same operation exhibit shared representational structure across different problems and reasoning contexts.

Prior work suggests that reasoning involves continuous latent dynamics~\cite{hao2025training,shen-etal-2025-codi, xu-etal-2025-softcot,sun-etal-2026-llm} and that hidden representations encode signals associated with answer correctness~\cite{zhang2025reasoning}. 
However, it remains unclear how the distinct reasoning operations expressed within a CoT trace are organized in representation space, or whether their organization generalizes across tokens and problems.
We therefore study the representational geometry of textually explicit reasoning operations in reasoning LLMs. Specifically, we ask: (1) Do instances of the same reasoning operation share geometric structure beyond their lexical and problem-specific content? (2) Where and when does this operation-level structure emerge across model layers and reasoning trajectories? (3) How is this structure affected by contextual factors such as token identity, operation position, and execution correctness?

To answer these questions, we categorize reasoning operations using Polya’s problem-solving framework from \textit{How to Solve It}~\cite{polya1945solve}, which provides a coarse-grained taxonomy of stages.
% such as understanding the problem, devising a plan, carrying out the plan, and looking back.
We apply this taxonomy to generated reasoning traces and analyze the corresponding hidden representations on mathematical reasoning datasets, including DAPO-MATH-17K~\citep{yu2025dapo} and TheoremQA~\citep{chen-etal-2023-theoremqa}.
We further examine multiple reasoning-oriented LLMs from the Qwen and Gemma families to assess whether operation-level structures are consistently observed across model families.

Our key findings are as follows: (1) reasoning operations are separable in held-out hidden representations, with separability peaking in middle layers (\S\ref{sec:separability}); (2) operation signals become distributed across spans and contextualize even identical surface tokens (\S\ref{sec:sep_evolve}); (3) attention masking shows that preceding reasoning context causally contributes to subsequent operation representations (\S\ref{sec:sep_trigger}); and (4) operation geometry persists but weakens under factual errors (\S\ref{sec:erroneous_execution}).

Our findings suggest that language models maintain a 
representational correspondence between explicitly expressed reasoning 
operations in text and their internal geometric organization.
By bringing the reasoning trace from the output text level down to the internal representation level, our work offers a new lens through which to examine LLM cognition. 
As emerging paradigms increasingly optimize the reasoning process itself, elucidating these internal mechanisms lays the vital groundwork for improving reasoning capabilities through direct latent space interventions.

\begin{table*}[t]
    \centering
    \setlength{\tabcolsep}{7pt}
    \hspace{-2mm}
    \scalebox{0.8}{
    \begin{tabular}{l|ll}
        \toprule
        Problem-solving stage & Reasoning Operation & Example \\
        \midrule
        \makecell[l]{Understanding the problem} & 
        \makecell[l]{\textcolor{extraction}{Extraction} \\ \textcolor{symbolization}{Direct mapping}} & 
        \makecell[l]{The problem says that \textcolor{extraction}{\textbf{two numbers add up to 10.}} \\ 
        x is greater than 3 $\rightarrow$ \textcolor{symbolization}{\textbf{$x > 3$}}} \\

        \midrule
        \makecell[l]{Planning the solution} & 
        \makecell[l]{\textcolor{structural}{Decomposition}} & 
        \makecell[l]{Get the area of Heptagon $\rightarrow$ \textcolor{structural}{\textbf{Divide it into triangles}} \\ 
        $\rightarrow$ \textcolor{structural}{\textbf{Solve each triangle}} $\rightarrow$ Combine results} \\

        \midrule
        \makecell[l]{Carrying out the plan} & 
        \makecell[l]{\textcolor{retrieval}{Recall} \\  
        \textcolor{deduction}{Deduction} \\ 
        \textcolor{algebraic}{Algebraic manipulation} \\ 
        \textcolor{arithmetic}{Arithmetic computation}} & 
        \makecell[l]{Area of a circle is needed $\rightarrow$ \textcolor{retrieval}{\textbf{Recall $A = \pi r^2$}} $\rightarrow$ apply to problem \\ 
        $x$ is greater than 3 and $x$ is an integer $\rightarrow$ \textcolor{deduction}{\textbf{$x \geq 4$}} \\ 
        $x + y = 10$ $\rightarrow$ \textcolor{algebraic}{\textbf{$y = 10 - x$}} \\ 
        $10 - 3$ \textcolor{arithmetic}{\textbf{$= 7$}}} \\

        \midrule
        \makecell[l]{Looking back and final answer} & 
        \textcolor{gray}{Final answer} & 
        Therefore, the values are \textcolor{gray}{\textbf{$x = 3$ and $y = 7$}}. \\

        \bottomrule
    \end{tabular}
    }
    \vspace{-.5em}
    \caption{\textbf{Overview of main reasoning operations for our analyses.} These eight frequent operation types are used as span labels in our representation analysis. }
    \label{tab:reasoning_ops_examples}
    \vspace{-1.0em}
\end{table*}

\section{Preliminary}
\paragraph{Reasoning LLMs.}
Given an input prompt $x$, a reasoning LLM generates an intermediate reasoning trace $r = (t_1, \ldots, t_N)$ before producing the final answer $y$:
\begin{equation}
    p(r, y \mid x) = \prod_{i=1}^{N} p(t_i \mid x, t_{<i}) \cdot p(y \mid x, r).
\end{equation}
For each token position $i$, we extract hidden representations $\mathbf{h}^{(\ell)}_i \in \mathbb{R}^d$ from layer $\ell$ of the model. 
Our study investigates whether and how the geometric organization of these representations reflects the functional roles of different reasoning steps expressed in the text.
% Our study investigates whether and how the geometric organization of these representations reflects the functional structure of the reasoning process expressed in text.

\paragraph{Reasoning Operation taxonomy.}
We view the reasoning trace $r$ not as a homogeneous token sequence but as a sequence of \textit{reasoning operation chunks}, $r = (c_1, \ldots, c_K)$, where each chunk $c_k$ corresponds to a contiguous span of tokens that serves a specific functional role in the problem-solving process. 
To systematically categorize these operations, we define a hierarchical taxonomy of reasoning operations for generated reasoning traces based on the four-stage structure~\cite{polya1945solve}.
The taxonomy is intended to capture the functional role of each reasoning step in the solution process, rather than its surface wording alone.
% For example, a step may extract given information from the problem, translate a natural-language statement into a symbolic form, decompose the problem into subproblems, recall a relevant formula or theorem, derive a conclusion, manipulate an equation, perform arithmetic computation, or present the final answer.

The full taxonomy, which includes a broader set of operation types and subtypes, is described in Appendix~\ref{app:full_taxonomy}.
In the main analysis, we focus on eight recurring operation types that appear frequently across the generated traces that cover different parts of four Polya’s problem-solving process, where the examples are shown in Table~\ref{tab:reasoning_ops_examples}. 
% The remaining types are used to organize the taxonomy and are described in Appendix~\ref{app:full_taxonomy}.
% The full taxonomy includes a broader set of operation types and subtypes. In the main analysis, we focus on eight recurring operation types that appear frequently across the generated traces that cover different parts of four Polya’s problem-solving process.
In the \textbf{understanding stage}, \textcolor{extraction}{Extraction} identifies information explicitly given in the problem, while \textcolor{symbolization}{Direct mapping} converts natural-language statements into formal expressions.
In the \textbf{planning stage}, \textcolor{structural}{Decomposition}  captures the act of breaking a complex problem into smaller subproblems.
In the \textbf{execution stage}, \textcolor{retrieval}{Recall} retrieves relevant formulas, definitions, or rules; \textcolor{deduction}{Deduction} derives conclusions from known premises; \textcolor{algebraic}{Algebraic manipulation} transforms symbolic expressions; and \textcolor{arithmetic}{Arithmetic computation} performs numerical calculations.
Finally, \textcolor{Gray}{Final answer} closes the reasoning process by presenting the derived result in the required format.
% The examples of these main eight reasoning taxonomy is shown in table~\ref{tab:reasoning_ops_examples}. 
% The remaining types are used to organize the taxonomy and are described in Appendix~\ref{app:full_taxonomy}.
We treat this taxonomy as an operational framework for analyzing textually expressed reasoning functions, rather than as a definitive cognitive model or an exhaustive account of the LLM’s latent computation.

% We emphasize that this taxonomy is an operational analysis framework, not a definitive cognitive model of LLM reasoning.
% Labels are assigned based on the textually explicit function of each step.
% Therefore, the taxonomy allows us to examine whether different explicit reasoning operations are associated with distinguishable hidden-state structures, without assuming that the labels exhaustively describe the model’s full latent computation.

% In the \textbf{understanding stage}, \textcolor{Red}{Extraction} identifies information explicitly given in the problem, while \textcolor{Blue}{Symbolization.direct-mapping} converts natural-language statements into formal expressions.
% In the \textbf{planning stage}, \textcolor{ForestGreen}{Structural-analysis.decomposition}  captures the act of breaking a complex problem into smaller subproblems.
% In the \textbf{execution stage}, \textcolor{Purple}{Retrieval.recall} retrieves relevant formulas, definitions, or rules; \textcolor{Orange}{Inference-flow.chaining.deduction} derives conclusions from known premises; \textcolor{Brown}{Execution.operation.algebraic-manipulation} transforms symbolic expressions; and \textcolor{Thistle}{Execution.operation.arithmetic-computation} performs numerical calculations.
% Finally, \textcolor{Gray}{Final-answer} closes the reasoning process by presenting the derived result in the required format.

% \textbf{structural-analysis.decomposition}
\section{Geometric Structure of Reasoning Operations in Reasoning LLMs}
We investigate how reasoning operations are organized in the hidden representation space of reasoning LLMs.
We first test whether different operations are separable in held-out representations(\S\ref{sec:main_separability}) and whether this structure can be explained by lexical(\S\ref{sec:lexical_confounds}) or positional confounds(\S\ref{sec:additional_robustness}).
We then characterize how operation-aligned signals evolve across layers, examining their distribution within spans(\S\ref{sec:variance}) and the representations of identical surface tokens(\S\ref{sec:scatter_surface}).
Finally, we use attention-masking interventions to test the contribution of preceding context (\S\ref{sec:sep_trigger}) and examine how operation geometry changes under factual errors (\S\ref{sec:erroneous_execution}).

\begin{figure*}[t]
    \centering

    % =========================================================
    % Shared legend
    % =========================================================
    
    % \vspace{-0.5em}

    % \vspace{0.5em}

    % =========================================================
    % Panel A: AUROC
    % =========================================================
    \begin{subfigure}[t]{0.61\textwidth}
    \centering

    \begin{subfigure}[b]{\textwidth}
        \centering
        \footnotesize
        \renewcommand{\arraystretch}{0.7}
        \begin{tabular}{@{}ll@{\hspace{0.9em}}ll@{}}
            \tikz{\draw[line width=0.8mm, color=extraction] (0,0) -- (0.35,0);} &
            \texttt{Extraction} &
            \tikz{\draw[line width=0.8mm, color=symbolization] (0,0) -- (0.35,0);} &
            \texttt{Direct mapping} \\

            \tikz{\draw[line width=0.8mm, color=structural] (0,0) -- (0.35,0);} &
            \texttt{Decomposition} &
            \tikz{\draw[line width=0.8mm, color=retrieval] (0,0) -- (0.35,0);} &
            \texttt{Recall} \\

            \tikz{\draw[line width=0.8mm, color=deduction] (0,0) -- (0.35,0);} &
            \texttt{Deduction} &
            \tikz{\draw[line width=0.8mm, color=algebraic] (0,0) -- (0.35,0);} &
            \texttt{Algebraic manipulation} \\

            \tikz{\draw[line width=0.8mm, color=arithmetic] (0,0) -- (0.35,0);} &
            \texttt{Arithmetic computation} &
            \tikz{\draw[line width=0.8mm, color=finalanswer] (0,0) -- (0.35,0);} &
            \texttt{Final answer}
        \end{tabular}
    \end{subfigure}

    \begin{minipage}[c]{0.055\linewidth}
        \centering
        \rotatebox{90}{\sffamily AUROC}
    \end{minipage}%
    \begin{minipage}[c]{0.935\linewidth}
        \centering
        \includegraphics[
            width=0.8\linewidth
        ]{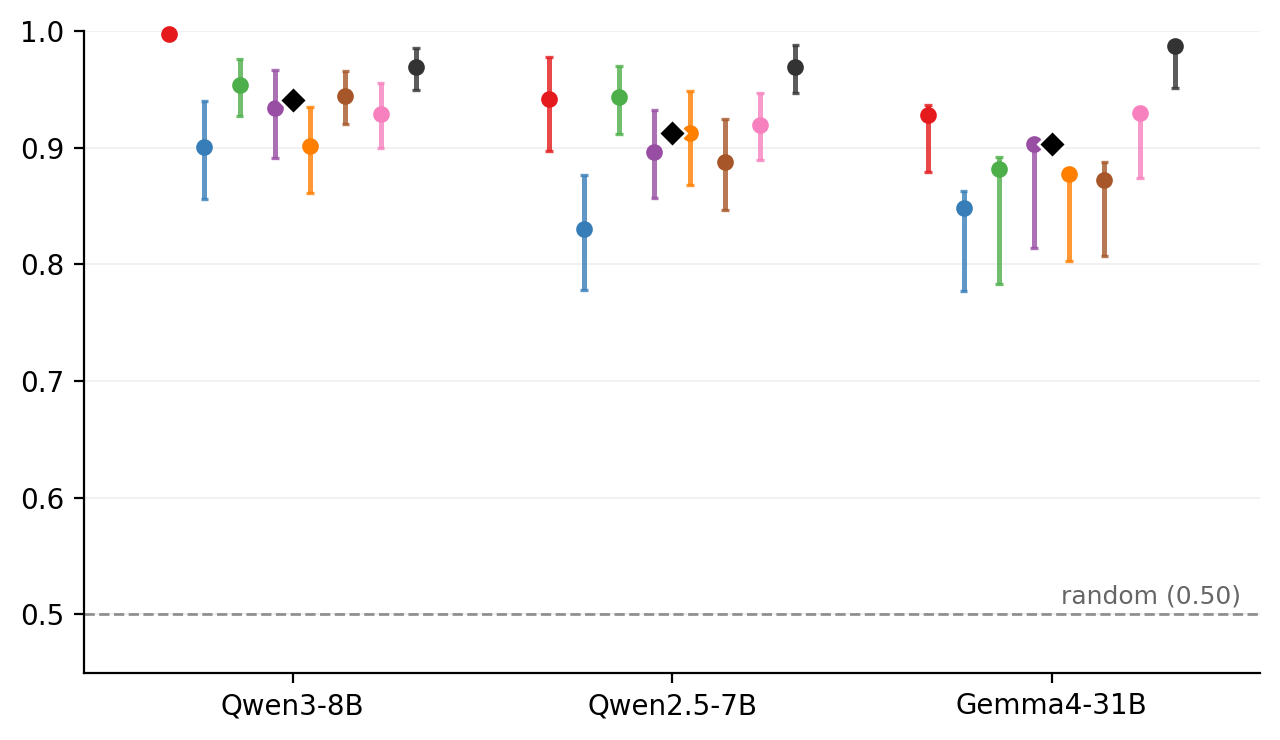}
    \end{minipage}

    \subcaption{Peak-layer AUROC (95\% CI)}
    \label{fig:auroc-panel}
\end{subfigure}
    \hfill
    % =========================================================
    % Panel B: AUPRC
    % =========================================================
    \begin{subfigure}[t]{0.37\textwidth}
    \centering

    \begin{subfigure}[b]{\textwidth}
        \centering
        \footnotesize
        \renewcommand{\arraystretch}{0.8}
        \begin{tabular}{@{}ll@{\hspace{1.2em}}ll@{}}
            \tikz{\draw[line width=0.8mm, color=qwenThree] (0,0) -- (0.5,0);} &
            \texttt{Qwen3-8B} &
            \tikz{\draw[line width=0.8mm, color=qwenTwoFive, dashed] (0,0) -- (0.5,0);} &
            \texttt{Qwen2.5-7B} \\

            \tikz{\draw[line width=0.8mm, color=gemmaFour, dashdotted] (0,0) -- (0.5,0);} &
            \texttt{Gemma4-31B} &
            \tikz{\draw[line width=0.6mm, color=black, densely dotted] (0,0) -- (0.5,0);} &
            \texttt{Random}
        \end{tabular}
    \end{subfigure}

    \begin{minipage}[c]{0.055\linewidth}
        \centering
    \end{minipage}%
    \begin{minipage}[c]{0.935\linewidth}
        \centering
        \includegraphics[
            width=0.85\linewidth
        ]{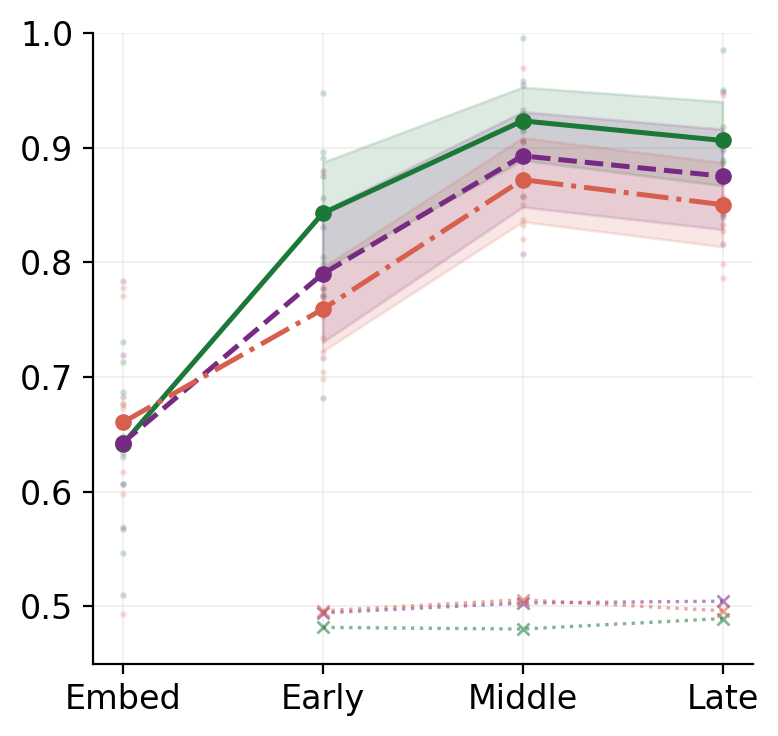}
    \end{minipage}

    \subcaption{Stage-wise mean AUROC}
    \label{fig:auprc-panel}
\end{subfigure}

    \caption{
\textbf{Layer-wise separability of reasoning operations across language models in middle token position.}
(A) Peak-layer one-vs-rest AUROC for each reasoning operation in each model, with 95\% confidence intervals.
(B) AUROC averaged across reasoning operations at different model depths (embedding, early, middle, and late layers), along with a random baseline.
Across models and reasoning operations, separability remains high and is strongest in the middle layers.
}
    \label{fig:overall-separability}
    \vspace{-1.3em}
\end{figure*}
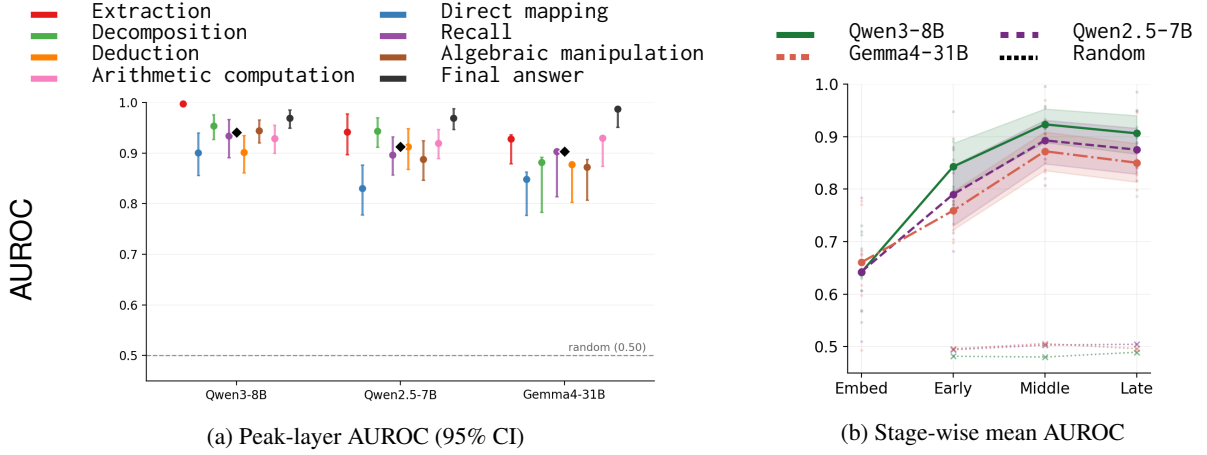
\begin{figure*}[t]
    \centering
    
    {
        \begin{subfigure}[b]{0.245\textwidth}
            \centering
            \includegraphics[width=\linewidth]{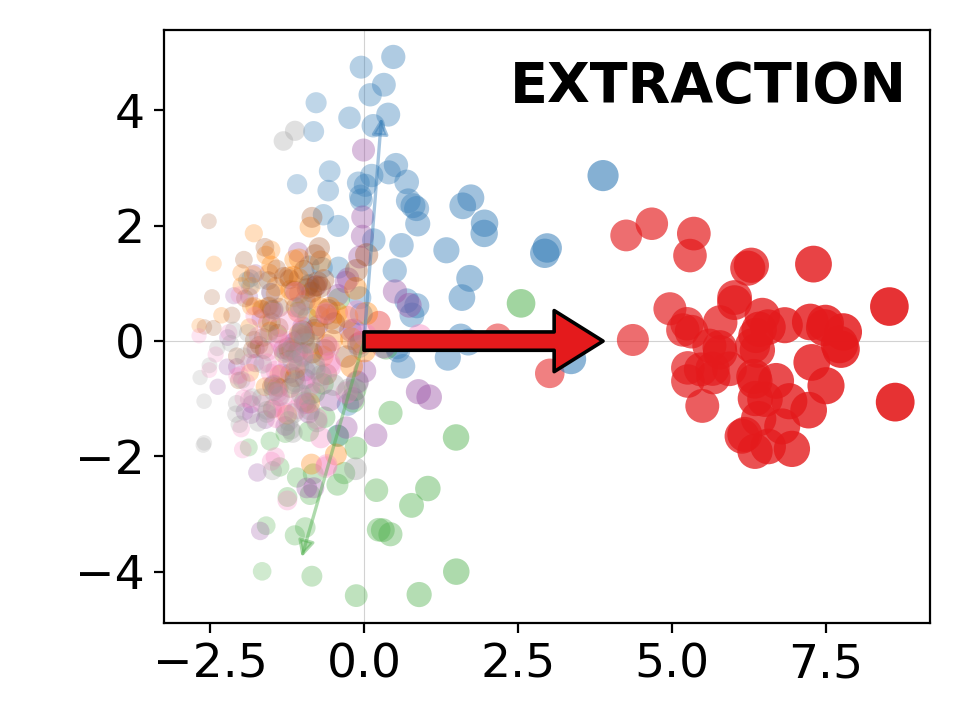}
            \caption{Extraction}
        \end{subfigure}
        \hfill
        \begin{subfigure}[b]{0.245\textwidth}
            \centering
            \includegraphics[width=\linewidth]{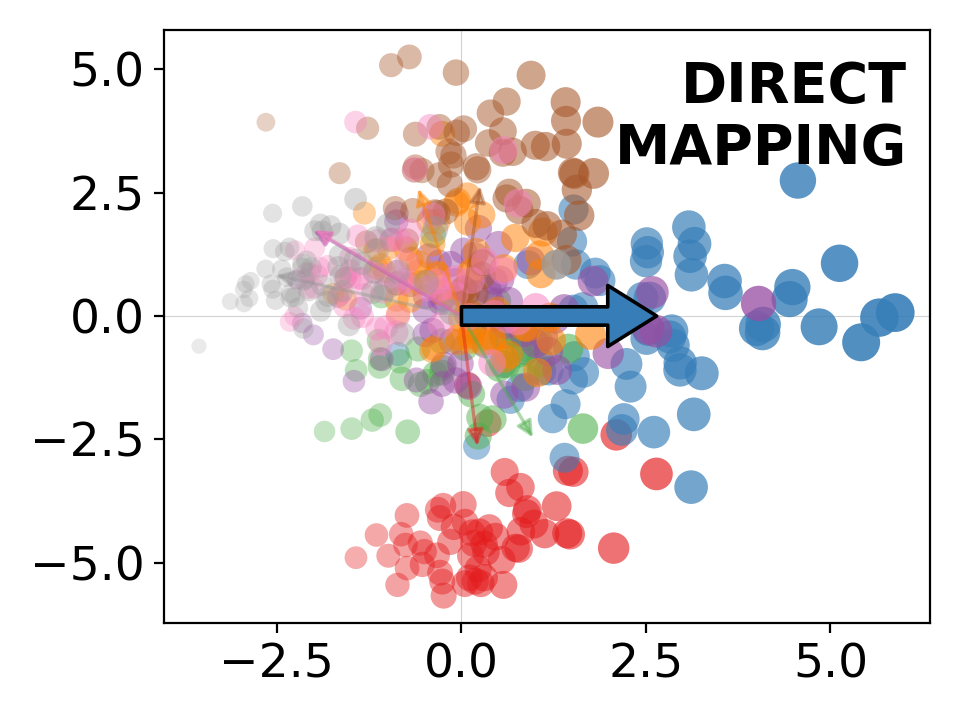}
            \caption{Direct mapping}
        \end{subfigure}
        \hfill
        \begin{subfigure}[b]{0.245\textwidth}
            \centering
            \includegraphics[width=\linewidth]{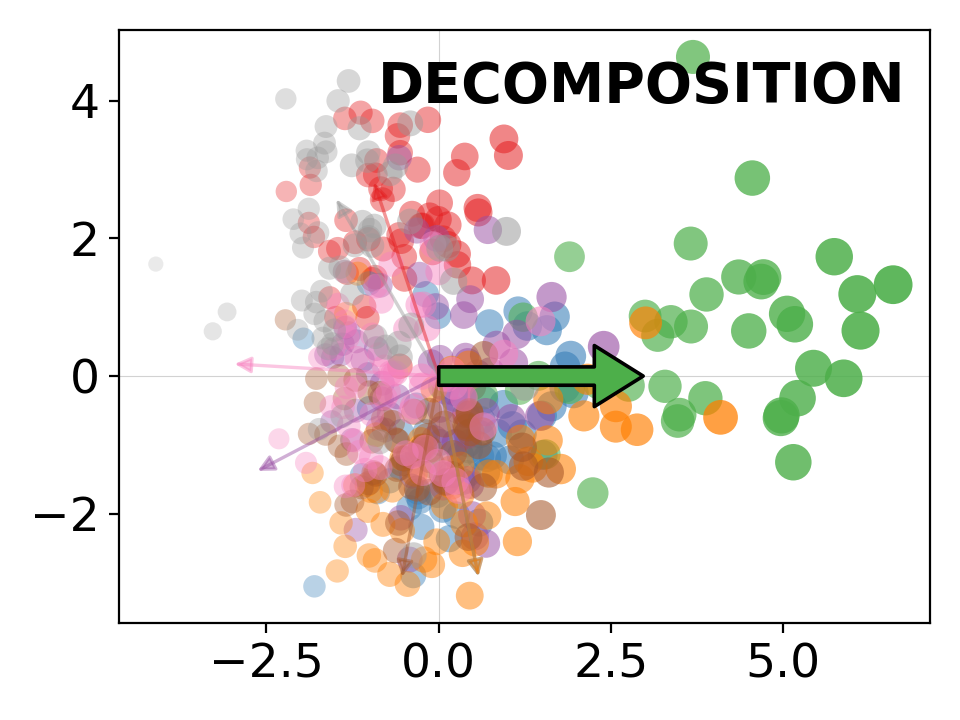}
            \caption{Decomposition}
        \end{subfigure}
        \hfill
        \begin{subfigure}[b]{0.245\textwidth}
            \centering
            \includegraphics[width=\linewidth]{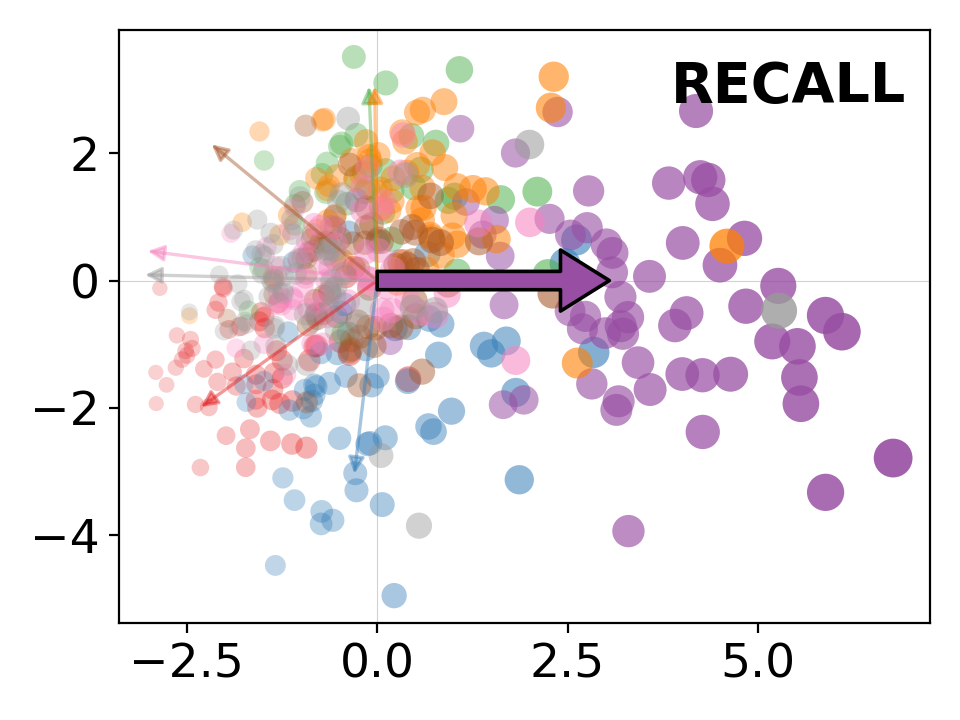}
            \caption{Recall}
        \end{subfigure}

        \vspace{4pt}

        \begin{subfigure}[b]{0.245\textwidth}
            \centering
            \includegraphics[width=\linewidth]{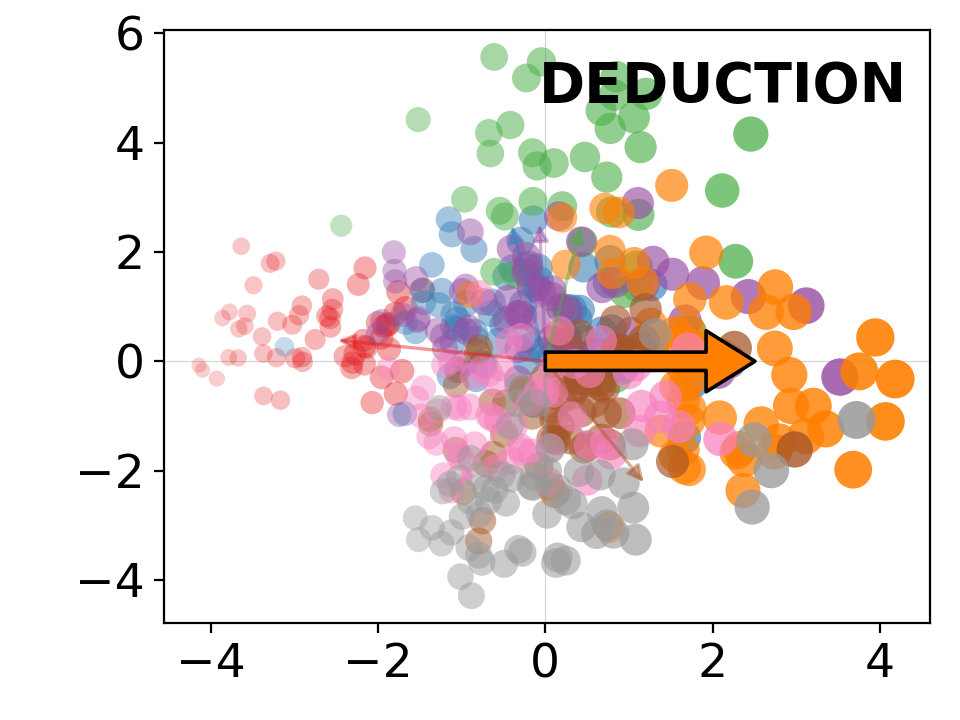}
            \caption{Deduction}
        \end{subfigure}
        \hfill
        \begin{subfigure}[b]{0.245\textwidth}
            \centering
            \includegraphics[width=\linewidth]{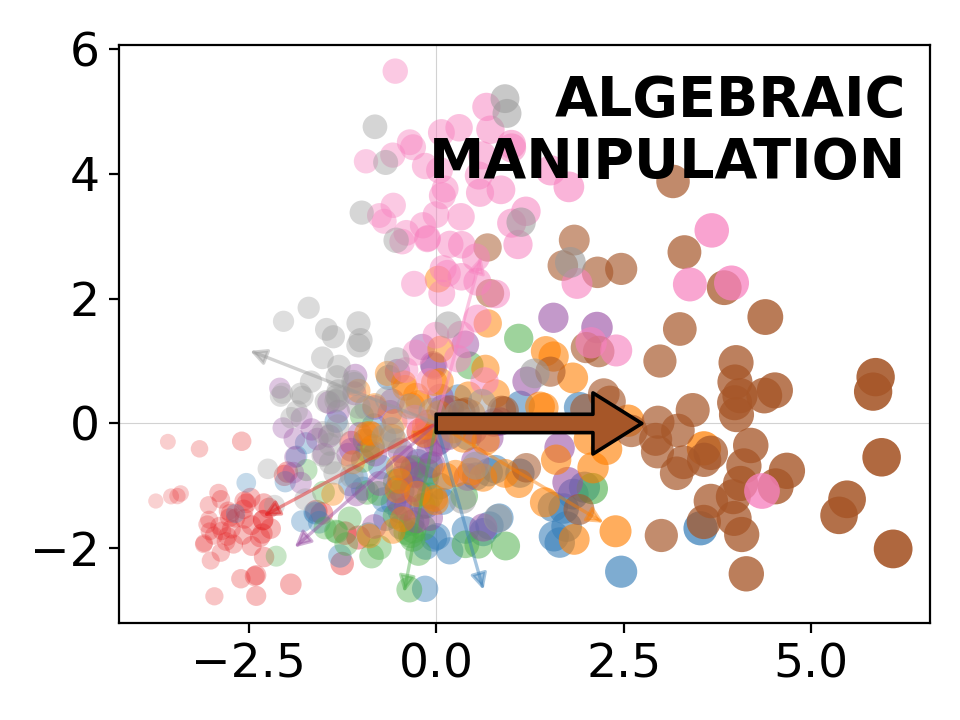}
            \caption{Algebraic Manipulation}
        \end{subfigure}
        \hfill
        \begin{subfigure}[b]{0.245\textwidth}
            \centering
            \includegraphics[width=\linewidth]{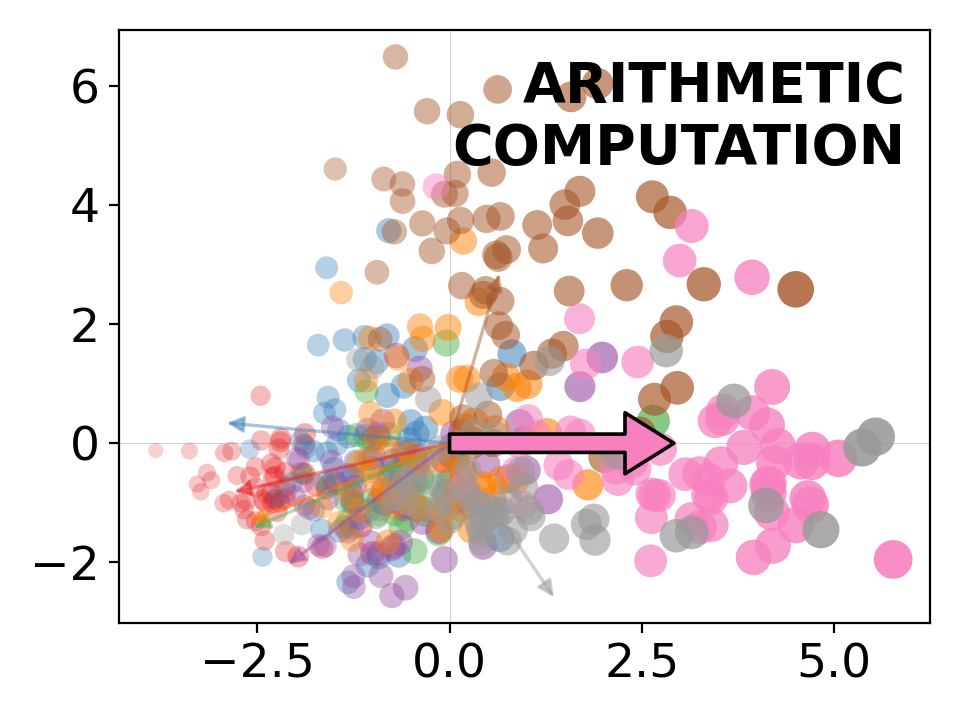}
            \caption{Arithmetic Computation}
        \end{subfigure}
        \hfill
        \begin{subfigure}[b]{0.245\textwidth}
            \centering
            \includegraphics[width=\linewidth]{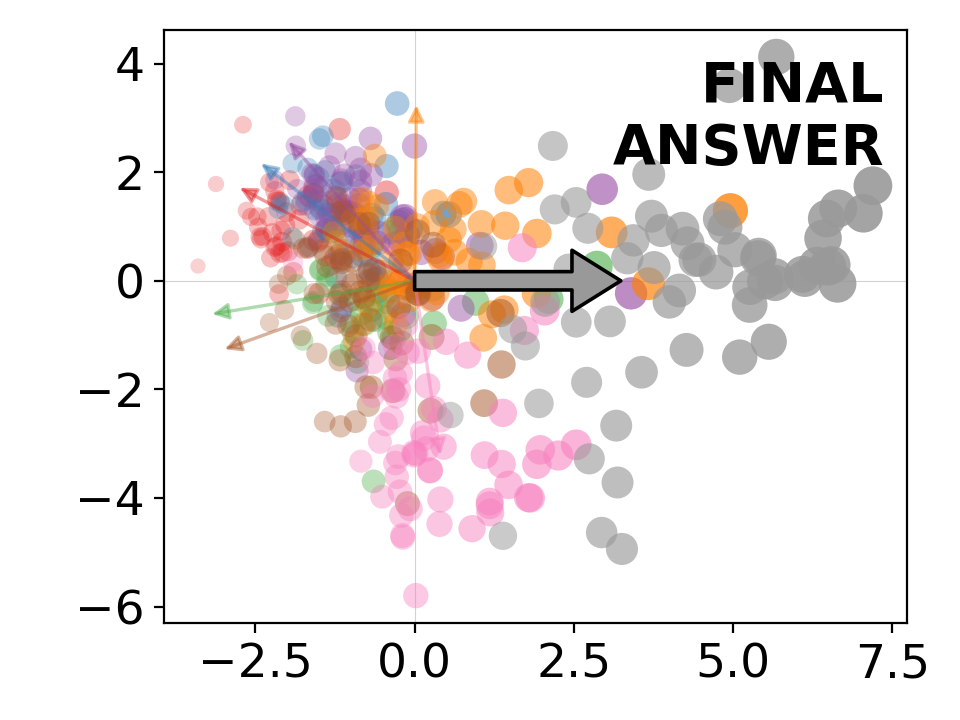}
            \caption{Final Answer}
        \end{subfigure}
    }
    \vspace{-1.5em}
    \caption{
\textbf{Visualization of reasoning-operation clusters in the learned LDA space.}
For each target operation, held-out span representations from \texttt{Qwen3-8B} are projected onto a two-dimensional plane whose $x$-axis is the corresponding reasoning operation vector $d_c$ and whose $y$-axis is the first principal component of the residual representations. 
Points are colored by their annotated reasoning operation labels.
}
    \label{fig:lda-2d-visualization}
    \vspace{-1.0em}
\end{figure*}
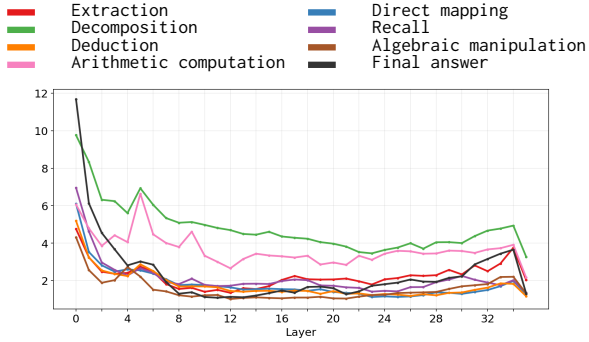
\begin{figure}[t]
    \centering

    {
        % \def\target{variance_layer}
        % \def\model{Qwen3-8B}

        % Legend
        \begin{subfigure}[b]{\linewidth}
            \centering
            \scriptsize
            \renewcommand{\arraystretch}{0.7}

            \resizebox{\linewidth}{!}{%
            \begin{tabular}{@{}ll@{\hspace{1em}}ll@{}}
                \tikz{\draw[line width=0.8mm, color=extraction] (0,0) -- (0.35,0);} &
                \texttt{Extraction} &
                \tikz{\draw[line width=0.8mm, color=symbolization] (0,0) -- (0.35,0);} &
                \texttt{Direct mapping} \\

                \tikz{\draw[line width=0.8mm, color=structural] (0,0) -- (0.35,0);} &
                \texttt{Decomposition} &
                \tikz{\draw[line width=0.8mm, color=retrieval] (0,0) -- (0.35,0);} &
                \texttt{Recall} \\

                \tikz{\draw[line width=0.8mm, color=deduction] (0,0) -- (0.35,0);} &
                \texttt{Deduction} &
                \tikz{\draw[line width=0.8mm, color=algebraic] (0,0) -- (0.35,0);} &
                \texttt{Algebraic manipulation} \\

                \tikz{\draw[line width=0.8mm, color=arithmetic] (0,0) -- (0.35,0);} &
                \texttt{Arithmetic computation} &
                \tikz{\draw[line width=0.8mm, color=finalanswer] (0,0) -- (0.35,0);} &
                \texttt{Final answer}
            \end{tabular}%
            }
        \end{subfigure}

        \vspace{0.3em}

        % Main image
        \includegraphics[width=0.9\linewidth]{
            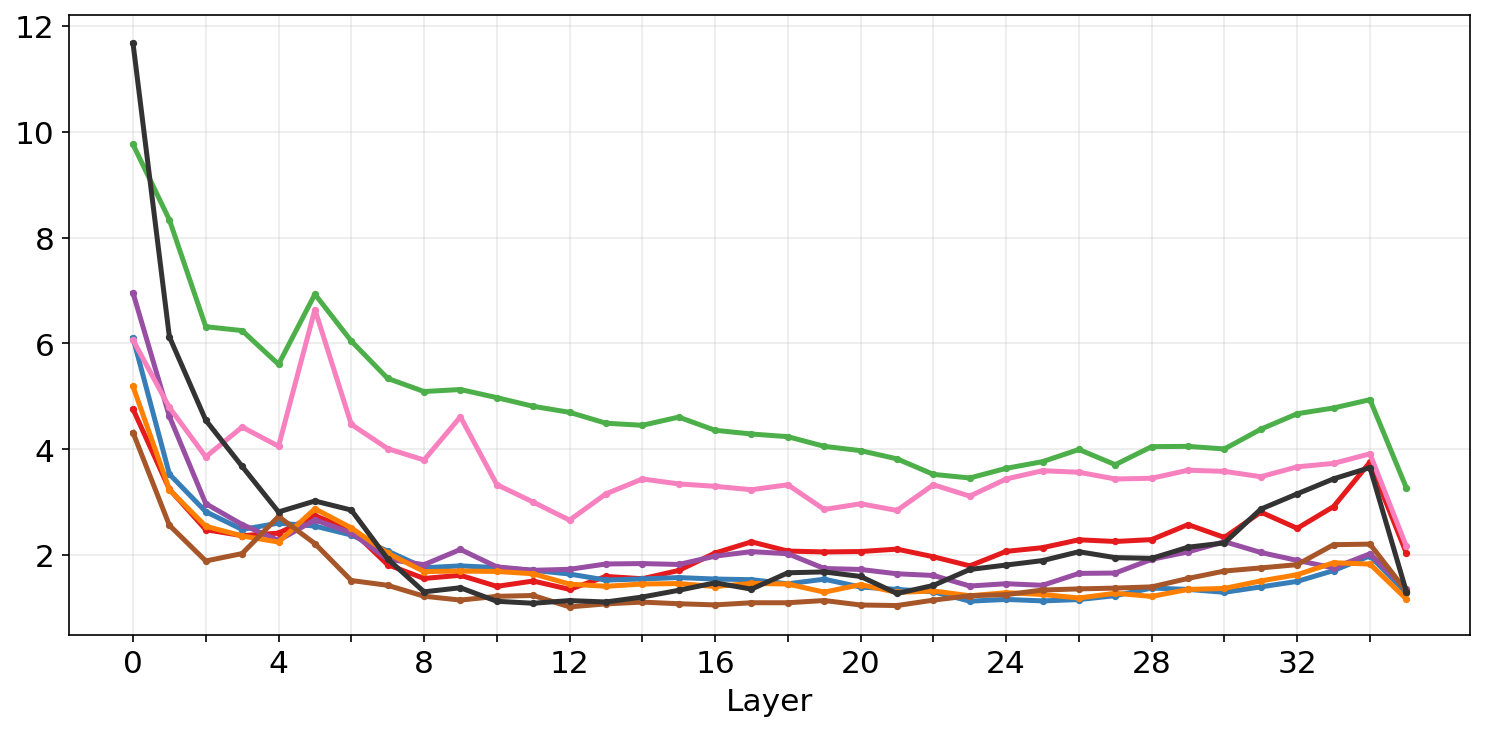
        }
    }

    \vspace{-1em}
    \caption{
        \textbf{Quantitative analysis on intra-span LDA score variances.}
        We investigate the LDA score variances within each span across layers         using Qwen3-8B. 
        Early-layer separability is often sparse and token-local, while mid--late-layer separability is distributed across the tokens.
        We further generalize the results for Qwen2.5-7B and Gemma4-31B in the Appendix.
    }
    \label{fig:lda-var-qwen3-8b}
    \vspace{-1.0em}
\end{figure}
\begin{figure*}[t]
    \centering
    {
        % \def\target{scatter_common_all}
        % \def\model{Qwen3-8B}

        % ── Row 2 ──────────────────────────────────────────────
        % \def\compare{DECOMPOSITION_vs_DEDUCTION}
        \begin{subfigure}[b]{0.05\textwidth}
            \centering
            \rotatebox{90}{%
                \parbox{2.5cm}{\centering
                    \small X:\textcolor{deduction}{Deduction}    \\
                    Y:\textcolor{structural}{Decomposition}   
                }
            }
        \end{subfigure}
        \begin{subfigure}[b]{0.15\textwidth}
            \centering
            \includegraphics[width=2.5cm, height=2.5cm]{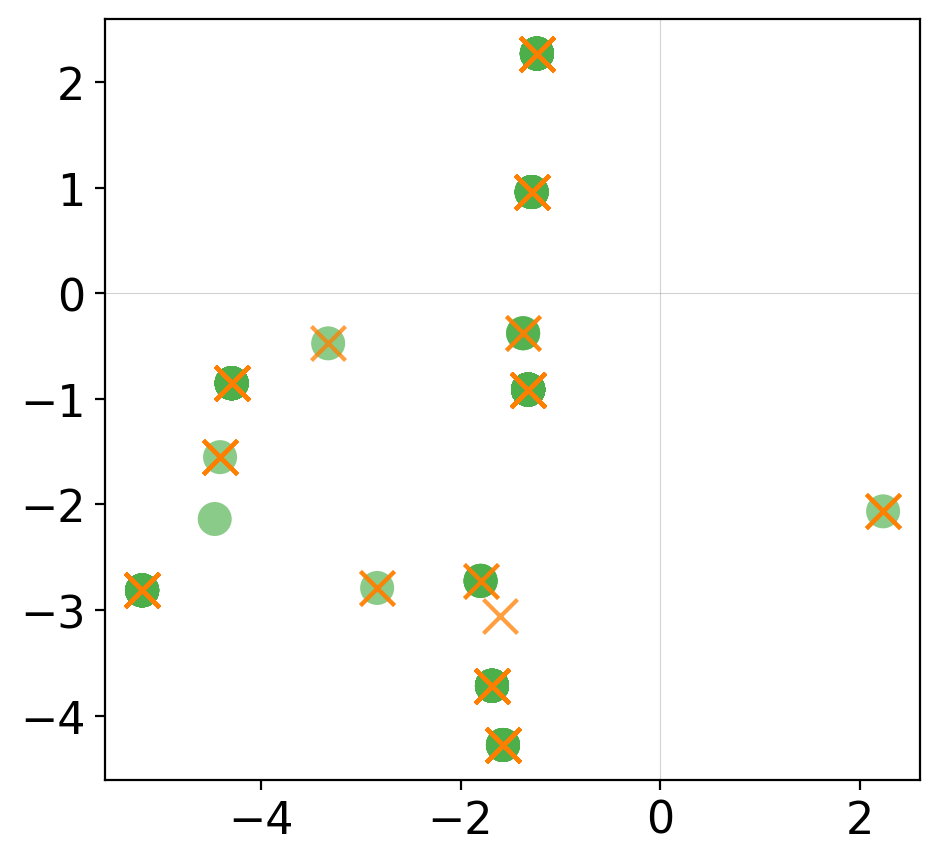}
        \end{subfigure}
        \begin{subfigure}[b]{0.15\textwidth}
            \centering
            \includegraphics[width=2.5cm, height=2.5cm]{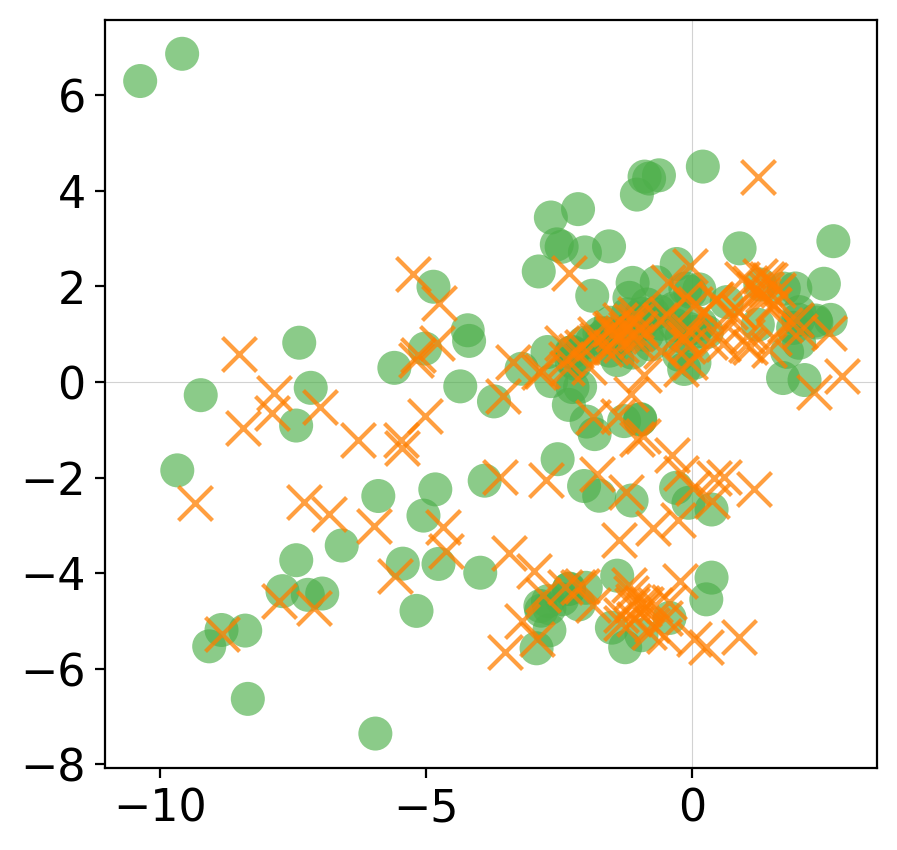}
        \end{subfigure}
        \begin{subfigure}[b]{0.15\textwidth}
            \centering
            \includegraphics[width=2.5cm, height=2.5cm]{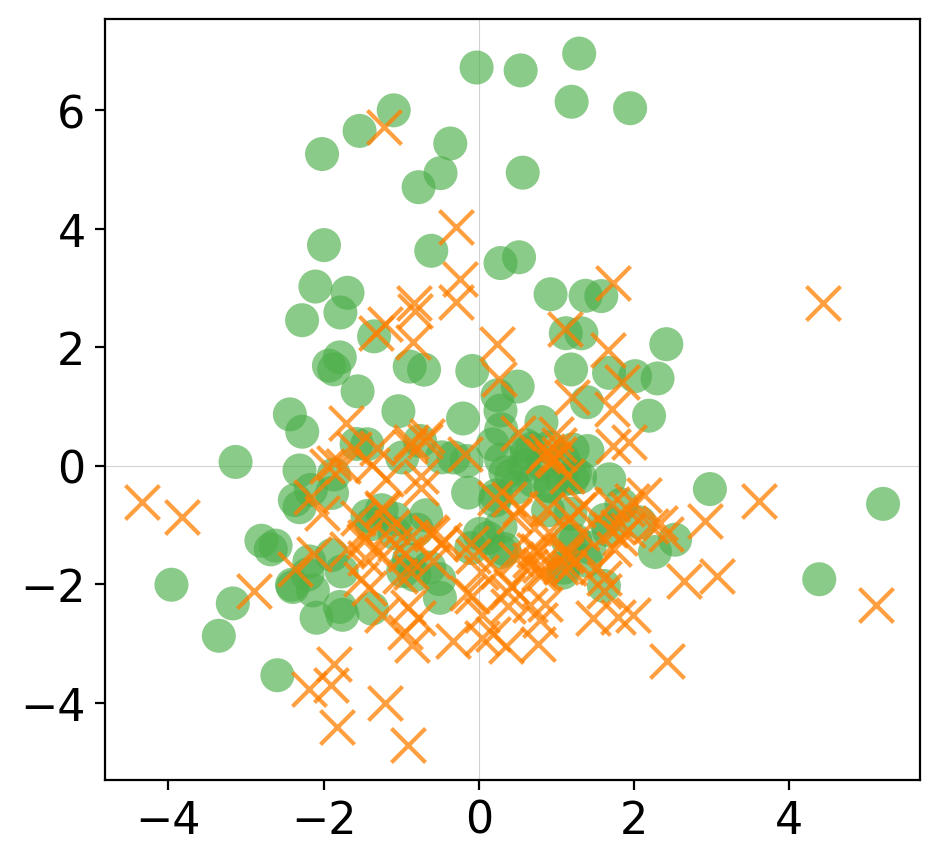}
        \end{subfigure}
        \begin{subfigure}[b]{0.15\textwidth}
            \centering
            \includegraphics[width=2.5cm, height=2.5cm]{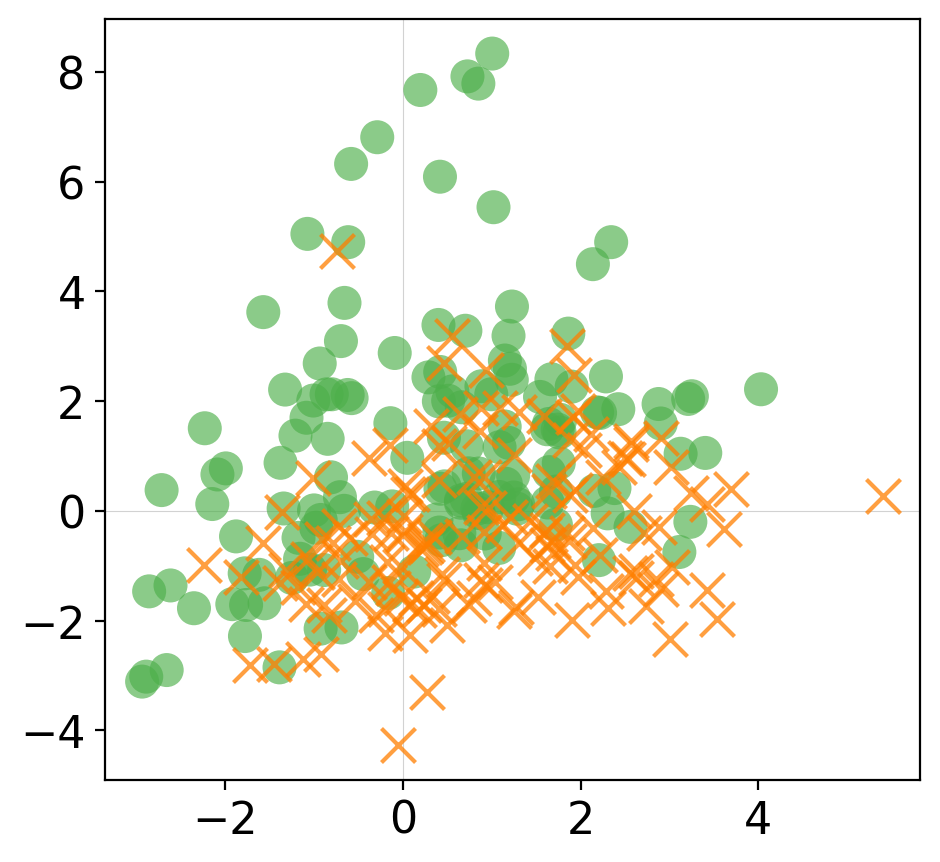}
        \end{subfigure}
        \begin{subfigure}[b]{0.15\textwidth}
            \centering
            \includegraphics[width=2.5cm, height=2.5cm]{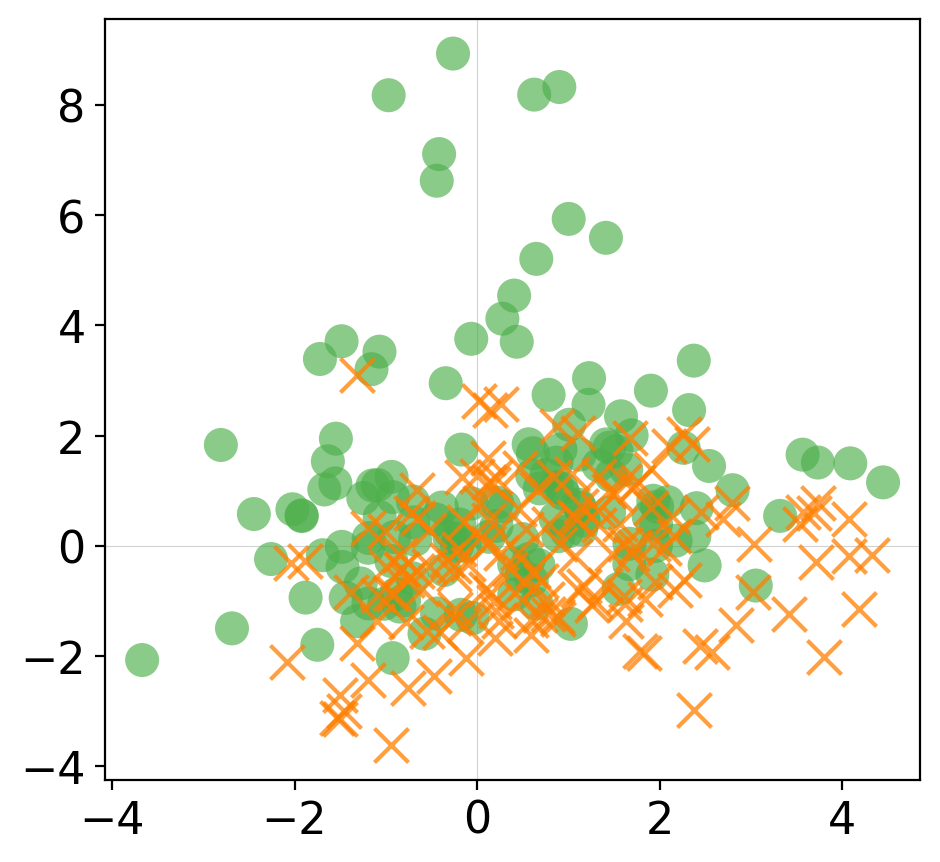}
        \end{subfigure}
        \begin{subfigure}[b]{0.15\textwidth}
            \centering
            \includegraphics[width=2.5cm, height=2.5cm]{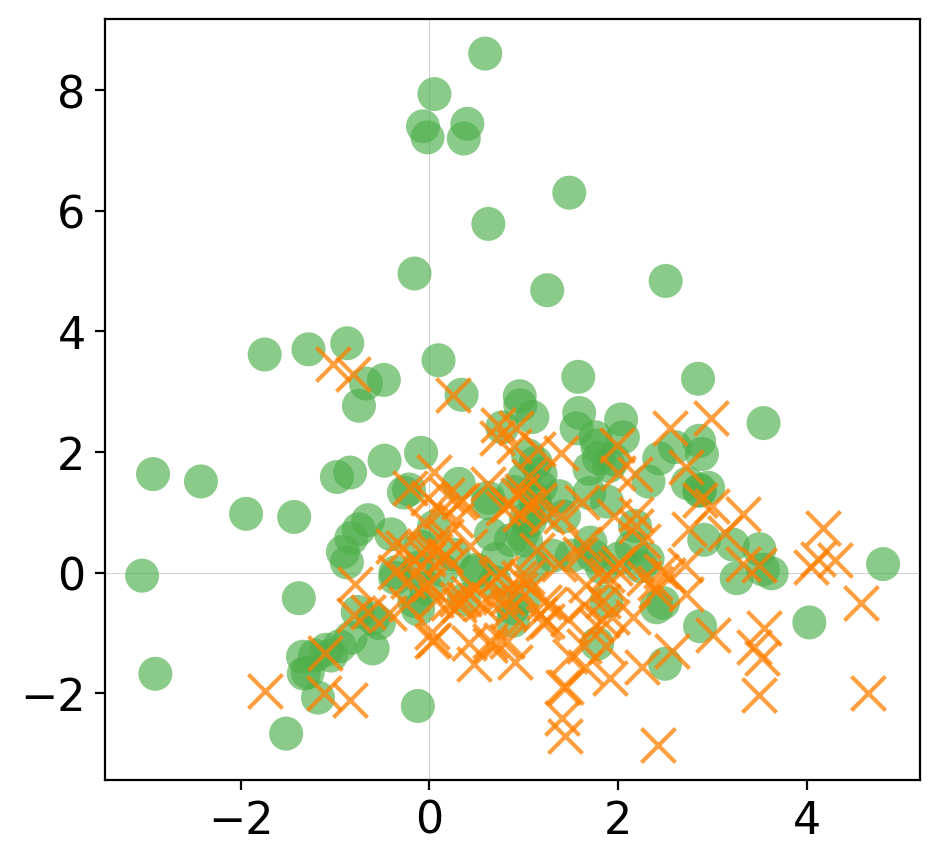}
        \end{subfigure}
        
        % ── Row 4  ───────────────────────────────
         % \def\compare{RECALL_vs_ALGEBRAIC-MANIPULATION}
        \begin{subfigure}[b]{0.05\textwidth}
            \centering
            \rotatebox{90}{%
                \parbox{2.5cm}{\centering
                    \small X:  \textcolor{symbolization}{Direct mapping}  \\
                    Y:  \textcolor{retrieval}{Recall}  
                }
            }
        \end{subfigure}
        \begin{subfigure}[b]{0.15\textwidth}
            \centering
            \includegraphics[width=2.5cm, height=2.5cm]{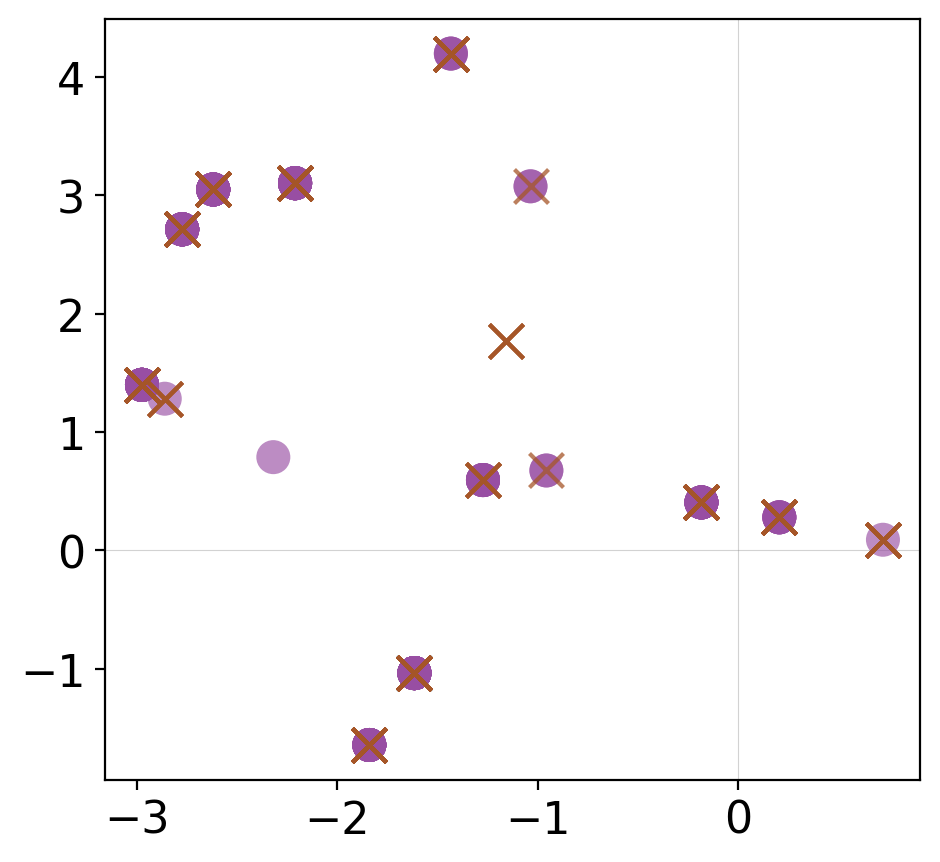}
        \end{subfigure}
        \begin{subfigure}[b]{0.15\textwidth}
            \centering
            \includegraphics[width=2.5cm, height=2.5cm]{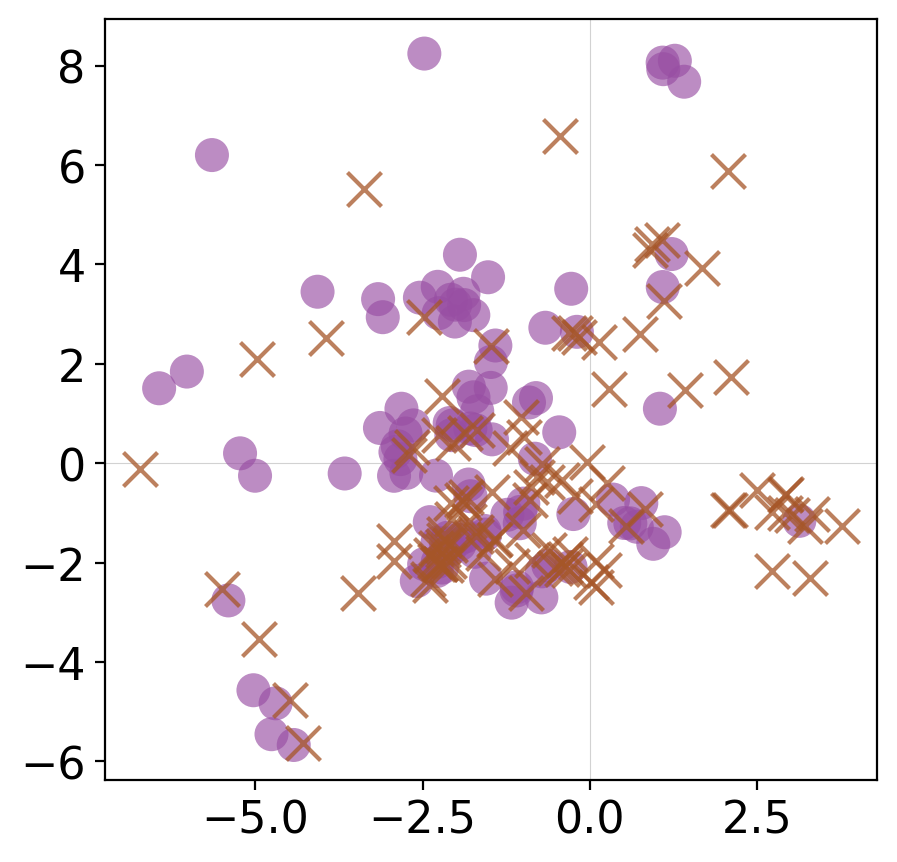}
        \end{subfigure}
        \begin{subfigure}[b]{0.15\textwidth}
            \centering
            \includegraphics[width=2.5cm, height=2.5cm]{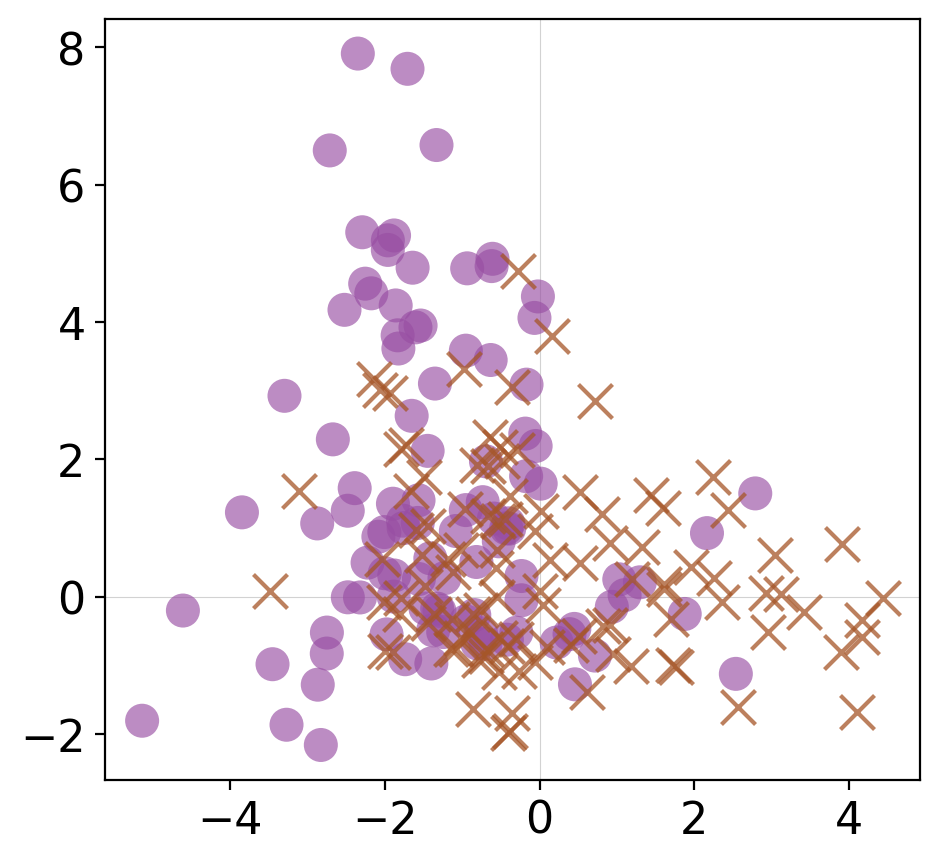}
        \end{subfigure}
        \begin{subfigure}[b]{0.15\textwidth}
            \centering
            \includegraphics[width=2.5cm, height=2.5cm]{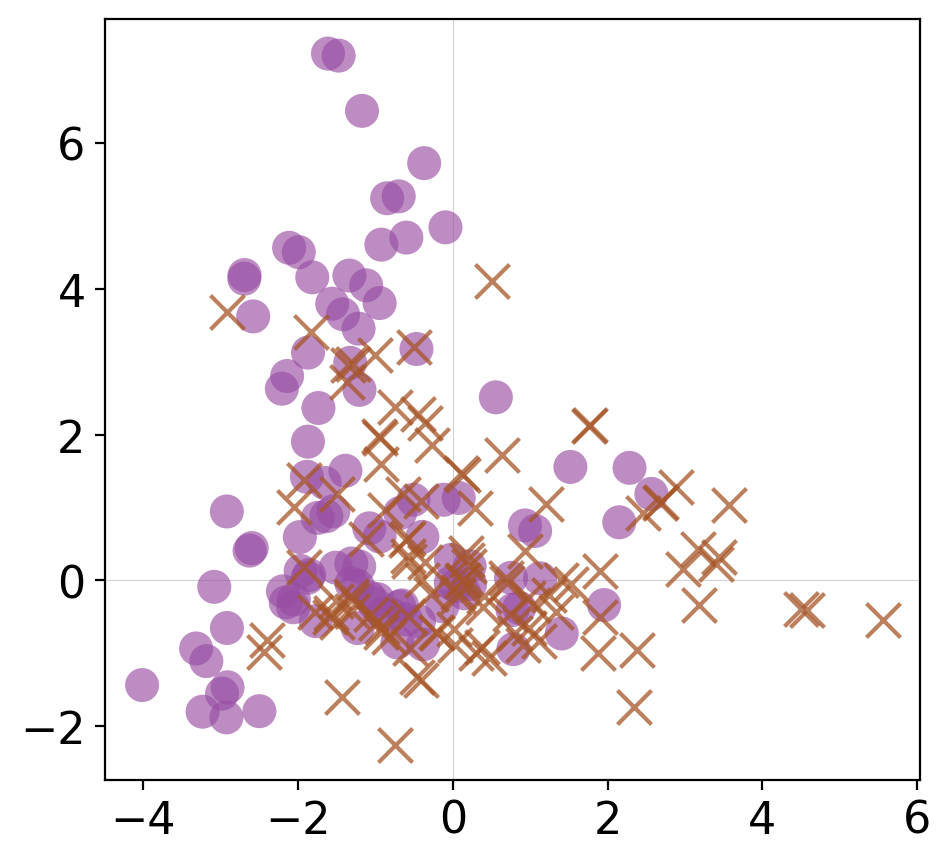}
        \end{subfigure}
        \begin{subfigure}[b]{0.15\textwidth}
            \centering
            \includegraphics[width=2.5cm, height=2.5cm]{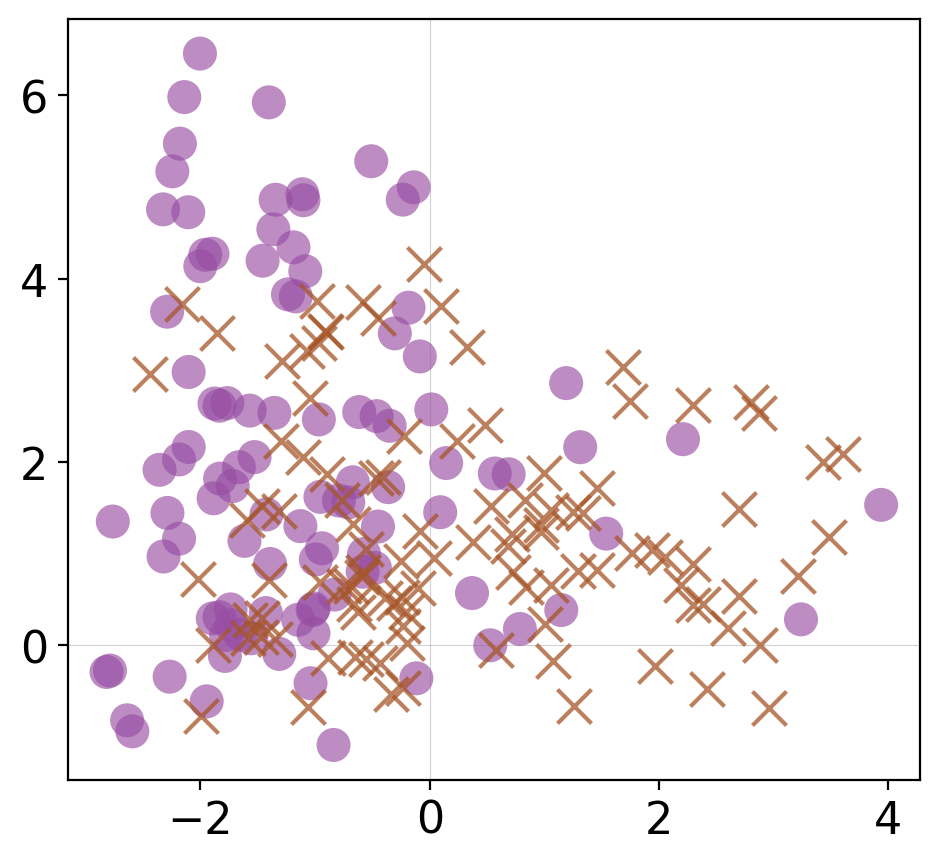}
        \end{subfigure}
        \begin{subfigure}[b]{0.15\textwidth}
            \centering
            \includegraphics[width=2.5cm, height=2.5cm]{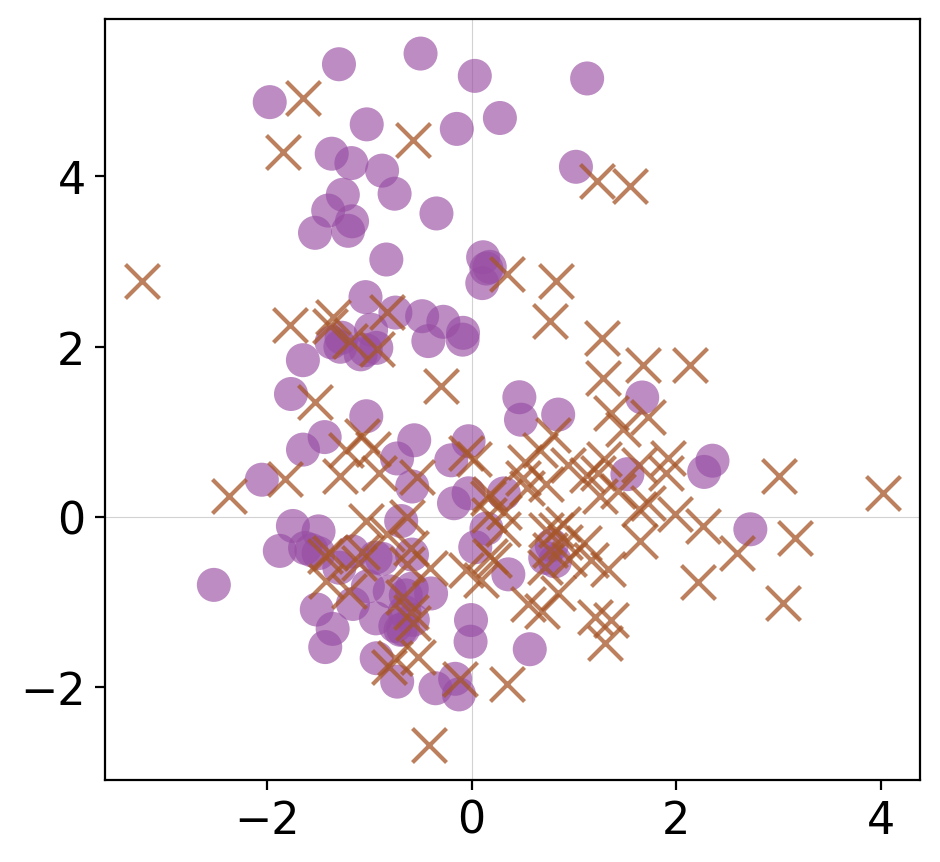}
        \end{subfigure}

        % ── Row 4 (with captions) ───────────────────────────────
        
         % \def\compare{RECALL_vs_FINAL-ANSWER}
        \begin{subfigure}[b]{0.05\textwidth}
            \centering
            \rotatebox{90}{%
                \parbox{2.5cm}{\centering
                    \small X:  \textcolor{Gray}{Final answer}   \\
                    Y:  \textcolor{retrieval}{Recall}  
                }
            }
        \end{subfigure}
        \begin{subfigure}[b]{0.15\textwidth}
            \centering
            \includegraphics[width=2.5cm, height=2.5cm]{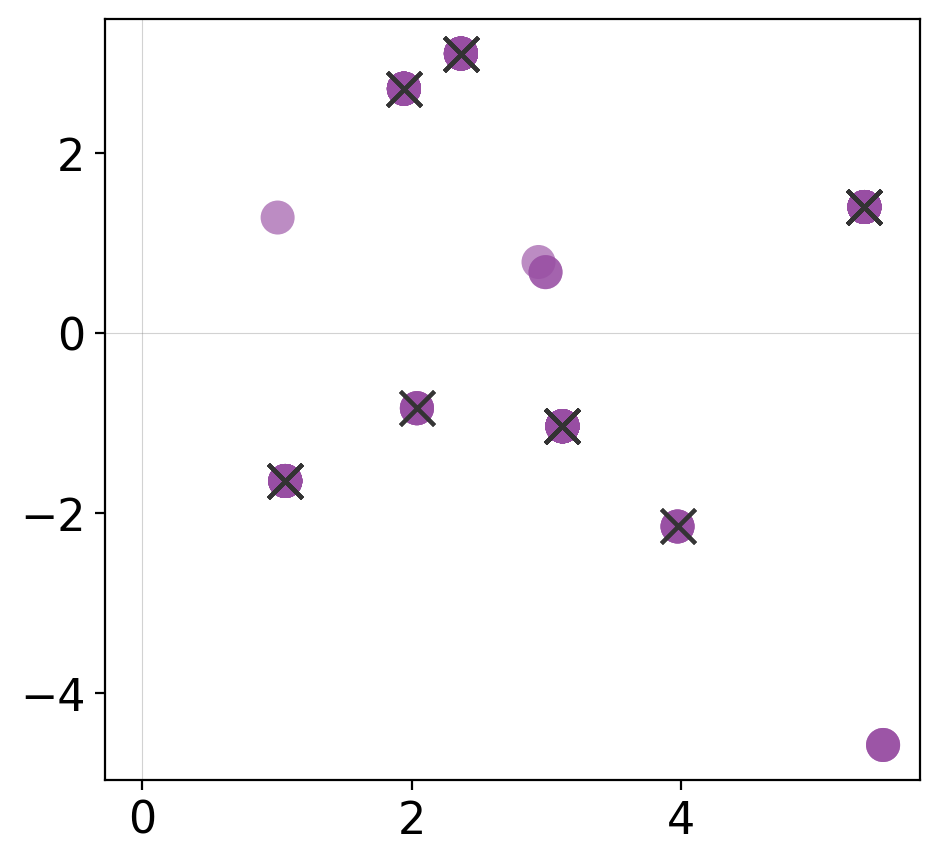}
            \caption{Emb. layer}
        \end{subfigure}
        \begin{subfigure}[b]{0.15\textwidth}
            \centering
            \includegraphics[width=2.5cm, height=2.5cm]{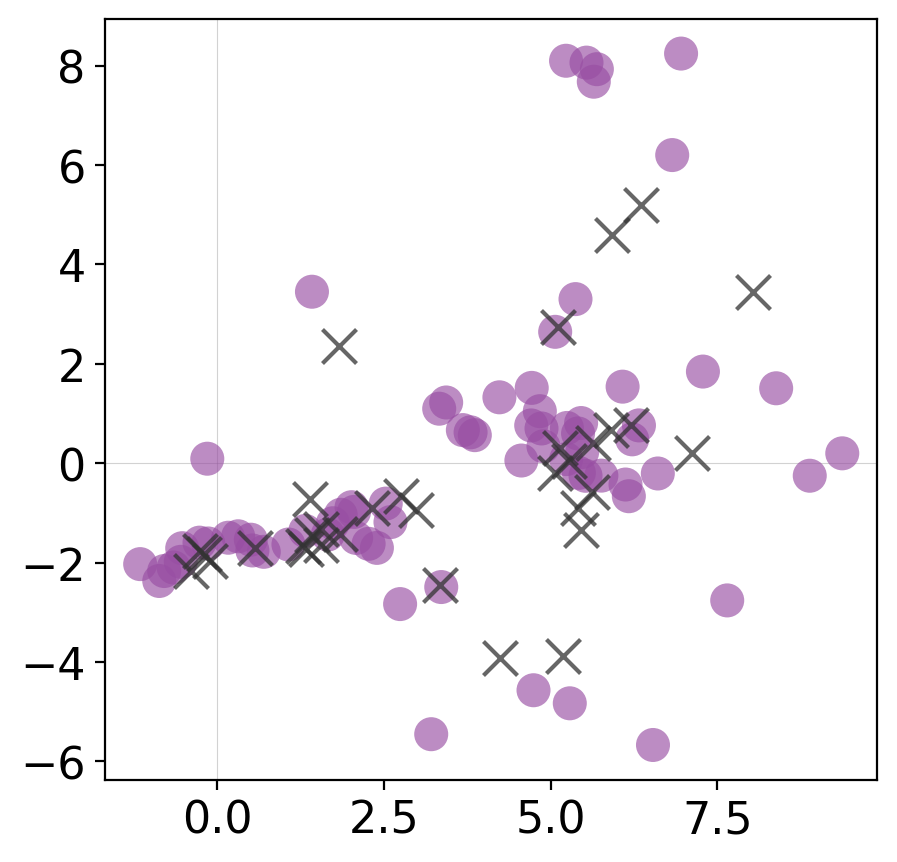}
            \caption{Layer 1}
        \end{subfigure}
        \begin{subfigure}[b]{0.15\textwidth}
            \centering
            \includegraphics[width=2.5cm, height=2.5cm]{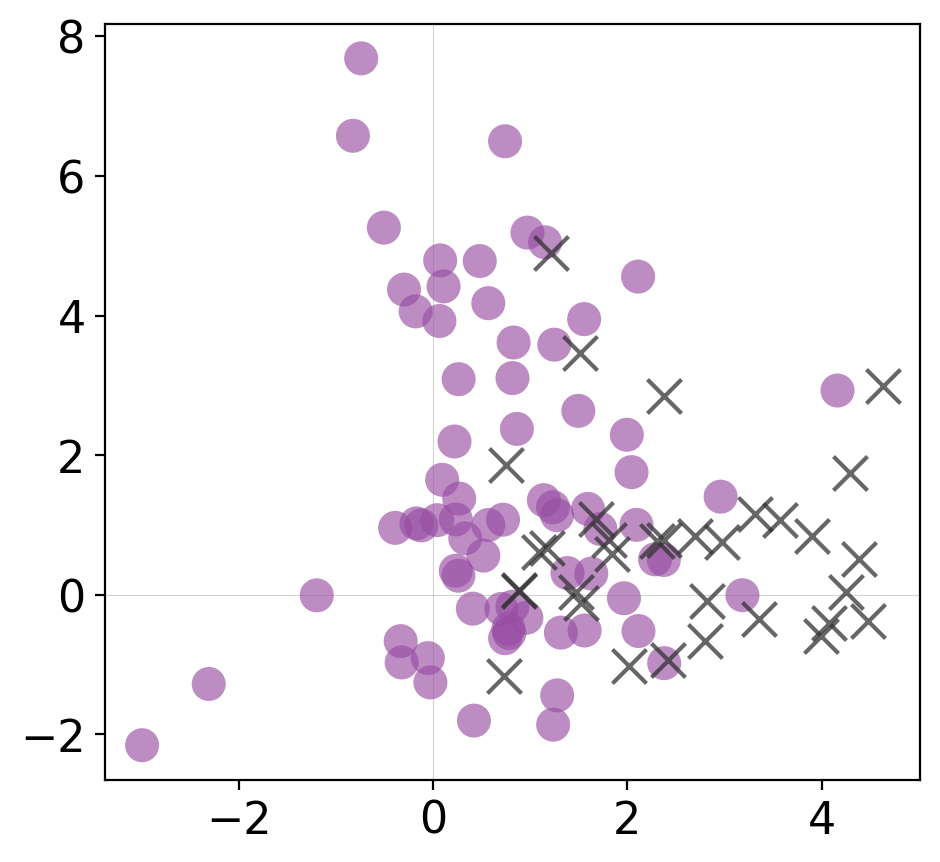}
            \caption{Layer 11}
        \end{subfigure}
        \begin{subfigure}[b]{0.15\textwidth}
            \centering
            \includegraphics[width=2.5cm, height=2.5cm]{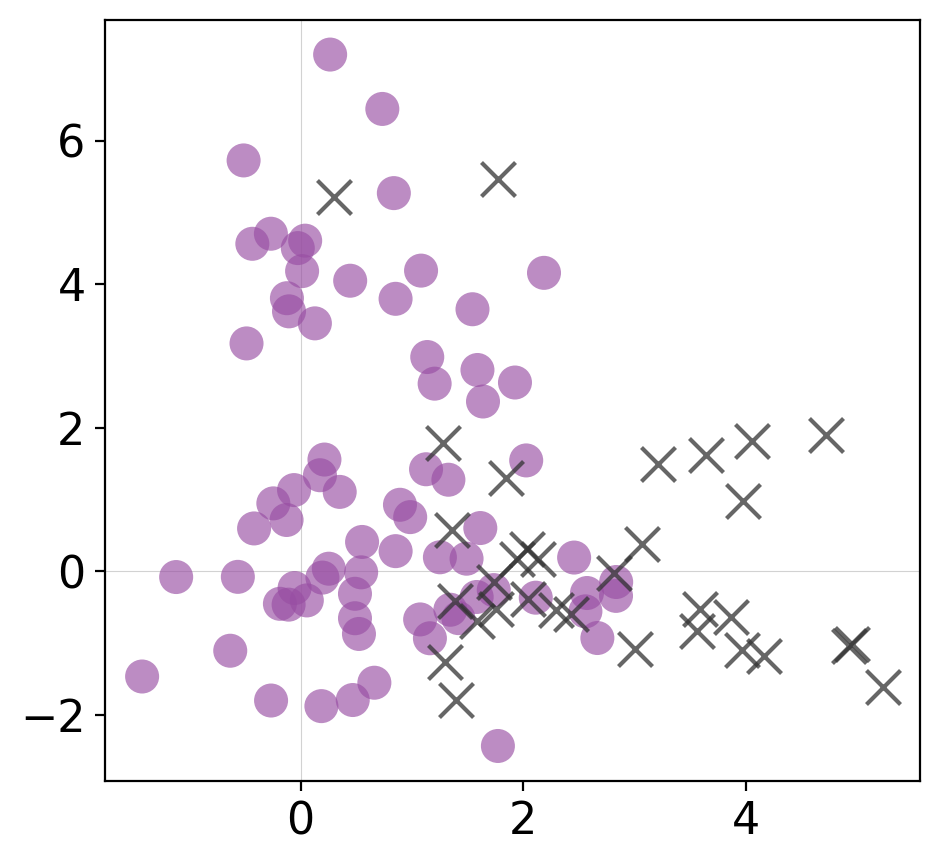}
            \caption{Layer 21}
        \end{subfigure}
        \begin{subfigure}[b]{0.15\textwidth}
            \centering
            \includegraphics[width=2.5cm, height=2.5cm]{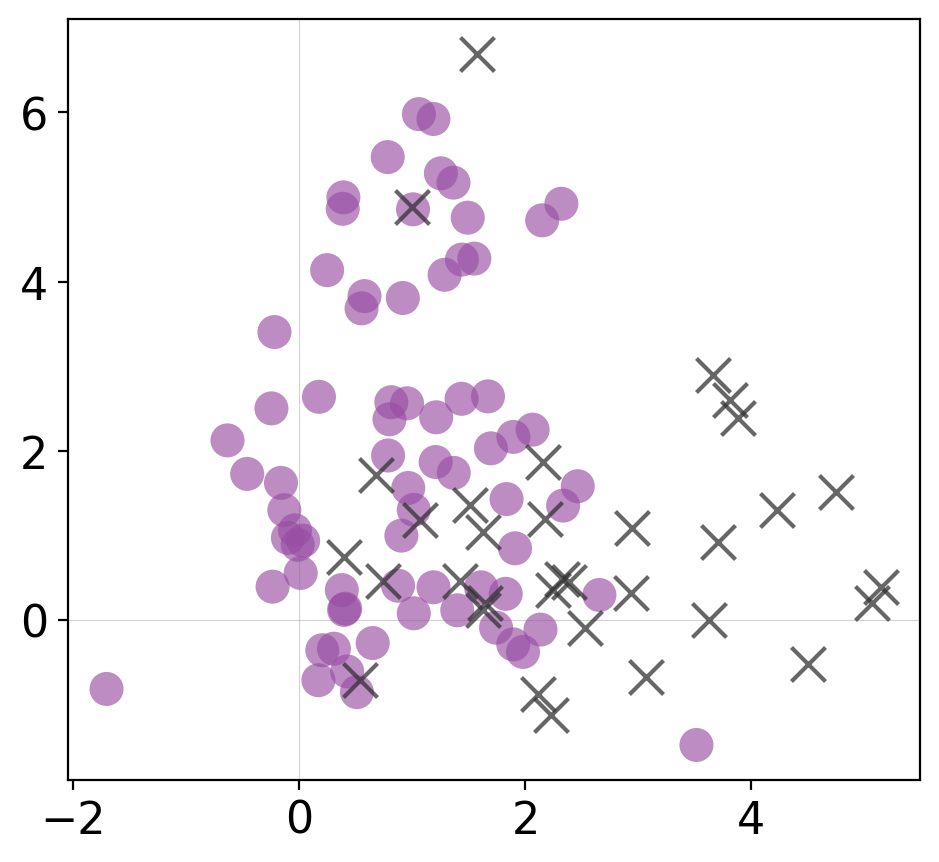}
            \caption{Layer 27}
        \end{subfigure}
        \begin{subfigure}[b]{0.15\textwidth}
            \centering
            \includegraphics[width=2.5cm, height=2.5cm]{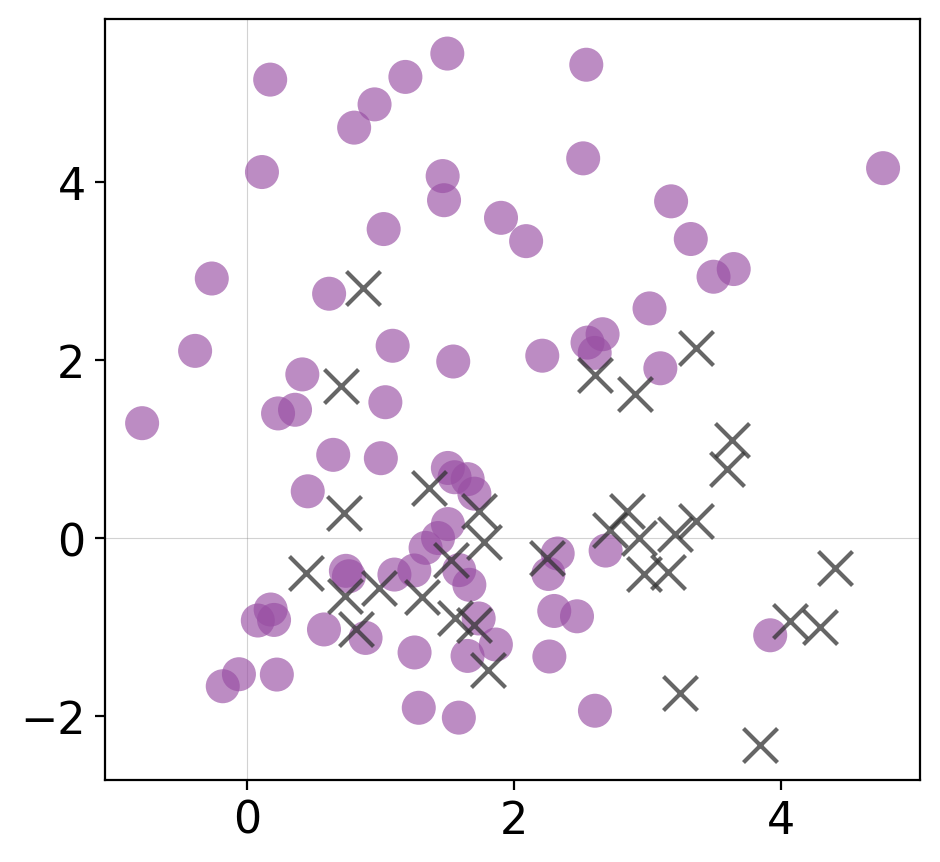}
            \caption{Layer 36}
        \end{subfigure}
        
    }
    \vspace{-.5em}
    \caption{
\textbf{Identical tokens acquire reasoning-dependent representations.}
For pairs of reasoning operations, we collect common surface tokens that appear in both operation contexts and visualize their layer-wise scores along the two corresponding reasoning operation vectors in \texttt{Qwen3-8B}. 
Each point is an occurrence of a shared token and is colored by its annotated reasoning label. 
}
    \label{fig:common-word-scatter}
    \vspace{-1.0em}
\end{figure*}

\subsection{Experimental Setup}

\subsubsection{Data, Models, and Reasoning Taxonomy}
\noindent\textbf{Tasks and data.}
We use two reasoning datasets, DAPO-Math-17K~\citep{yu2025dapo} and TheoremQA~\citep{chen-etal-2023-theoremqa}. DAPO-Math-17K consists of mathematical reasoning problems, while TheoremQA contains theorem-driven questions that require applying domain knowledge to solve problems.
Using both datasets allows us to examine reasoning operations in computation-heavy mathematical reasoning as well as more theorem-based reasoning settings.

\noindent\textbf{Models.}
We analyze three reasoning LLMs: \texttt{Qwen2.5-7B}~\citep{qwen2025qwen25technicalreport}, \texttt{Qwen3-8B}~\citep{yang2025qwen3technicalreport}, and \texttt{Gemma4-31B}~\citep{gemmateam2026gemma4technicalreport}. 
This validates the generalizability of our observations beyond a specific model family.

\noindent\textbf{Reasoning Taxonomy.} We use the eight main reasoning operation types introduced in the Preliminary section: 
\textcolor{extraction}{Extraction}, \textcolor{symbolization}{Direct mapping},  \textcolor{structural}{Decomposition}, \textcolor{retrieval}{Recall}, \textcolor{deduction}{Deduction}, \textcolor{algebraic}{Algebraic manipulation}, \textcolor{arithmetic}{Arithmetic computation}, and \textcolor{Gray}{Final-answer}.

\subsubsection{Operation-Span Annotation and Human Validation}

\noindent\textbf{Operation-span annotation.}
For each model and dataset, we generate reasoning traces and retain 500--700
correct responses. Each trace $r$ is segmented into non-overlapping reasoning
operation spans $\{s_1, s_2, \ldots, s_K\}$, where
$s_i = (t_i^{\mathrm{start}}, t_i^{\mathrm{end}}, y_i)$ denotes a contiguous
token range $[t_i^{\mathrm{start}}, t_i^{\mathrm{end}}]$ with operation label
$y_i$. We use GPT-5 to assign each span one of the eight main reasoning operation labels, where the prompts are in Appendix~\ref{app:annotation_prompt}.
For spans longer than 50 tokens, we select a 50-token window centered around the token with the highest entropy; shorter spans are used in full. Spans longer than 300 tokens are excluded.

\noindent\textbf{Validation of LLM annotation.}
To assess annotation reliability, seven human annotators, including three authors, all Korean with at least a bachelor's degree in engineering or mathematics, annotated 84 sampled spans, each independently labeled by three annotators.
A human-majority label was defined as agreement by at least two annotators on the same canonical operation label.
Annotators reached majority agreement for 81 of 84 spans (96.4\%; Fleiss' $\kappa=0.666$), and GPT-5 matched the human-majority label in 64 cases (76.2\%; Cohen's $\kappa=0.715$).
Uniform random guessing over eight labels would yield 12.5\% expected exact agreement.
These results support aggregate representation-level analyses while indicating non-negligible annotation uncertainty.
Full instructions, annotator assignments, and agreement analyses are in Appendix~\ref{app:human-validation}.

\subsubsection{Hidden-State Probing and Evaluation}

For each annotated span, we extract hidden representations from every layer and representation type. 
In the main results, we use the middle-token representation of each span. Let $x_i^{(l,m)}$ denote the representation of span $i$ at layer $l$ and representation type $m$.
For each layer and representation type, we fit the probe using only the training split. 
We $L_2$-normalize span representations, apply PCA to 128 dimensions, and fit supervised LDA using operation labels. 
We then apply the learned PCA and LDA projections to the test split. 
Since there are $C=8$ operation classes, the LDA space has at most $C-1=7$ dimensions; accordingly, we use all seven LDA dimensions in our analysis. 
Let
$z_i^{(l,m)}$ denote the projected representation of $x_i^{(l,m)}$.

For each operation $c$, we define a one-vs-rest operation vector:
{
\setlength{\abovedisplayskip}{4pt}
\setlength{\belowdisplayskip}{4pt}
\[
d_c^{(l,m)}
=
\frac{
\mu_c^{(l,m)} - \mu_{\neg c}^{(l,m)}
}{
\left\|
\mu_c^{(l,m)} - \mu_{\neg c}^{(l,m)}
\right\|_2
},
\]
}
where $\mu_c^{(l,m)}$ is the mean representation of training spans labeled as $c$, and
$\mu_{\neg c}^{(l,m)}$ is the mean representation of all remaining training spans. For
each held-out span $i$, its alignment with operation $c$ is
{
\setlength{\abovedisplayskip}{4pt}
\setlength{\belowdisplayskip}{4pt}
\[
a(i,c)
=
\left(z_i^{(l,m)}\right)^\top d_c^{(l,m)}.
\]
}
We evaluate each operation as a one-vs-rest classification task, with binary
target $g(i,c)=1_{[y_i=c]}$ and prediction score $a(i,c)$. We report AUROC for middle-token representations in the main text.
All normalization statistics, PCA components, LDA projections, and operation directions are estimated exclusively from the training split and then applied without refitting to the held-out test split.
To limit class imbalance, we sample at most 300 training and 60 test spans per operation without replacement. All splits are constructed before probe fitting.
As matched controls, we repeat the pipeline with randomly assigned labels and randomly selected token positions. 
Full setup and AUROC/AUPRC results for first-token, last-token, and mean-pooled representations are reported in
Appendix~\ref{app:separability_details}. 

% \subsection{Representational Separability of Reasoning Operations}
\subsection{Reasoning Operations Are Separable in Held-Out Hidden Representations}
\label{sec:separability}

\subsubsection{Separability of Reasoning Operations based on LDA}
\label{sec:main_separability}

We first ask whether textually annotated reasoning operations are separable in held-out hidden representations. To assess this, we evaluate operation-level separability using middle-token representations across \texttt{Qwen2.5-7B}, \texttt{Qwen3-8B}, and \texttt{Gemma4-31B}. Figure~\ref{fig:overall-separability} summarizes the results across models, reasoning operations, and model depth, showing consistently high peak-layer one-vs-rest AUROC across a broad range of reasoning operations in all three models, with separability strongest in the middle layers. Together, these results indicate that \textit{reasoning operations are consistently separable across models and operation types, with operation-level information most strongly expressed in intermediate representations}. Detailed results across individual layers and span-representation choices are reported in Appendix~\ref{app:position_variants}.

\noindent\textbf{Qualitative visualization.}
Figure~\ref{fig:lda-2d-visualization} visualizes held-out spans in the learned LDA probe space for \texttt{Qwen3-8B}. 
For each target operation, the $x$-axis is its operation vector $d_c$, and the $y$-axis is the first principal component of the residual representation after removing the projection onto $d_c$. 
The resulting projections illustrate operation-aligned organization consistent with the held-out AUROC and AUPRC results. 
We use this visualization as a qualitative illustration; the quantitative evidence for separability comes from the held-out evaluation above.

\noindent\textbf{Statistical reliability.}
To ensure these gaps are not artifacts of the modest and imbalanced span counts available for some operations (Table~\ref{tab:train-test-reasoning-occurrence}), we assess the statistical reliability of our results using stratified bootstrap confidence intervals and random-label permutation tests.
Full procedures and per-operation results are provided in Appendix~\ref{app:statistical-tests}.

%%%%%
\subsubsection{Robustness to Lexical Confounds}
\label{sec:lexical_confounds}

A central alternative explanation is that the observed separability reflects
surface lexical or statistical regularities associated with each reasoning
operation, rather than operation-related structure in the hidden
representations. We therefore evaluate this possibility using four
complementary controls: text-only classification, lexically matched
comparisons, competing-operation vocabulary subsets, and a targeted control
for digit and formula density.

\noindent\textbf{Text-only classification.}
Lexical content is informative, but does not account for the full hidden-state separability. 
We train bag-of-words and TF--IDF logistic-regression classifiers using the same splits, class balancing, and operation labels as the hidden-state analysis. 
As in Table~\ref{tab:surface-controls}, across all three models, the mean-pooled hidden-state probe outperforms the strongest text-only baseline in both macro AUROC and AUPRC. 
The improvement ranges from $0.041$ to $0.097$ in AUROC and from $0.084$ to $0.193$ in AUPRC, indicating that the hidden representations encode operation-relevant information beyond what can be recovered from lexical content alone.
Thus, although surface lexical features carry substantial information about operation identity, they do not fully account for the information captured by hidden representations.

\begin{table}[t]
\centering
\small
\setlength{\tabcolsep}{3pt}
\begin{tabular}{lccc}
\toprule
\textbf{Model}
& \textbf{Position-only}
& \textbf{Text-only}
& \textbf{Hidden(Ours)} \\
\midrule
Qwen3-8B
& $0.718/0.279$
& $0.849/0.549$
& $0.937/0.742$ \\
Qwen2.5-7B
& $0.708/0.269$
& $0.854/0.562$
& $0.895/0.646$ \\
Gemma4-31B
& $0.751/0.311$
& $0.802/0.466$
& $0.899/0.641$ \\
\bottomrule
\end{tabular}

\caption{
Macro AUROC/AUPRC of position-only logistic regression, the stronger
of bag-of-words and TF--IDF logistic regression, and mean-pooled
hidden-state probes. Position and lexical content are informative, but
hidden-state probes achieve higher aggregate performance across all
three models.
}
\label{tab:surface-controls}
\vspace{-1.0em}
\end{table}
% \vspace{-1cm}

\noindent\textbf{Lexically matched comparisons.}
We next directly control for broad lexical similarity between spans.
For each target span, we compare a lexically similar span with a different operation label to a lexically dissimilar span with the same operation label, using the previously trained probe without any additional fitting.
Across all eight reasoning operations, the median effect favors operation identity over broad lexical similarity, with confidence intervals excluding zero for five.
% Across all eight reasoning operations, representations align more strongly with operation identity than with broad lexical similarity.
Full matching procedures and per-operation results are reported in Appendix~\ref{app:lexical-controls}.

\noindent\textbf{Competing-operation vocabulary.}
Broad lexical matching may still leave open the possibility that the probe
relies on a small set of highly operation-specific cue words. We therefore
evaluate held-out spans that contain vocabulary strongly associated with a
\emph{competing} reasoning operation. The original probe remains strongly
predictive on these adversarial subsets, reaching macro AUROC/AUPRC of
$0.919/0.848$ for \texttt{Qwen3-8B} and $0.884/0.789$ for
\texttt{Qwen2.5-7B}. Thus, the presence of lexical cues associated with a
different operation is generally insufficient to override the operation
identity encoded in the hidden representation.

\noindent\textbf{Digit and formula density.}
Finally, we control for numerical and mathematical notation by restricting evaluation to spans with 50--75\% digit or mathematical-token density.
Separability remains strong across the five operation types retained in this subset, with a macro AUROC/AUPRC of $0.917/0.789$ using mean-pooled representations (Table~\ref{tab:density-control}).
All five operations remain above chance, though the margin is smallest for Arithmetic Computation (AUPRC 0.510 vs. 0.204), indicating that digit and formula density alone does not account for the observed operation-level separability.
Full experimental details and per-operation results are provided in Appendix~\ref{app:lexical-controls}.

\subsubsection{Additional Robustness and Generalization}
\label{sec:additional_robustness}

We further test whether operation separability depends on the supervised LDA
projection or relative position within the reasoning trace, and whether it
generalizes beyond the original models and datasets.

\noindent\textbf{Supervised projection.}
Separability persists without supervised LDA. We remove LDA and evaluate the same held-out representations using only PCA and training-set class-mean directions. 
With 128 principal components and mean-pooled representations, macro AUROC/AUPRC remains $0.938/0.716$ for \texttt{Qwen3-8B}, $0.914/0.700$ for \texttt{Qwen2.5-7B}, and $0.872/0.552$ for \texttt{Gemma4-31B}. 
Thus, LDA sharpens the operation-level structure but is not necessary for the central separability result. 
Results across PCA dimensions and span-representation choices (middle-token, mean-pooled representations) are reported in Appendix~\ref{app:pca-only}.

\noindent\textbf{Relative trace position.}
Position alone is predictive of some reasoning operations, but does not account for hidden-state separability. 
Position-only classifiers achieve substantially lower aggregate performance than the hidden-state probes across all three models.
Moreover, when evaluation is restricted to spans occurring within similar relative-position intervals, the original frozen probes remain strongly predictive: across five position bins, macro AUROC ranges from $0.916$ to $0.973$ and macro AUPRC from $0.769$ to $0.898$. 
Full results are reported in Appendix~\ref{app:position-controls}.

\noindent\textbf{Additional models and datasets.}
Operation separability also generalizes beyond the original experimental
settings. Applying the same model-specific probing procedure to
\texttt{Llama-3-8B} yields macro AUROC/AUPRC of $0.958/0.840$ with
mean-pooled representations. In addition, \texttt{Qwen3-8B} probes trained on
the original tasks transfer without task-specific retraining to
GPQA-Diamond ($0.938/0.764$) and MATH-500 ($0.948/0.799$).
Additional results are reported in Appendix~\ref{app:generalization}.

\subsection{Within-Span Structure of Reasoning Operation Signals}
\label{sec:sep_evolve}
Section~\ref{sec:separability} showed that reasoning operations are separable in held-out hidden representations, with separability peaking in middle layers.
What does this signal reflect? It may arise from a small number of token-local cues or from lexical identity. 
We test these alternatives by examining how operation-alignment signals are distributed and evolve across layers.

% \subsubsection{Intra-span variance of reasoning operation-alignment scores}
\subsubsection{From Token-Local Cues to Span-Distributed Signals}
\label{sec:variance}
\noindent\textbf{Experiment.}
We test whether early-layer separability is concentrated on a small number of cue tokens. 
For each token $t$ in an operation span $s_i$, we compute
its alignment with the span's target operation:
$
a_t^{(l,y_i)}
=
\left(z_t^{(l)}\right)^\top d_{y_i}^{(l)}.
$
We then measure the variance of these token-level scores within the span:
$
\mathrm{Var}_i^{(l)}
=
\mathrm{Var}_{t \in s_i}
\left[
a_t^{(l,y_i)}
\right].
$
High intra-span variance indicates that the signal is concentrated on a small
number of tokens, whereas low variance indicates that it is distributed more
uniformly across the operation span.

\noindent\textbf{Result.} We find out that \textit{operation-alignment becomes increasingly distributed across tokens within a reasoning span in middle-layer, while early-layer signals are cue-local}. As shown in Figure~\ref{fig:lda-var-qwen3-8b}, intra-span variance is relatively high in early layers, decreases toward the middle layers, and increases slightly again near the final layers. Thus, the strong separability observed in middle layers is not driven primarily by a small number of isolated cue tokens; instead, operation-aligned information is distributed more broadly across the span. The same qualitative pattern is observed in other models (Appendix~\ref{app:generalization_llm_families}).

% \subsubsection{Identical Tokens Exhibit Different Score Trends Across Reasoning Contexts}
\subsubsection{Identical Surface Tokens Acquire Operation-Dependent Representations}
\label{sec:scatter_surface}
\noindent\textbf{Experiment.}
We next test whether operation-alignment signals can be explained by lexical identity alone. 
For each pair of reasoning operations, we construct a shared-token set by intersecting the ten most frequent token identities from balanced test spans of the two operations. 
For every occurrence of a shared token, we project its hidden representation at each layer into the probe space and compute its alignment with the two corresponding operation vectors.

\noindent\textbf{Result.}
We found that \textit{shared surface-token occurrences increasingly align with their surrounding reasoning operation in middle layers.}
Figure~\ref{fig:common-word-scatter} shows that shared token occurrences are largely intermixed in early layers, but gradually separate along the operation-specific directions in middle-to-late layers. 
The final layer becomes more mixed again. 
Holding surface-token identity fixed, this pattern shows that operation-alignment is not determined by lexical identity alone. 
The shared tokens used for this analysis are listed in Appendix~\ref{app:shared_token_list}.

\begin{figure}[t]
    \centering
    % Local scope for \dataset, \target, and \modelpath
    {
        
        % \begin{subfigure}[b]{0.45\textwidth}
        %     \centering
        %     \rule{8cm}{3cm} 
        %     \caption{Qwen2.5-7B}
        % \end{subfigure}
        
        \begin{subfigure}[b]{0.45\textwidth}
            \centering
            \includegraphics[width=\textwidth]{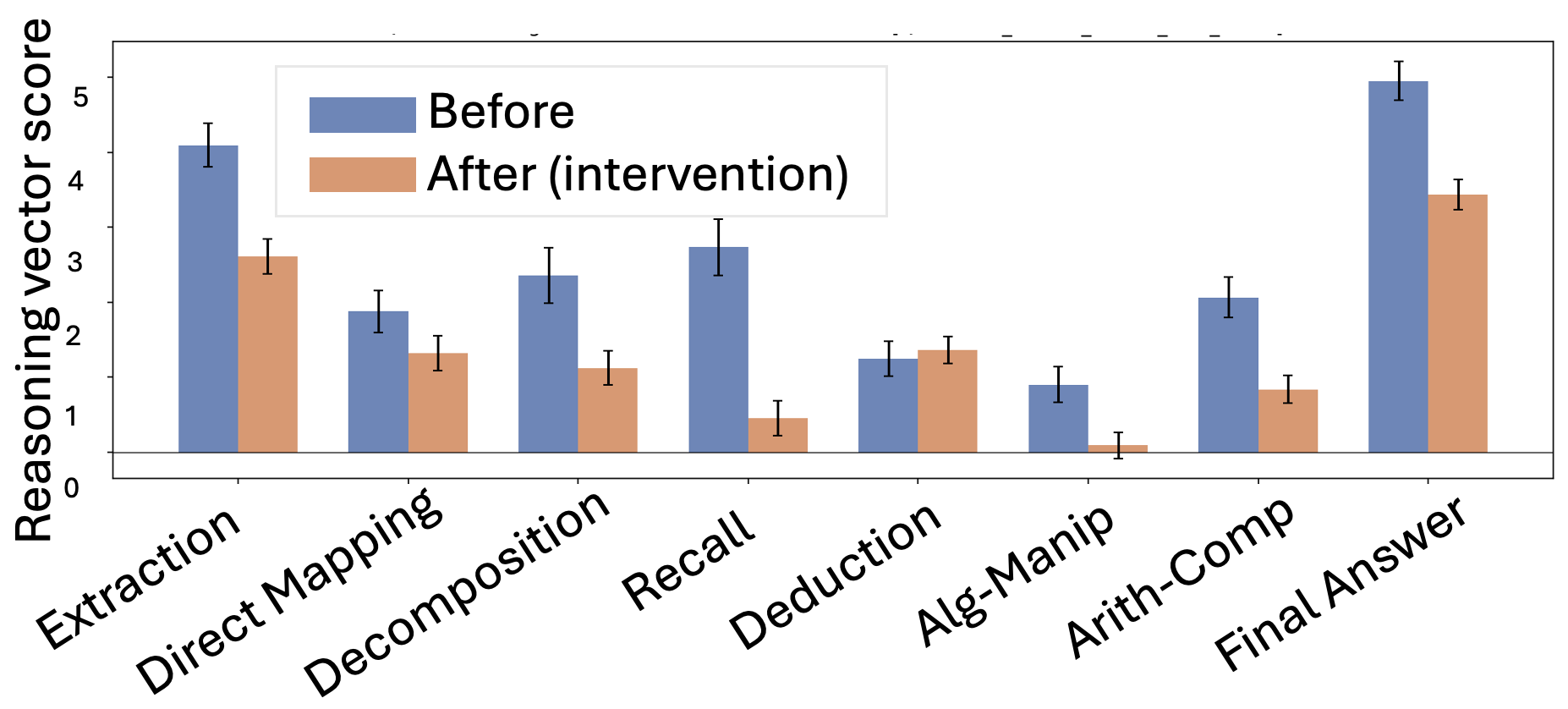}
            % \caption{Qwen3-8b}
        \end{subfigure}
        
    %     \begin{subfigure}[b]{0.45\textwidth}
    %         \centering
    %         \rule{8cm}{3cm} 
    %         \caption{Gemma4-31B}
    %     \end{subfigure}
    }

    \caption{
\textbf{Preceding context shapes operation representations.}
Masking attention to preceding context reduces target reasoning operation-alignment score, indicating that subsequent reasoning operations depend on prior reasoning context.
}
    \label{tab:pretoken_intervention}
    \vspace{-1.0em}
\end{figure}

Representative layer-by-token heatmaps showing the same qualitative pattern are provided in Appendix ~\ref{app:token_heatmap}.
Taken together, operation-alignment signals are not reducible to isolated cue tokens or lexical identity. 
We therefore next examine how preceding reasoning context causally contributes to the formation of these operation representations.

\subsection{Preceding Context Contributes to Operation-Aligned Representations}
\label{sec:sep_trigger}
\noindent\textbf{Experiment.}
We test whether the operation-aligned representation at the onset of a reasoning chunk can be formed independently of its preceding context. 
For each held-out reasoning trace, we re-run the model on the same fixed generated sequence while masking attention from the first token of a target reasoning operation chunk to a selected preceding context region. 
We measure the change in alignment with the target chunk's annotated operation:
$
\Delta_c = a_{\mathrm{after}}(i,c) - a_{\mathrm{before}}(i,c),
$
where $i$ is the first token of the target chunk and $c$ is its operation label. 
A negative $\Delta_c$ indicates that the masked context contributed to the target operation-aligned representation at the chunk onset.

Our primary intervention masks the $M$ tokens immediately preceding the target chunk (\textit{preceding-token} masking). 
As robustness checks, we additionally mask the entire immediately preceding annotated chunk (\textit{preceding-chunk} masking) and use a random between-chunk block masking matched to the preceding-chunk length (\textit{random-chunk} masking). 
We report the primary pre-token results in Figure~\ref{tab:pretoken_intervention}; the additional interventions are reported in Appendix~\ref{app:context_intervention}.

% \subsubsection{Result}

\noindent\textbf{Result.}
% \textit{Preceding reasoning context shapes subsequent operation representations, with the strongest effect from the immediately preceding tokens.}
We found that \textit{immediately preceding context causally contributes to the
operation-aligned representation at the onset of a reasoning chunk.}
Masking the preceding 30 tokens reduces the target reasoning operation-alignment score across reasoning operations. 
Thus, the operation signal at the beginning of a new chunk is not formed solely from the chunk-local token content: it depends
causally on information available in the immediately preceding context.

Masking the entire preceding annotated chunk yields a qualitatively similar direction of change (Appendix~\ref{app:context_intervention}). 
However, the preceding-token masking and preceding-chunk masking interventions differ in masked-span length and eligible examples, so their effect magnitudes should not be directly compared.
The random between-chunk control provides additional context, but is based on a substantially smaller and selectively eligible sample; we therefore interpret it as a supplementary robustness check rather than as the primary basis for the causal claim. 
Overall, these results show that preceding context helps shape the onset of subsequent reasoning-operation representations.

\subsection{Operation Geometry Persists but Weakens under Erroneous Execution}
\label{sec:erroneous_execution}
The preceding analyses characterize how reasoning operations are represented and contextually formed.
We finally ask whether operation-level geometry is specific to successfully executed reasoning steps.
This setting allows us to test whether functional operation identity
remains recoverable when factual execution fails.

\noindent\textbf{Experiment.}
We analyze Qwen3-8B reasoning traces whose final answers are incorrect.
Within these traces, we distinguish spans containing explicit factual errors from operation-matched spans that do not contain factual errors using GPT-5.
Factual errors include incorrect arithmetic, misapplied formulas or theorems, incorrect factual recall, and invalid logical steps.
For each retained operation, we sample equal numbers of factual-error and non-error spans.
We then apply probes trained exclusively on correct reasoning traces to both groups without retraining.
Full annotation criteria, sampling procedures, per-operation results, and statistical tests are provided in Appendix~\ref{app:erroneous_traces}.

\noindent\textbf{Result.} We found that \textit{operation geometry persists under factual errors, but its separability is modestly attenuated.}
Operation identity remains strongly detectable even when an operation
is executed incorrectly.
With mean-pooled representations, probes achieve macro AUROC/AUPRC of $0.955/0.877$ on factual-error spans, compared with $0.971/0.901$ on operation-matched non-error spans from the same incorrect traces.
With middle-token representations, the corresponding scores are $0.920/0.759$ and $0.937/0.808$.
Thus, factual errors do not eliminate operation-level geometry, but reduce its aggregate separability.

In addition, the attenuation is operation-dependent. 
Deduction and Arithmetic Computation show the clearest and most consistent reductions across representation choices and evaluation metrics, whereas Recall shows little difference between factual-error and non-error spans.
These results support a partial dissociation between the functional identity of a reasoning operation and the factual correctness with which it is executed.

\section{Related Work}
\label{sec:rel_work}
\paragraph{Structure and behaviors in reasoning traces.}
Prior work has increasingly treated chain-of-thought reasoning as a structured process rather than a homogeneous sequence of tokens. 
Beyond methods that elicit or organize intermediate reasoning \cite{wei2022chain,yao2022react,yao2023tree,Guo_2025}, recent studies characterize reasoning traces through sentence-level influence, cognitive behaviors, hierarchical episodes, and discourse structure.
Thought Anchors~\cite{bogdan2025thought} identifies reasoning steps that influence subsequent reasoning and final answers; \citet{gandhi2025cognitive} and \citet{zhang2025understanding} analyze reasoning behaviors such as verification, backtracking, and subgoal setting; and ReasoningFlow~\cite{lee2026reasoningflow} represents reasoning traces as discourse graphs to analyze relations among reasoning steps. 
Related work also organizes traces through cognitive taxonomies and hierarchical episodes~\cite{kargupta2025cognitivefoundationsreasoningmanifestation,zhang2026reasoninglenshierarchicalvisualizationdiagnostic,wang2026cognitiveepisodesllmreasoning}.
These studies reveal recurring functional structure in observable reasoning traces, but primarily characterize textual or trajectory-level organization.

\paragraph{Representation and reasoning geometry.}
A complementary line of work studies structured information in LLM hidden representations. 
Prior studies have shown that semantic and behavioral concepts can exhibit linear or otherwise structured geometry in representation space~\cite{park2024the,jiang2024origins,park2025geometry}.
For reasoning models, hidden states have been used to predict answer correctness~\cite{zhang2025reasoning}, while recent work characterizes reasoning through representation trajectories, logical progress, step-specific
geometry, and correctness signals \cite{zhou2026geometry,sun-etal-2026-llm,damirchi-etal-2026-truth}.
Continuous-reasoning approaches further demonstrate that intermediate reasoning can be represented beyond discrete textual tokens \cite{hao2025training,shen-etal-2025-codi,xu-etal-2025-softcot}.
These works establish geometric structure in reasoning representations, but focus primarily on semantic concepts, global trajectories, step identity, or correctness rather than the functional operation performed by a reasoning span.

Our work connects these two lines of research by asking whether recurring functional operations expressed in reasoning text also exhibit distinguishable structure in hidden representation space. 
Unlike trajectory-level behaviors such as backtracking or verification, which may span multiple reasoning steps, we analyze local operation spans that can recur at different positions within a trajectory. 
Thus, our analysis provides an operation-level bridge between the functional structure of observable reasoning traces and their internal representation geometry.

\section{Conclusion}
We investigated whether reasoning operations explicitly distinguished in text are organized as distinct geometric structures in the hidden representations of large language models. 
Through reasoning operation-based analysis, we found that reasoning operations form separable clusters in hidden representation space where even identical tokens acquire operation-specific representations depending on their surrounding reasoning context. 
Causal interventions further revealed that these operations are not locally self-contained but emerge through information propagation from preceding context. 
Consequently, our findings demonstrate that language models maintain a faithful representational correspondence between linguistic reasoning expressions and their internal geometric organization, offering a foundation for interpreting reasoning as a structured, layered, and context-dependent process.

\section{Limitations}
Our study has several limitations.
First, the reasoning-operation annotations are generated by GPT-5 and validated against human annotations on a limited subset of 84 spans. Although the agreement is substantial, it is not perfect, and our human validation focuses on operation labels rather than span boundaries. The annotations should therefore be viewed as approximate labels of textually expressed reasoning functions rather than ground-truth labels of latent cognitive states. Larger-scale validation of both labels and boundaries remains an important direction for future work.

Second, our experiments are limited to mathematical and theorem-driven reasoning tasks and a small set of reasoning-oriented LLMs. While these settings provide structured reasoning traces, the observed geometry may differ in commonsense reasoning, planning, code generation, or interactive tasks, as well as in models with different architectures or training procedures.

Third, our analysis is primarily diagnostic rather than interventional or application-oriented. The learned reasoning-operation vectors reveal representational separability and context dependence, but we do not evaluate whether they can be used to improve model behavior. For example, future work could investigate whether these probes can support reasoning failure detection, verification, decoding-time control, or activation-based steering.

\section{Ethical Considerations}

We used AI assistants during the preparation of this work. ChatGPT was used for writing support, including language polishing and drafting assistance, and Claude Code was used as a coding assistant during implementation. All experimental design, analyses, claims, and final manuscript content were reviewed and verified by the authors.
\section*{Acknowledgements}

This work was supported by the National Research Foundation of Korea(NRF) grant funded by the Korea government(MSIT) (No. RS-2026-25478254).
\paragraph{Generative AI disclosure.} We used generative AI tools during the preparation of this work. GPT-5 was used for reasoning-operation and factual-error annotation as described in the corresponding methodology sections. ChatGPT was used for language polishing and drafting assistance, and Claude Code was used as a coding assistant during implementation. All annotations used in the analyses were subject to the validation procedures described in the paper, and all experimental design, analyses, claims, code, and final manuscript content were reviewed and verified by the authors.

\bibliography{custom}
% \bibliography{custom,anthology-1,anthology-2}

\newpage

\clearpage

\appendix
\section*{Appendix}

\setcounter{table}{0}
\renewcommand{\thetable}{\Alph{table}}
\setcounter{section}{0}
\renewcommand\thesection{\Alph{section}}
\setcounter{figure}{0}
\renewcommand{\thefigure}{\AlphAlph{\value{figure}}}

\begin{itemize}
    \item \S~\ref{app:emergence_dynamics}: Token-Level Emergence and Dynamics of Reasoning-Operation Signals
    \item \S~\ref{app:context_formulation}: Contextual Formation of Reasoning-Operation Representations
    \item \S~\ref{app:generalization}: Generalization Across Models, Tasks, and Erroneous Reasoning
    \item \S~\ref{app:robustness}: Robustness and Alternative Explanations for Representational Separability
    \item \S~\ref{app:full_taxonomy_annotation}: Reasoning-Operation Taxonomy and Annotation Process
    \item \S~\ref{app:setup_details}: Experimental Setup and Implementation Details
\end{itemize}

\section{Token-Level Emergence and Dynamics of Reasoning-Operation Signals}
\label{app:emergence_dynamics}

\subsection{Layer-by-Token Visualization of Reasoning-Operation Signals}
\label{app:token_heatmap}

\begin{figure*}[p]
    \centering

        \begin{subfigure}[b]{0.95\textwidth}
            \centering
            \includegraphics[width=\linewidth, height=0.26\textheight, keepaspectratio]{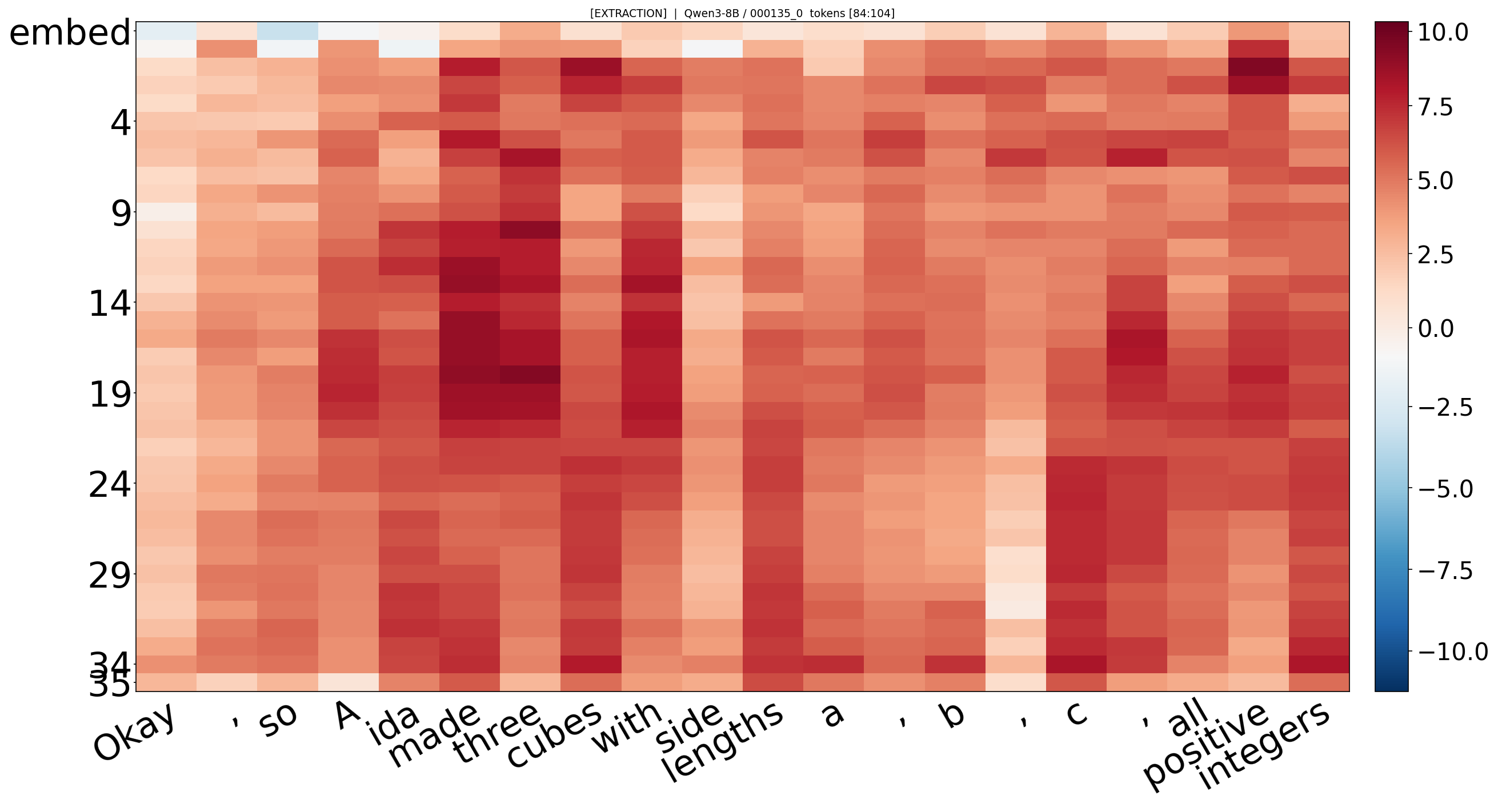}
            \caption{Extraction vector - Example 1}
        \end{subfigure}

        \vspace{0.8em}

        \begin{subfigure}[b]{0.95\textwidth}
            \centering
            \includegraphics[width=\linewidth, height=0.26\textheight, keepaspectratio]{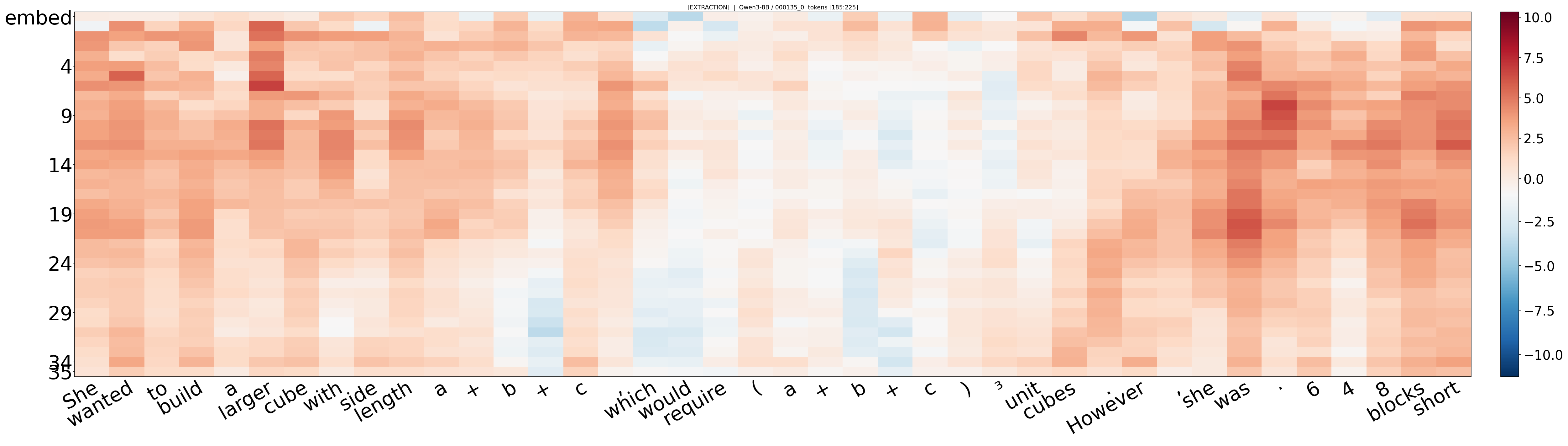}
            \caption{Extraction vector - Example 2}
        \end{subfigure}

        \vspace{0.8em}

        \begin{subfigure}[b]{0.95\textwidth}
            \centering
            \includegraphics[width=\linewidth, height=0.26\textheight, keepaspectratio]{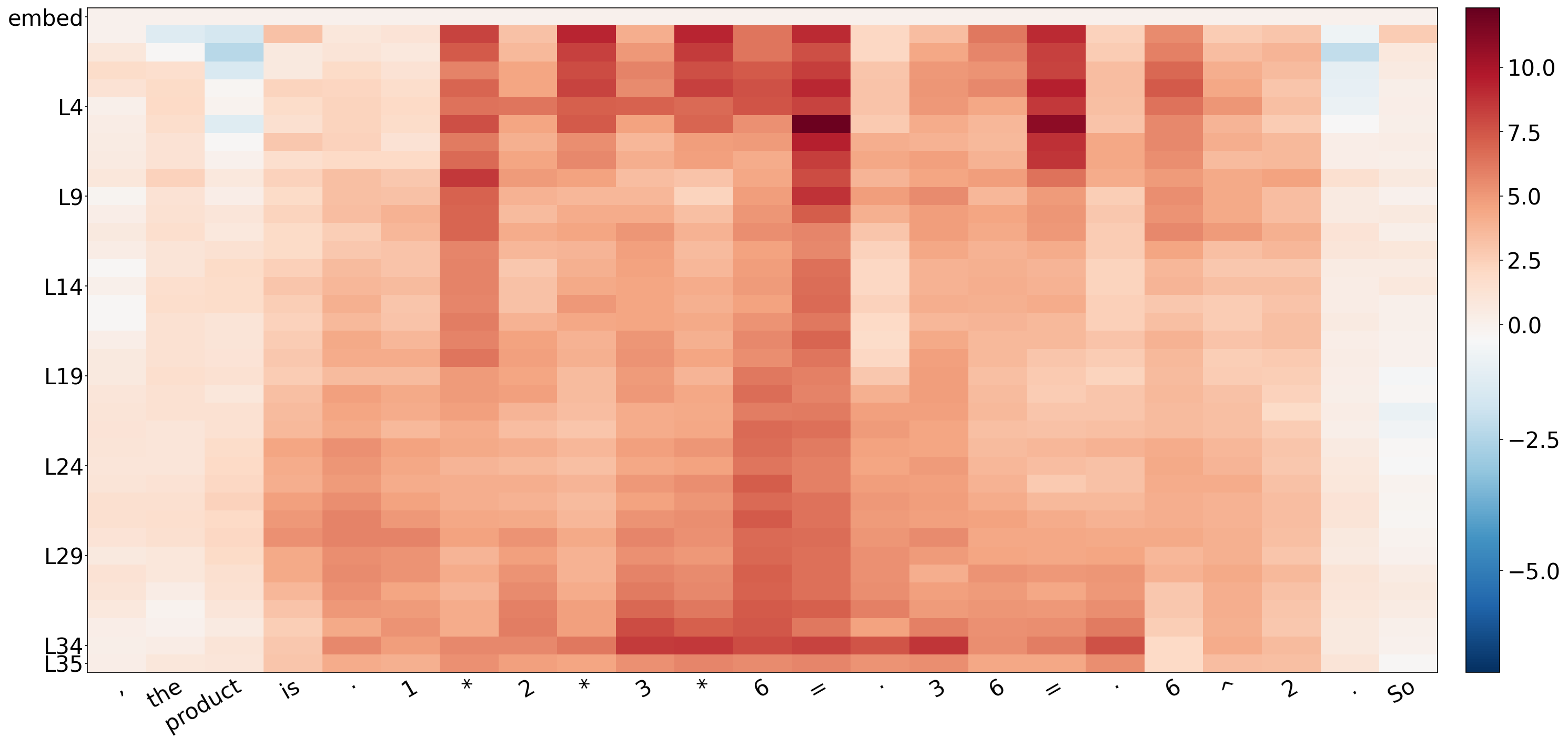}
            \caption{Arithmetic Computation}
        \end{subfigure}

    \caption{
\textbf{Layer-by-token heatmaps of reasoning-operation alignment scores.}
We visualize token-level alignment scores with the corresponding reasoning-operation vector across layers for representative reasoning traces.
Each heatmap shows how strongly each generated token aligns with a target reasoning operation vector at each layer.
The examples illustrate that operation-specific scores are often localized or weak in early layers, become more coherent over contiguous token spans in middle layers, and may become less sharply localized in later layers.
This supports the view that reasoning-operation representations emerge through contextual processing rather than being attached only to isolated cue tokens.
}
\label{fig:token_heatmap_1}
\end{figure*}
\begin{figure*}[p]
    \centering

    \begin{subfigure}[b]{0.95\textwidth}
        \centering
        \includegraphics[width=\linewidth, height=0.20\textheight]{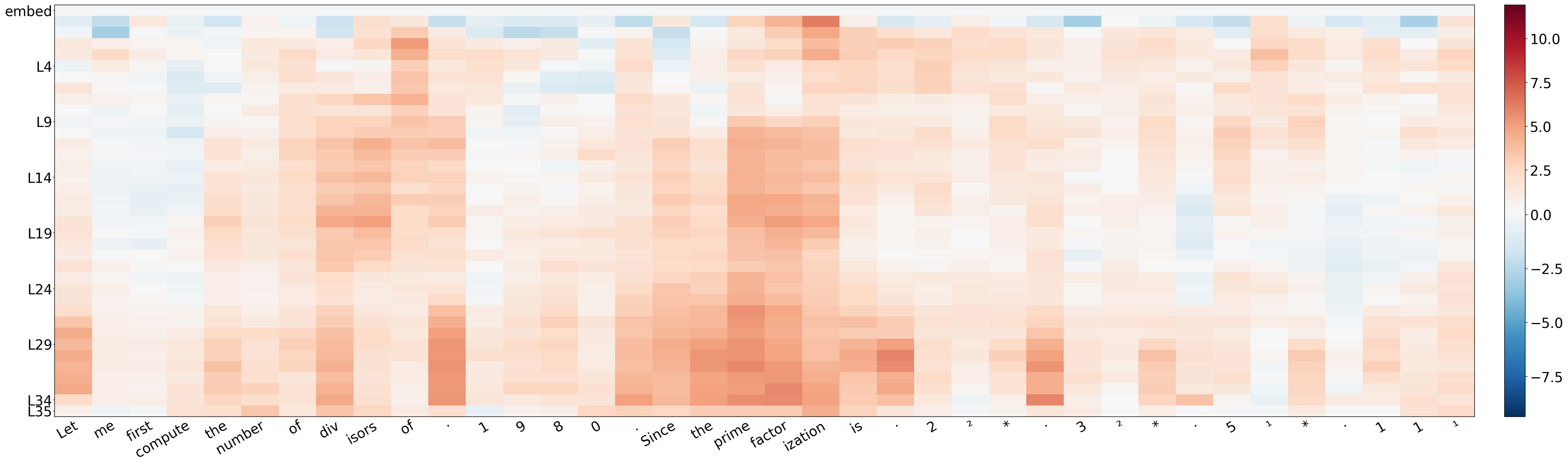}
        \caption{Recall}
    \end{subfigure}
    \vspace{0.8em}
    \begin{subfigure}[b]{0.95\textwidth}
        \centering
        \includegraphics[width=0.7\linewidth, height=0.20\textheight]{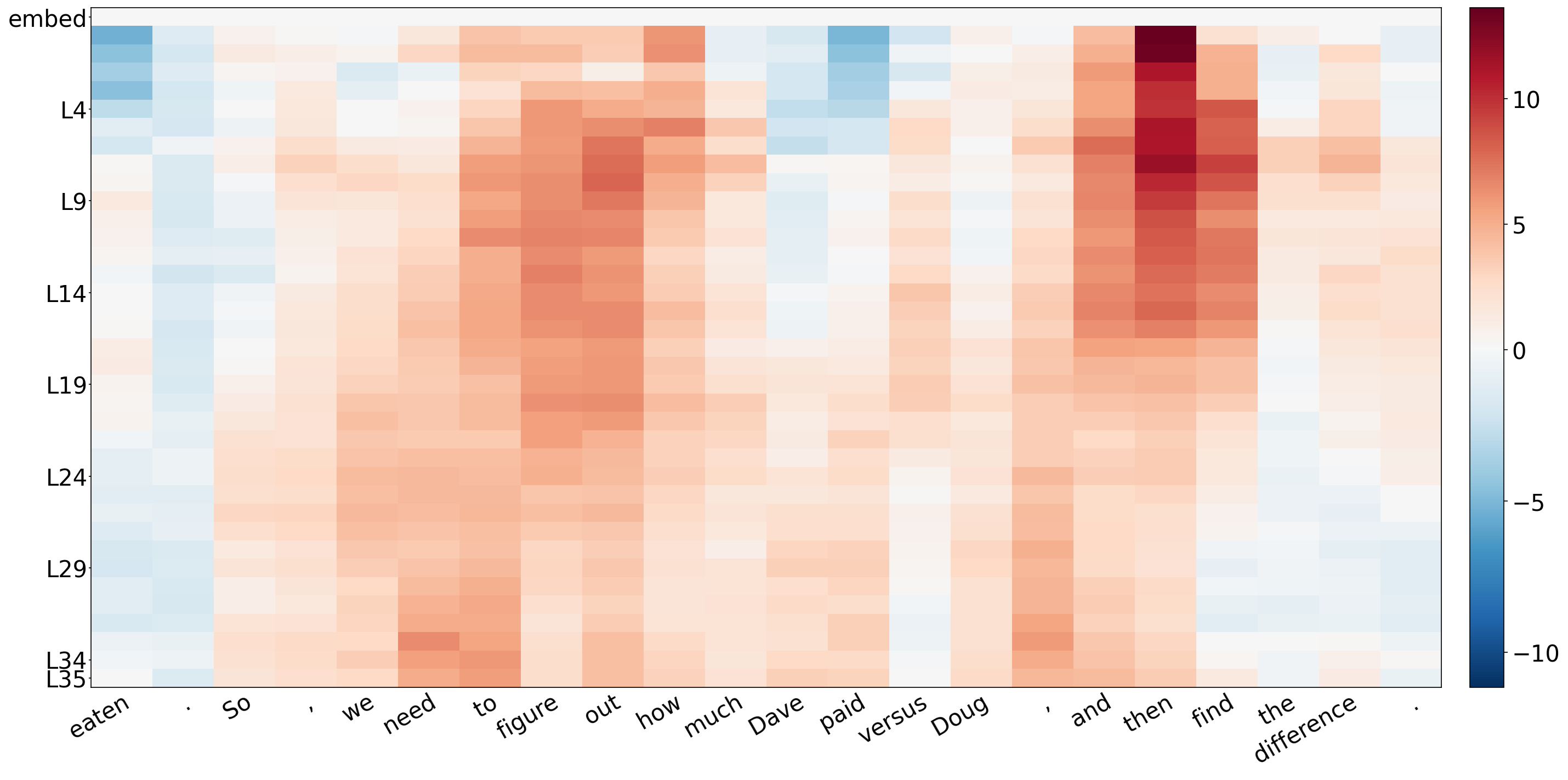}
        \caption{Decomposition}
    \end{subfigure}
    \vspace{0.8em}
    \begin{subfigure}[b]{0.95\textwidth}
        \centering
        \includegraphics[width=\linewidth, height=0.20\textheight]{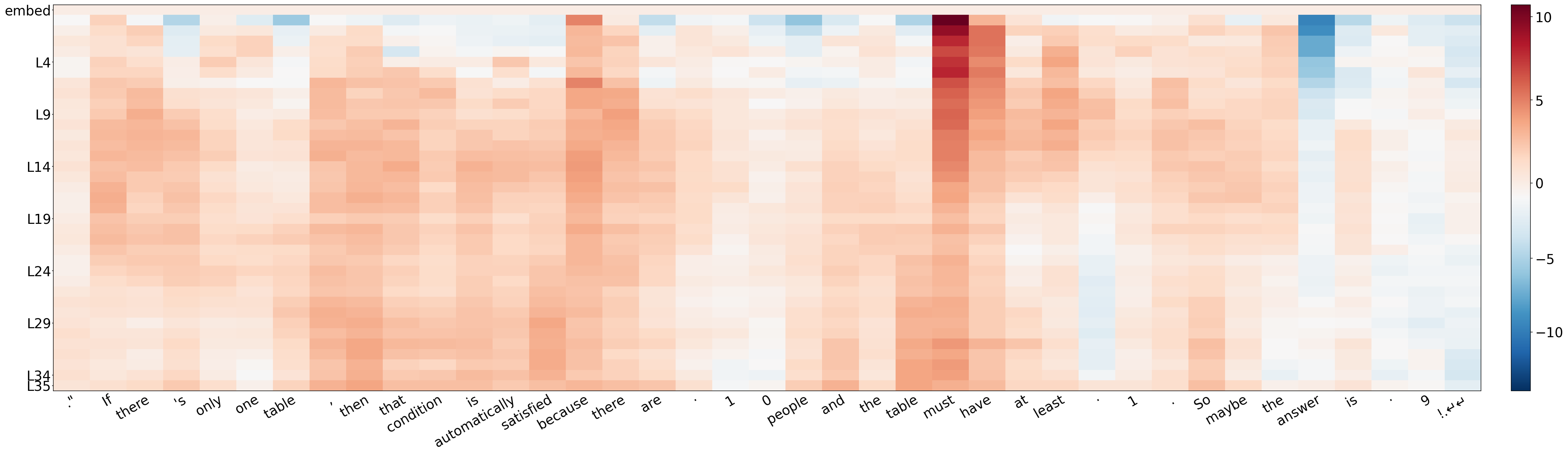}
        \caption{Deduction}
    \end{subfigure}

    \caption{
\textbf{Additional layer-by-token heatmaps of reasoning-operation alignment scores.}
We provide additional representative examples for other reasoning-operation types.
Consistent with Figure~\ref{fig:token_heatmap_1}, the alignment scores become more structured across contiguous generated tokens in middle layers, while early-layer scores are more sparse and token-local.
These qualitative patterns are consistent with the span-level separability and intra-span variance analyses.
}
\label{fig:token_heatmap_2}
\end{figure*}

In addition to span-level analyses, we visualize how reasoning operation-alignment scores evolve over continuously generated tokens in Figure~\ref{fig:token_heatmap_1} and ~\ref{fig:token_heatmap_2}.
For selected reasoning traces, we compute token-level alignment scores with the corresponding reasoning-operation vector at every layer and visualize the resulting layer-by-token score maps.

These heatmaps provide a qualitative view of whether an operation signal is concentrated on a small number of cue tokens or sustained across a broader generated segment.
Across the examples, operation-specific scores are often weak or localized in early layers, become more coherent across contiguous tokens in middle layers, and may become less sharply tied to the annotated operation in later layers.
This pattern is consistent with our span-level separability and intra-span variance analyses, suggesting that reasoning-operation representations emerge through contextual processing rather than being attached only to isolated lexical cues.

The examples also show that identical surface tokens can receive different operation-score intensities depending on their surrounding reasoning context.
For instance, in Figure~\ref{fig:token_heatmap_2}, the comma token appears multiple times in the same trace, but its score intensity differs across occurrences.
This supports the interpretation that token-level reasoning operation-alignment scores are modulated by contextualized reasoning states, rather than determined solely by token identity.

\subsection{Cross-Model Generalization of Span-Distributed Operation Signals}
\label{app:generalization_llm_families}

\begin{figure*}[t]
    \centering
     % Legend
        % Legend
    \begin{subfigure}[b]{\linewidth}
        \centering
        \footnotesize
        \renewcommand{\arraystretch}{0.8}

        \begin{tabular}{
            @{}ll@{\hspace{1.5em}}
            ll@{\hspace{1.5em}}
            ll@{\hspace{1.5em}}
            ll@{}
        }
            \tikz{\draw[line width=0.8mm, color=extraction]
                (0,0) -- (0.35,0);} &
            \texttt{Extraction} &

            \tikz{\draw[line width=0.8mm, color=symbolization]
                (0,0) -- (0.35,0);} &
            \texttt{Direct mapping} &

            \tikz{\draw[line width=0.8mm, color=structural]
                (0,0) -- (0.35,0);} &
            \texttt{Decomposition} &

            \tikz{\draw[line width=0.8mm, color=retrieval]
                (0,0) -- (0.35,0);} &
            \texttt{Recall}
            \\

            \tikz{\draw[line width=0.8mm, color=deduction]
                (0,0) -- (0.35,0);} &
            \texttt{Deduction} &

            \tikz{\draw[line width=0.8mm, color=algebraic]
                (0,0) -- (0.35,0);} &
            \texttt{Algebraic manipulation} &

            \tikz{\draw[line width=0.8mm, color=arithmetic]
                (0,0) -- (0.35,0);} &
            \texttt{Arithmetic computation} &

            \tikz{\draw[line width=0.8mm, color=finalanswer]
                (0,0) -- (0.35,0);} &
            \texttt{Final answer}
        \end{tabular}
    \end{subfigure}

    \vspace{0.5em}

    % Qwen2.5-7B
    \begin{subfigure}[t]{0.49\textwidth}
        \centering
        \includegraphics[width=\linewidth]{
            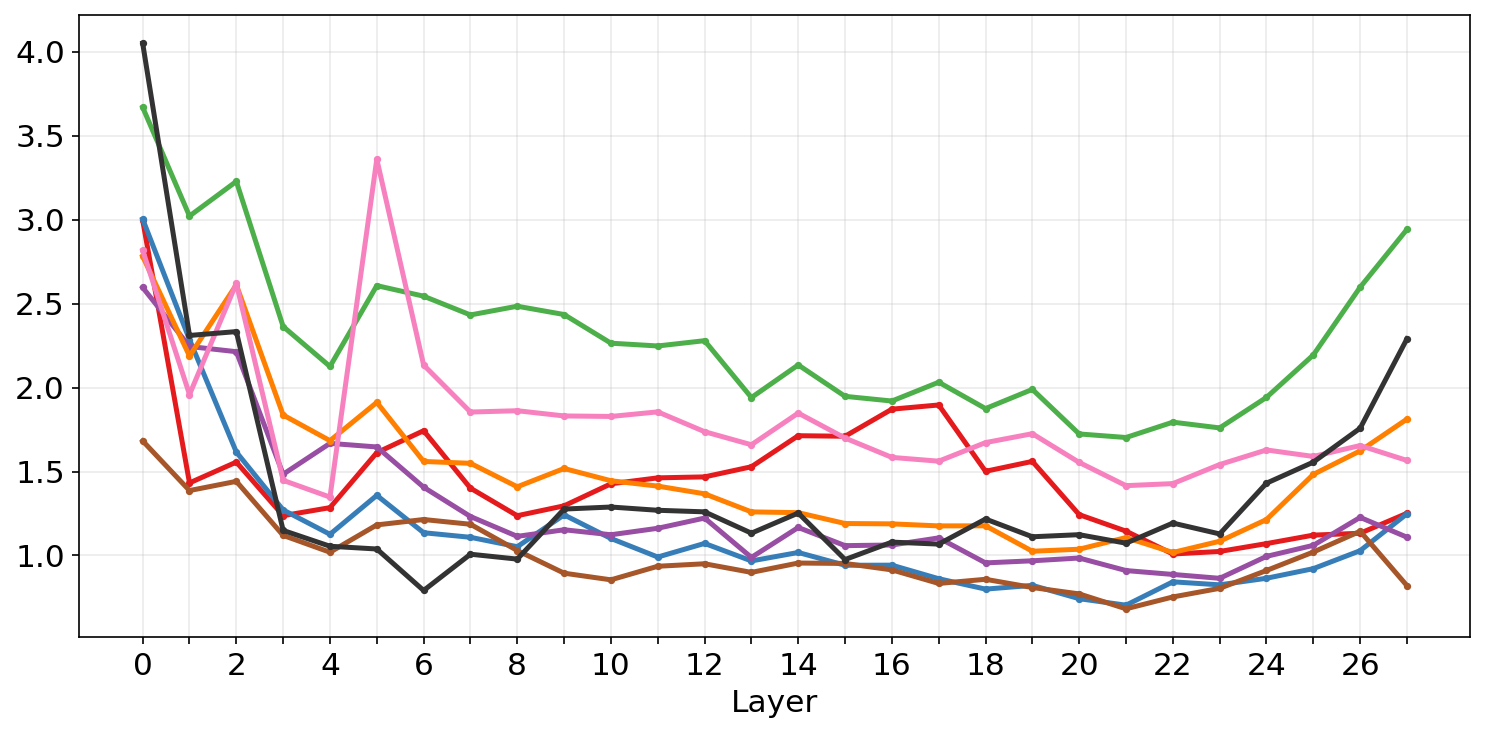
        }
        \caption{Qwen2.5-7B}
        \label{fig:lda-var-qwen25-7b}
    \end{subfigure}
    \hfill
    % Gemma4-31B
    \begin{subfigure}[t]{0.49\textwidth}
        \centering
        \includegraphics[width=\linewidth]{
            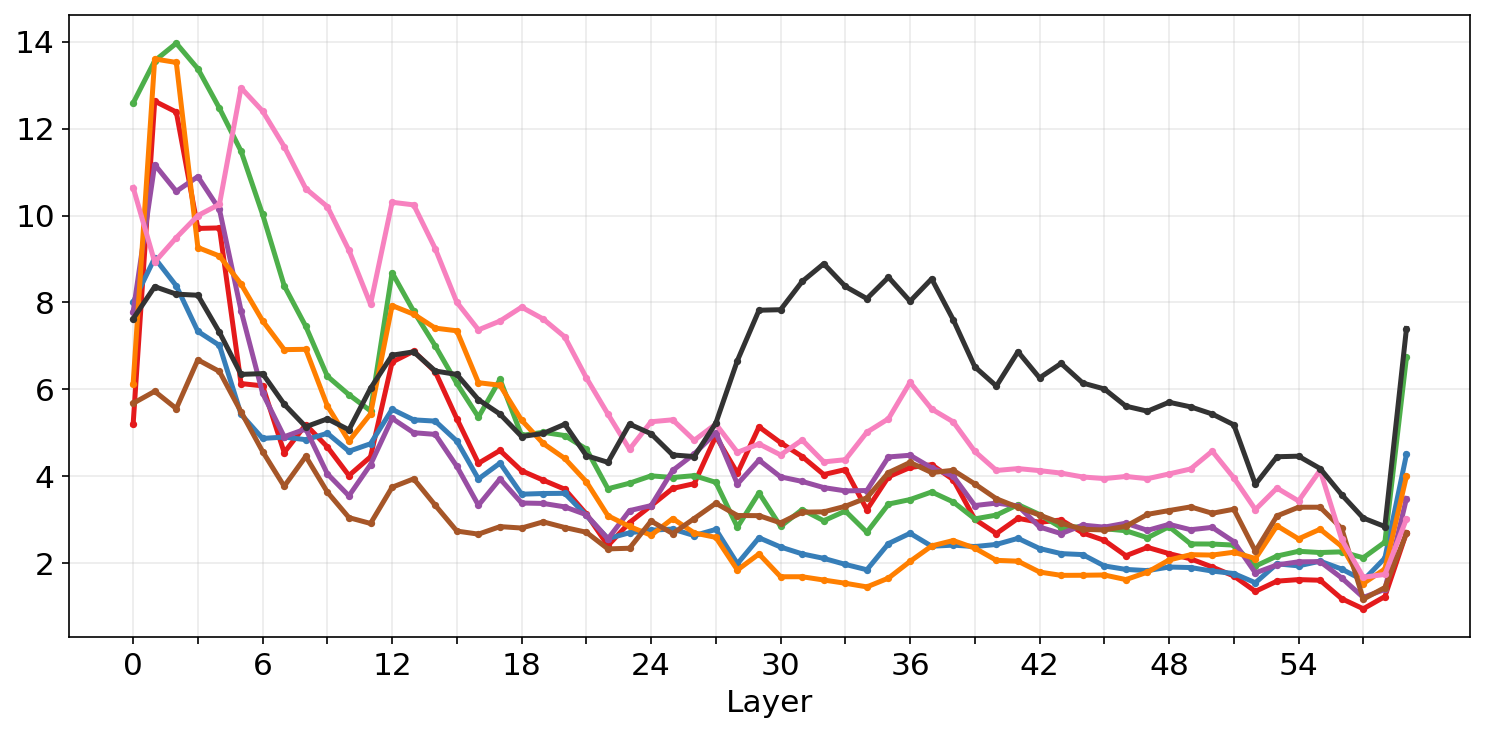
        }
        \caption{Gemma4-31B}
        \label{fig:lda-var-gemma4-31b}
    \end{subfigure}

    \caption{
        \textbf{Quantitative analysis on intra-span LDA score variances
        across additional models.}
        We report the intra-span LDA score variances across layers for
        Qwen2.5-7B and Gemma4-31B. Consistent with Qwen3-8B, variance is high in early layers and decreases toward middle layers, indicating a shift from token-local cues to operation signals distributed across the span.
    }
    \label{fig:lda-var-additional-models}
\end{figure*}

We additionally examine whether the intra-span variance trend observed in \texttt{Qwen3-8B} also appears in other model families.
Figure~\ref{fig:lda-var-additional-models} shows the per-operation variance trends for \texttt{Qwen2.5-7B} and \texttt{Gemma4-31B}, respectively.
Overall, both models exhibit the same qualitative pattern: intra-span variance is relatively high in early layers, decreases toward middle layers, and slightly increases again in later layers.
This indicates that the reduction of intra-span LDA-score variance is a general phenomenon across LLM families rather than a model-specific artifact.

Consistent with the main-text analysis, this trend suggests that early-layer separability is often supported by sparse and token-local cues, whereas middle-layer representations become more distributed across the reasoning operation chunk.
The fact that the same pattern appears across multiple models supports the robustness of our interpretation of reasoning-operation signal formation.

\subsection{Exploratory Temporal Ordering of Early- and Late-Layer Signals}
\label{app:temporal-ordering}

The preceding analyses show that early-layer operation-alignment scores are often concentrated at a small number of token positions, whereas middle-to-late-layer scores tend to be more sustained across contiguous tokens. 
We conduct an exploratory analysis to examine the relative token positions at which sustained operation-alignment signals are first detected in the two layer regimes.

\paragraph{Experiment.}
For each reasoning trace and operation $c$, we construct an early-layer score sequence by averaging the operation-alignment scores over layers $0$--$4$. 
We construct a later-layer score sequence by averaging the five layers with the highest mean held-out AUROC for the corresponding operation.

Let $s_{t}^{c,r}$ denote the target operation-alignment score at token position $t$, where $r \in \{\mathrm{early}, \mathrm{late}\}$ denotes the layer regime.
For each regime, we compute a causal rolling mean with window size
$w=3$:
\begin{equation}
\bar{s}_{t}^{c,r}
=
\frac{1}{w}
\sum_{j=0}^{w-1}
s_{t-j}^{c,r}.
\label{eq:temporal-rolling-mean}
\end{equation}

We define the onset of the signal in regime $r$ as the first token position at which the rolling-mean score exceeds the corresponding high-score threshold:
\begin{equation}
t_{r}^{c}
=
\min
\left\{
t :
\bar{s}_{t}^{c,r}
>
\mu_{r}^{c} + 1.5\sigma_{r}^{c}
\right\},
\label{eq:temporal-onset}
\end{equation}
where $\mu_{r}^{c}$ and $\sigma_{r}^{c}$ are the mean and standard deviation of the score sequence for operation $c$ in regime $r$.

We then calculate the onset difference
\begin{equation}
\delta_{\mathrm{onset}}^{c}
=
t_{\mathrm{late}}^{c}
-
t_{\mathrm{early}}^{c}.
\label{eq:temporal-onset-difference}
\end{equation}
A negative value indicates that the sustained late-layer signal is detected before the sustained early-layer signal under this onset criterion, whereas a positive value indicates the reverse ordering.

For each operation, we report the fraction of examples for which $\delta_{\mathrm{onset}}^{c}<0$, along with the mean and median onset difference. 
We additionally apply a one-sided Wilcoxon signed-rank test to test whether the distribution exhibits a negative location shift.
The reported $p$-values are uncorrected.

\paragraph{Results.}
\begin{table*}[t]
\centering
\small
\begin{tabular}{lrrrr}
\toprule
\textbf{Reasoning operation}
&
\(\boldsymbol{\Pr(\delta_{\mathrm{onset}} < 0)}\)
&
\textbf{Mean \(\boldsymbol{\delta_{\mathrm{onset}}}\)}
&
\textbf{Median \(\boldsymbol{\delta_{\mathrm{onset}}}\)}
&
\textbf{One-sided \(p\)-value}
\\
\midrule
Extraction
& 0.596
& $-3.418$
& $-1.0$
& $7.16 \times 10^{-15}$
\\
Direct Mapping
& 0.391
& $-3.793$
& $0.0$
& $1.90 \times 10^{-6}$
\\
Decomposition
& 0.531
& $-2.344$
& $-1.0$
& $0.0173$
\\
Recall
& 0.406
& $-1.048$
& $0.0$
& $7.11 \times 10^{-6}$
\\
Deduction
& 0.441
& $-2.441$
& $0.0$
& $0.0181$
\\
Algebraic Manipulation
& 0.601
& $-4.017$
& $-1.0$
& $1.23 \times 10^{-5}$
\\
Arithmetic Computation
& 0.770
& $-9.399$
& $-4.0$
& $4.24 \times 10^{-24}$
\\
Final Answer
& 0.875
& $-2.719$
& $-1.0$
& $2.87 \times 10^{-5}$
\\
\bottomrule
\end{tabular}

\caption{
Exploratory onset-to-onset temporal-ordering analysis.
For each operation,
$\delta_{\mathrm{onset}}
=
t_{\mathrm{late}}-t_{\mathrm{early}}$,
where the two onsets are detected using the same causal three-token
rolling-mean procedure and regime-specific thresholds.
Negative values indicate that the sustained late-layer signal is
detected before the sustained early-layer signal under this criterion.
The fraction column reports
$\Pr(\delta_{\mathrm{onset}}<0)$.
The reported $p$-values are from uncorrected one-sided Wilcoxon
signed-rank tests for a negative location shift. A significant
$p$-value does not necessarily indicate that negative onset
differences occur in a majority of examples.
}
\label{tab:onset-ordering}
\end{table*}
As shown in Table~\ref{tab:onset-ordering}, the mean onset difference is negative for all eight operations. 
However, the strength and consistency of this ordering differ substantially across operations.

Arithmetic Computation and Final Answer show the clearest late-before-early tendency: the onset difference is negative for $77.0\%$ and $87.5\%$ of examples, respectively.
Extraction, Decomposition, and Algebraic Manipulation also have negative median onset differences, although their fractions of negative examples are closer to one half.

In contrast, Direct Mapping, Recall, and Deduction have median onset differences of zero, with fewer than half of the examples exhibiting a negative difference.
The significant Wilcoxon results for these operations therefore should not be interpreted as indicating that the late-layer onset occurs first in a majority of examples. 
The signed-rank test can detect a negative location shift when negative differences tend to have greater magnitudes or ranks than positive differences, even if their frequency is below one half.

Overall, the results indicate an operation-dependent negative temporal shift, most clearly for Arithmetic Computation and Final Answer, rather than a consistent onset ordering shared by all operations. 
Moreover, early-layer signals are relatively spike-like and have high token-level variance, whereas later-layer signals tend to be more plateau-like. 
Applying the same three-token averaging and thresholding procedure to these different signal shapes does not guarantee equivalent detection sensitivity across layer regimes.

We therefore interpret this analysis as a supplementary, detector-dependent characterization of the temporal organization of operation-alignment signals. 
It is consistent with earlier detection of sustained later-layer signals for some operations, but does not by itself establish or rule out autoregressive cue propagation.

\subsection{Shared Tokens Used in Pairwise Analysis}
\label{app:shared_token_list}

In Section~\ref{sec:sep_evolve}, we analyze whether identical surface tokens acquire different reasoning operation-alignment scores depending on their surrounding reasoning context.
For each pair of reasoning-operation labels, we first collect the most frequent tokens from balanced test spans of each label and then take their intersection.
The resulting shared-token set is used to compare token representations while controlling for lexical identity.

In the main text, we visualize three representative operation pairs:
\textsc{Deduction} vs. \textsc{Decomposition},
\textsc{Direct Mapping} vs. \textsc{Recall},
and \textsc{Final Answer} vs. \textsc{Recall}.
The shared tokens used for these pairs are listed below:
\[
\begin{aligned}
&\textsc{Deduction} \text{ vs. } \textsc{Decomposition}:\\[-2pt]
&\qquad \{\texttt{a}, \texttt{is}, \texttt{of}, \texttt{so},
\texttt{that}, \texttt{the}, \texttt{to}\},\\[3pt]
&\textsc{Direct Mapping} \text{ vs. } \textsc{Recall}:\\[-2pt]
&\qquad \{\texttt{a}, \texttt{is}, \texttt{let}, \texttt{of},
\texttt{so}, \texttt{that}, \texttt{the}, \texttt{to}\},\\[3pt]
&\textsc{Final Answer} \text{ vs. } \textsc{Recall}:\\[-2pt]
&\qquad \{\texttt{is}, \texttt{of}, \texttt{so}, \texttt{that},
\texttt{the}, \texttt{to}\}.
\end{aligned}
\]
These tokens are mostly function words or common reasoning connectives.
Therefore, any separation observed in the LDA score space is unlikely to be explained by token identity alone.

\section{Contextual Formation of Reasoning-Operation Representations}
\label{app:context_formulation}
\subsection{Additional Context-Intervention Results}
\label{app:context_intervention}
\begin{figure*}[t]
    \centering

    \begin{subfigure}[t]{0.65\textwidth}
        \centering
        \includegraphics[width=\textwidth]{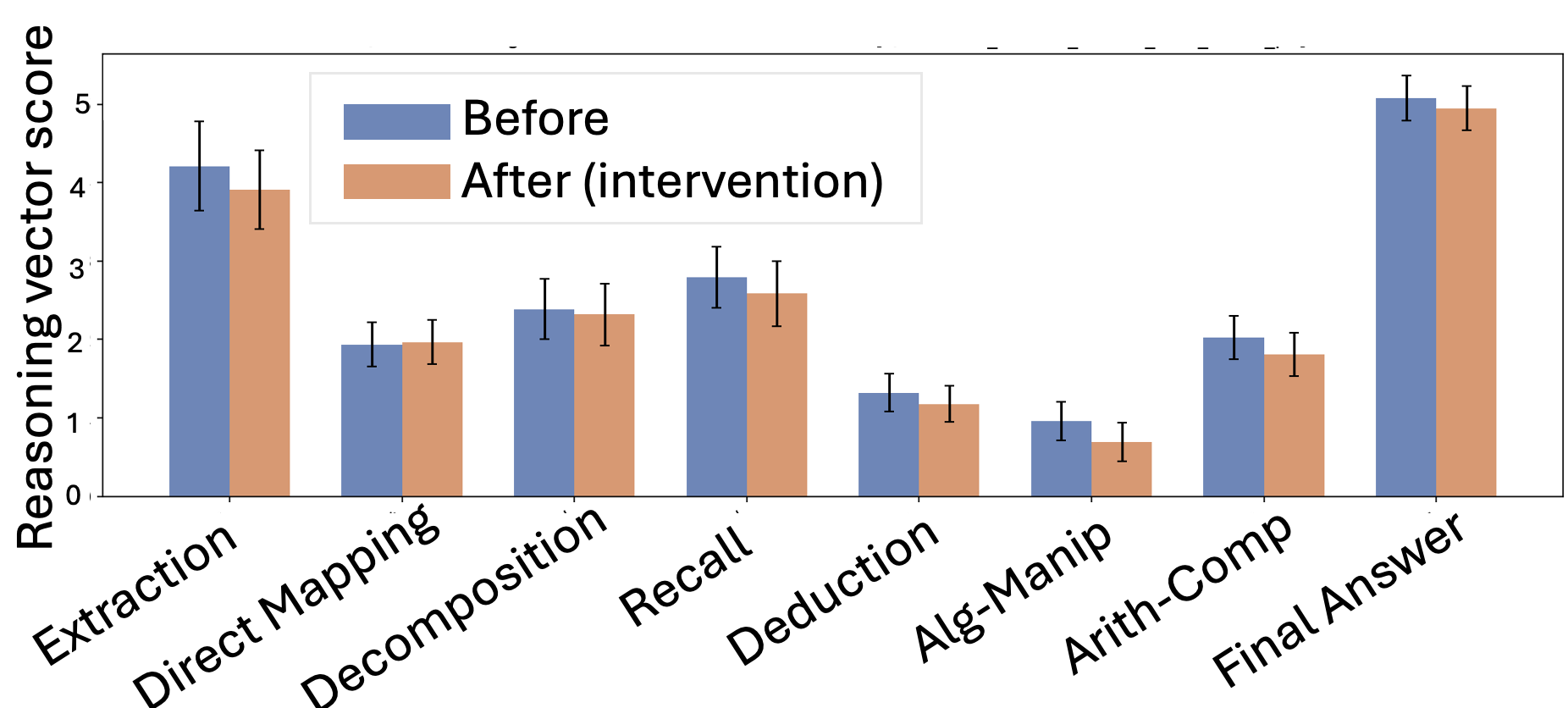}
        \caption{Preceding-chunk masking}
        \label{fig:intervention_prechunk}
    \end{subfigure}

    \vspace{1em}

    \begin{subfigure}[t]{0.65\textwidth}
        \centering
        \includegraphics[width=\textwidth]{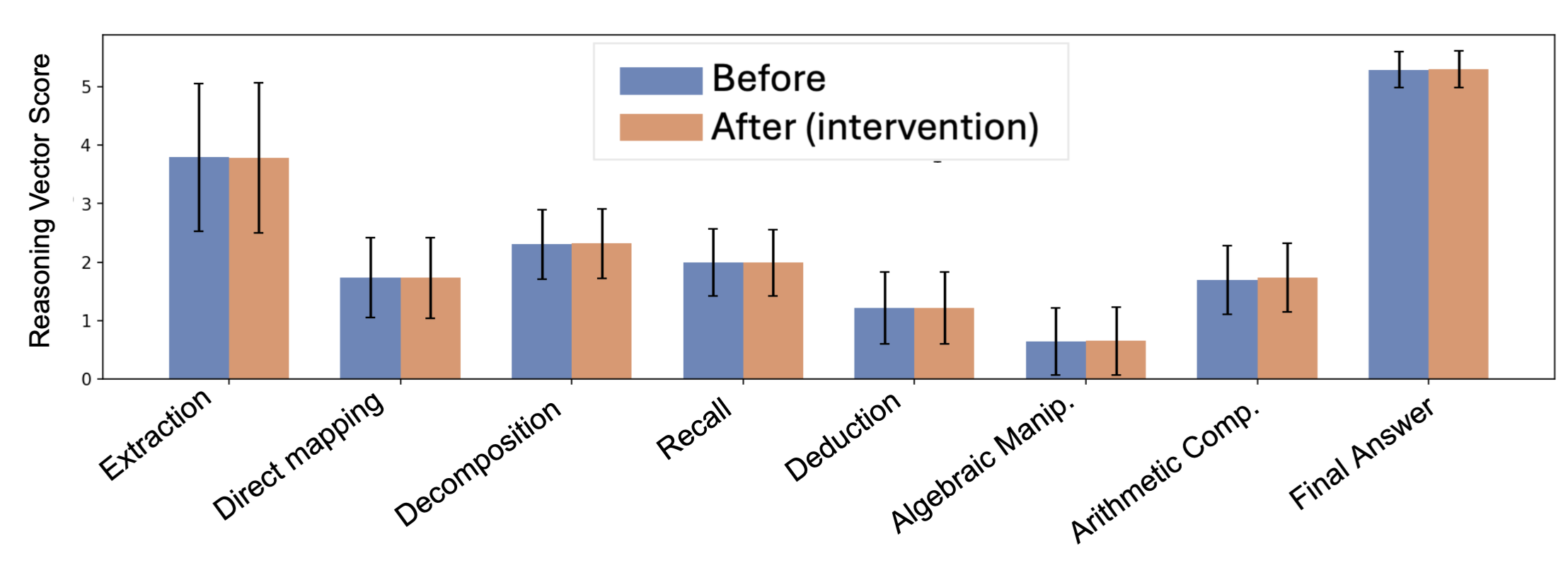}
        \caption{Random preceding-block masking}
        \label{fig:intervention_random}
    \end{subfigure}

    \caption{
    \textbf{Additional attention-masking intervention results.}
    We report context-intervention results beyond the pre-token setting shown in the main text.
    In the pre-chunk setting, we mask attention from the first token of a target reasoning operation chunk to the entire immediately preceding annotated reasoning operation chunk.
    In the random-control setting, we mask a randomly selected preceding block with a comparable length.
    Across settings, blocking preceding context generally reduces the target reasoning operation-alignment score, indicating that subsequent reasoning-operation representations depend on prior context.
    The reduction is stronger for structured preceding-context masking than for the random control, supporting the view that relevant preceding reasoning context contributes to the formation of the next operation representation.
    }
    \label{fig:context_intervention_appendix}
\end{figure*}
As shown in Figure~\ref{fig:context_intervention_appendix}, the additional intervention results show the same ordering as the main pre-token result.
The magnitude of the score decrease is generally smallest under the random preceding-block masking, larger under preceding-chunk masking, and largest under immediate-preceding-token masking:
\[
\text{random} < \text{preceding-chunk} < \text{preceding-token}
\]
This ordering suggests that the formation of a target reasoning-operation representation depends most strongly on the immediately preceding local context, while the previous reasoning operation chunk still contributes meaningful but more diffuse information.
In contrast, randomly selected preceding blocks have a weaker effect, indicating that the score reduction is not simply caused by removing arbitrary past tokens.

One exception is \textsc{Recall}, where the random-control intervention produces a relatively large decrease.
This may indicate that recall operations depend on information accumulated across a broader preceding context, rather than only on the immediately adjacent span or local transition tokens.
Since recall often retrieves formulas, definitions, or theorems that can be cued by earlier problem statements or previously established subgoals, masking even a random preceding block can remove context that remains relevant for forming the recall representation.
Nevertheless, the overall trend across operation types supports the main conclusion: subsequent reasoning-operation representations are context-dependent, and structured preceding context plays a stronger role than arbitrary preceding tokens.

\subsection{Specificity Control with Non-Reasoning Discourse Functions}
\label{app:nonreasoning-control}

We construct a non-reasoning control using five context-dependent discourse functions: \textsc{Question}, \textsc{Response}, \textsc{Assertion}, \textsc{Speculation}, and \textsc{Info}. 
We annotate all 17 chapters of \textit{Harry Potter and the Philosopher's Stone}, retaining up to 5,000 tokens per chapter. 
GPT-5 produces the annotations using the same prompt structure and span format as in the reasoning analysis, changing only the label definitions and examples. Chapters 2, 10, and 12 are held out for testing, and the remaining chapters are used for training. 
Multi-label and duplicate spans are removed, leaving 100 training and 20 test spans per label.

We apply the same mean-pooled residual-stream representation, $L_2$ normalization, PCA, LDA, and one-vs-rest scoring pipeline. 
For a matched comparison, the reasoning probe is evaluated using the same per-label sample budget. 
The literary discourse control reaches a best-layer macro AUROC/AUPRC of $0.665/0.326$, above chance levels of $0.500/0.200$, showing that the control itself is moderately decodable rather than a failed probe.
However, its layer-averaged performance is only $0.551/0.260$, compared with $0.936/0.752$ for reasoning operations. 
The discourse labels also peak at widely distributed layers (2, 3, 11, 22, and 23), whereas six of the eight reasoning operations peak between layers 16 and 22. 
Thus, reasoning operations exhibit stronger and more consistent organization across intermediate layers than the non-reasoning discourse control.

We then apply the same preceding-context intervention. 
For each target span, we prevent its first token from attending to the preceding 30 tokens while
holding the token sequence fixed. 
The literary discourse probe shows no aggregate reduction in target-label score ($\overline{\Delta}=+0.0409$, $p=0.8092$), with a significant decrease only for \textsc{Assertion}. 
In contrast, the reasoning probe shows an aggregate decrease ($\overline{\Delta}=-0.4356$, $p<0.0001$), with significant decreases for Arithmetic Computation, Extraction, Final Answer, Recall, and Direct Mapping.

Because the two domains contain different numbers and distributions of eligible spans, we do not directly compare intervention effect magnitudes or $p$-values across domains. 
Instead, the relevant contrast is the direction and consistency of the within-probe effects: preceding-context masking systematically weakens reasoning-operation representations but not the non-reasoning discourse representations.

\section{Generalization Across Models, Tasks, and Erroneous Reasoning}
\label{app:generalization}

\subsection{Cross-Model Replication on Llama-3-8B}
\begin{table*}[t]
\centering
\small
\begin{tabular}{lcccc}
\toprule
\textbf{Operation}
& \textbf{AUROC-M}
& \textbf{AUROC-Mid}
& \textbf{AUPRC-M}
& \textbf{AUPRC-Mid} \\
\midrule
Extraction             & 0.959 & 0.903 & 0.880 & 0.711 \\
Direct Mapping         & 0.936 & 0.890 & 0.821 & 0.721 \\
Decomposition          & 0.987 & 0.943 & 0.852 & 0.659 \\
Recall                 & 0.974 & 0.918 & 0.857 & 0.785 \\
Deduction              & 0.869 & 0.766 & 0.652 & 0.470 \\
Algebraic Manipulation & 0.985 & 0.926 & 0.907 & 0.739 \\
Arithmetic Computation & 0.970 & 0.958 & 0.812 & 0.844 \\
Final Answer           & 0.984 & 0.979 & 0.935 & 0.919 \\
\midrule
Macro                  & 0.958 & 0.910 & 0.840 & 0.731 \\
\bottomrule
\end{tabular}
\caption{
Llama-3-8B replication. M and Mid denote mean-pooled and middle-token
span representations.
}
\label{tab:llama-generalization}
\end{table*}

We apply the same model-specific probing procedure to Llama-3-8B~\cite{grattafiori2024llama3herdmodels}, using up to 100 training and 20 test examples per operation. 
PCA, LDA, and operation directions are fitted on Llama-3-8B training representations and evaluated on its held-out test spans. 
Results are averaged over layers 10--28 and is in Table~\ref{tab:llama-generalization}.

\subsection{Cross-Task Transfer to GPQA-Diamond and MATH-500}
\begin{table*}[t]
\centering
\scriptsize
\setlength{\tabcolsep}{3pt}
\begin{tabular}{@{}lcccccccc@{}}
\toprule
& \multicolumn{4}{c}{\textbf{GPQA-Diamond}}
& \multicolumn{4}{c}{\textbf{MATH-500}} \\
\cmidrule(lr){2-5}\cmidrule(lr){6-9}
\textbf{Operation}
& \textbf{AUROC-M}
& \textbf{AUROC-Mid}
& \textbf{AUPRC-M}
& \textbf{AUPRC-Mid}
& \textbf{AUROC-M}
& \textbf{AUROC-Mid}
& \textbf{AUPRC-M}
& \textbf{AUPRC-Mid} \\
\midrule
Extraction
& 0.983 & 0.967 & 0.925 & 0.861
& 0.986 & 0.983 & 0.983 & 0.960 \\
Direct Mapping
& 0.907 & 0.835 & 0.699 & 0.506
& 0.902 & 0.837 & 0.732 & 0.455 \\
Decomposition
& 0.951 & 0.871 & 0.859 & 0.650
& 0.970 & 0.881 & 0.862 & 0.615 \\
Recall
& 0.903 & 0.844 & 0.675 & 0.538
& 0.954 & 0.902 & 0.845 & 0.707 \\
Deduction
& 0.902 & 0.846 & 0.585 & 0.427
& 0.921 & 0.850 & 0.646 & 0.518 \\
Algebraic Manipulation
& 0.947 & 0.891 & 0.792 & 0.592
& 0.937 & 0.897 & 0.724 & 0.573 \\
Arithmetic Computation
& 0.941 & 0.897 & 0.708 & 0.618
& 0.924 & 0.902 & 0.685 & 0.593 \\
Final Answer
& 0.974 & 0.950 & 0.872 & 0.785
& 0.987 & 0.977 & 0.914 & 0.868 \\
\midrule
Macro
& 0.938 & 0.888 & 0.764 & 0.622
& 0.948 & 0.904 & 0.799 & 0.661 \\
\bottomrule
\end{tabular}
\caption{
Test-only transfer of Qwen3-8B probes to GPQA-Diamond and MATH-500
without task-specific probe retraining.
}
\label{tab:task-transfer}
\end{table*}

We apply Qwen3-8B probes trained on the original tasks to annotated GPQA-Diamond~\cite{rein2024gpqa} and MATH-500~\cite{lightman2024let} spans without task-specific probe retraining. 
Results that are averaged over layers 10--30 are reported in Table~\ref{tab:task-transfer}.

\subsection{Operation Geometry under Factual Errors}
\label{app:erroneous_traces}

We define an incorrect trace as a complete Qwen3-8B reasoning trajectory that ends with an incorrect final answer.
GPT-5 segments each trace using the original reasoning-operation taxonomy and identifies spans containing explicit factual errors, including incorrect arithmetic, misapplied formulas or theorems, incorrect factual recall, and invalid logical steps.
Inefficient strategies and vague expressions are not classified as factual errors.
The authors manually screened a subset of the resulting segmentations and error annotations.

We divide operation spans into factual-error spans and non-error spans from the same incorrect traces.
We retain operations with at least 15 factual-error spans, sample at most 30 factual-error spans per operation, and sample an equal number of non-error spans with the same operation label, using replacement only when necessary.

PCA--LDA probes trained exclusively on correct reasoning traces are applied to both groups without retraining.
We report one-vs-rest AUROC and AUPRC averaged across layers 10--30 using middle-token and mean-pooled span representations.
For each operation and metric, we perform a one-sided paired Wilcoxon signed-rank test of whether non-error separability exceeds factual-error separability.
Full results are reported in Table~\ref{tab:erroneous_operation_full}.

\begin{table*}[t]
\centering
\small
\setlength{\tabcolsep}{4.7pt}
\begin{tabular}{lcccccc}
\toprule
Operation
& Error AUROC
& Non-error AUROC
& $p_{\mathrm{AUROC}}$
& Error AUPRC
& Non-error AUPRC
& $p_{\mathrm{AUPRC}}$ \\
\midrule

\multicolumn{7}{l}{\textit{(a) Middle-token representation}} \\

Recall
& 0.946 & 0.949 & 0.446
& 0.862 & 0.857 & 0.695 \\

Deduction
& 0.867 & 0.906 & $<0.001$
& 0.592 & 0.755 & $<0.001$ \\

Algebraic Manipulation
& 0.897 & 0.915 & $<0.001$
& 0.632 & 0.630 & 0.607 \\

Arithmetic Computation
& 0.926 & 0.957 & $<0.001$
& 0.784 & 0.866 & $<0.001$ \\

Final Answer
& 0.964 & 0.960 & 0.773
& 0.924 & 0.932 & 0.406 \\

\midrule
\textbf{Macro}
& \textbf{0.920} & \textbf{0.937} & \textbf{$<0.001$}
& \textbf{0.759} & \textbf{0.808} & \textbf{$<0.001$} \\

\midrule
\multicolumn{7}{l}{\textit{(b) Mean-pooled representation}} \\

Recall
& 0.978 & 0.979 & 0.196
& 0.957 & 0.943 & 0.987 \\

Deduction
& 0.927 & 0.950 & $<0.001$
& 0.768 & 0.847 & $<0.001$ \\

Algebraic Manipulation
& 0.935 & 0.952 & $<0.001$
& 0.790 & 0.805 & 0.041 \\

Arithmetic Computation
& 0.980 & 0.989 & $<0.001$
& 0.934 & 0.953 & 0.002 \\

Final Answer
& 0.955 & 0.985 & $<0.001$
& 0.937 & 0.958 & $<0.001$ \\

\midrule
\textbf{Macro}
& \textbf{0.955} & \textbf{0.971} & \textbf{$<0.001$}
& \textbf{0.877} & \textbf{0.901} & \textbf{$<0.001$} \\

\bottomrule
\end{tabular}

\caption{
Transfer of reasoning-operation probes trained exclusively on correct reasoning traces to factual-error and operation-matched non-error spans from incorrect Qwen3-8B reasoning traces.
Results are averaged across layers 10--30.
The error and non-error groups contain equal numbers of spans for each
operation. 
Reported $p$-values are from one-sided paired Wilcoxon signed-rank tests of whether separability is higher for non-error spans than for factual-error spans.
Values smaller than $0.001$ are reported as $<0.001$.
}
\label{tab:erroneous_operation_full}
\end{table*}

Operation identity remains strongly detectable in factual-error spans.
The macro AUROC/AUPRC is $0.920/0.759$ with middle-token representations and $0.955/0.877$ with mean-pooled representations.
However, aggregate separability is significantly lower than for operation-matched non-error spans under both representation choices.

The effect varies across operations.
Deduction and Arithmetic Computation show significant reductions in both AUROC and AUPRC for both representation choices.
Recall shows no significant reduction under either representation.
Algebraic Manipulation shows a significant AUROC reduction for both representations, but its AUPRC difference is small and inconsistent.
Final Answer shows a significant reduction with mean pooling, whereas the middle-token results do not show a consistent reduction.
These results indicate that factual errors modestly attenuate, rather than eliminate, operation-level geometry.

\section{Robustness and Alternative Explanations for Representational Separability}
\label{app:robustness}

\subsection{Robustness to Projection Choice: PCA-Only Evaluation}
\label{app:pca-only}
\begin{table*}[t]
\centering
\small
\begin{tabular}{lcccccc}
\toprule
& \multicolumn{3}{c}{\textbf{Mean pooling}}
& \multicolumn{3}{c}{\textbf{Middle token}} \\
\cmidrule(lr){2-4}\cmidrule(lr){5-7}
\textbf{Model}
& \textbf{8 PCs}
& \textbf{32 PCs}
& \textbf{128 PCs}
& \textbf{8 PCs}
& \textbf{32 PCs}
& \textbf{128 PCs} \\
\midrule
Qwen3-8B
& $0.908/0.605$
& $0.934/0.699$
& $0.938/0.716$
& $0.807/0.434$
& $0.861/0.549$
& $0.873/0.583$ \\
Qwen2.5-7B
& $0.886/0.622$
& $0.909/0.683$
& $0.914/0.700$
& $0.794/0.439$
& $0.849/0.563$
& $0.862/0.601$ \\
Gemma4-31B
& $0.846/0.479$
& $0.867/0.538$
& $0.872/0.552$
& $0.744/0.336$
& $0.783/0.399$
& $0.793/0.424$ \\
\bottomrule
\end{tabular}
\caption{
Macro AUROC/AUPRC of PCA-only evaluation. Operation separability
remains above chance without supervised LDA across all models,
component counts, and span representations.
}
\label{tab:pca-only-summary}
\end{table*}

The main probe uses supervised LDA after PCA. 
To test whether the reported separability is induced by the supervised projection, we repeat the held-out analysis without LDA. 
We preserve the original question-level split, class balancing, span-selection procedure, and hidden representations. 
PCA is fitted only on the training representations and applied without refitting to the test set.

We evaluate 8, 32, and 128 principal components in Table~\ref{tab:pca-only-summary}. 
In the resulting PCA space, we construct the same one-vs-rest class-centroid directions using the training examples and evaluate their scores on the held-out test spans. 
Results are averaged over the selected intermediate-layer ranges for each model, which are: layers 10–30 for Qwen3-8B, 8–23 for Qwen2.5-7B, and 13–40 for Gemma4-31B.

Operation labels remain substantially separable in the PCA-only space.
With mean pooling, macro AUROC ranges from $0.846$ to $0.908$ using 8 components and from $0.872$ to $0.938$ using 128 components. 
The largest gains generally occur between 8 and 32 components, with smaller improvements thereafter. 
LDA therefore provides a compact supervised coordinate system that sharpens the structure, but the central held-out separability result does not depend on LDA.

\subsection{Robustness to Span-Representation Choice}
\label{app:position_variants}

In the main text, we report separability results using the middle-token representation of each reasoning operation chunk.
Here, we provide additional results using mean-pooled, first-token, and last-token span representations.
Overall, the results are qualitatively consistent with the middle-token setting: reasoning-operation labels remain substantially more separable than the random-label baseline across models and layers, indicating that the observed operation-level geometry is not specific to a single choice of span position.

\begin{figure*}[t]
    \centering
    
    % \def\mode{res}
    % \def\position{middle}
    
    % legend
    {
    \begin{subfigure}[b]{\textwidth}
    \centering
    \footnotesize
    \renewcommand{\arraystretch}{0.7}
    \begin{tabular}{@{}ll@{\hspace{0.9em}}ll@{\hspace{0.9em}}ll@{\hspace{0.9em}}ll@{}}
        \tikz{\draw[line width=0.8mm, color=extraction] (0,0) -- (0.35,0);} &
        \texttt{Extraction} &
        \tikz{\draw[line width=0.8mm, color=symbolization] (0,0) -- (0.35,0);} &
        \texttt{Direct mapping} &
        \tikz{\draw[line width=0.8mm, color=structural] (0,0) -- (0.35,0);} &
        \texttt{Decomposition} &
        \tikz{\draw[line width=0.8mm, color=retrieval] (0,0) -- (0.35,0);} &
        \texttt{Recall} \\
    
        \tikz{\draw[line width=0.8mm, color=deduction] (0,0) -- (0.35,0);} &
        \texttt{Deduction} &
        \tikz{\draw[line width=0.8mm, color=algebraic] (0,0) -- (0.35,0);} &
        \texttt{Algebraic manipulation} &
        \tikz{\draw[line width=0.8mm, color=arithmetic] (0,0) -- (0.35,0);} &
        \texttt{Arithmetic computation} &
        \tikz{\draw[line width=0.8mm, color=finalanswer] (0,0) -- (0.35,0);} &
        \texttt{Final answer}
    \end{tabular}
    \end{subfigure}
    }
    \vspace{-0.5em}

    \begin{subfigure}[b]{0.48\textwidth}
        \centering
        \includegraphics[width=\textwidth,height=2.5cm]{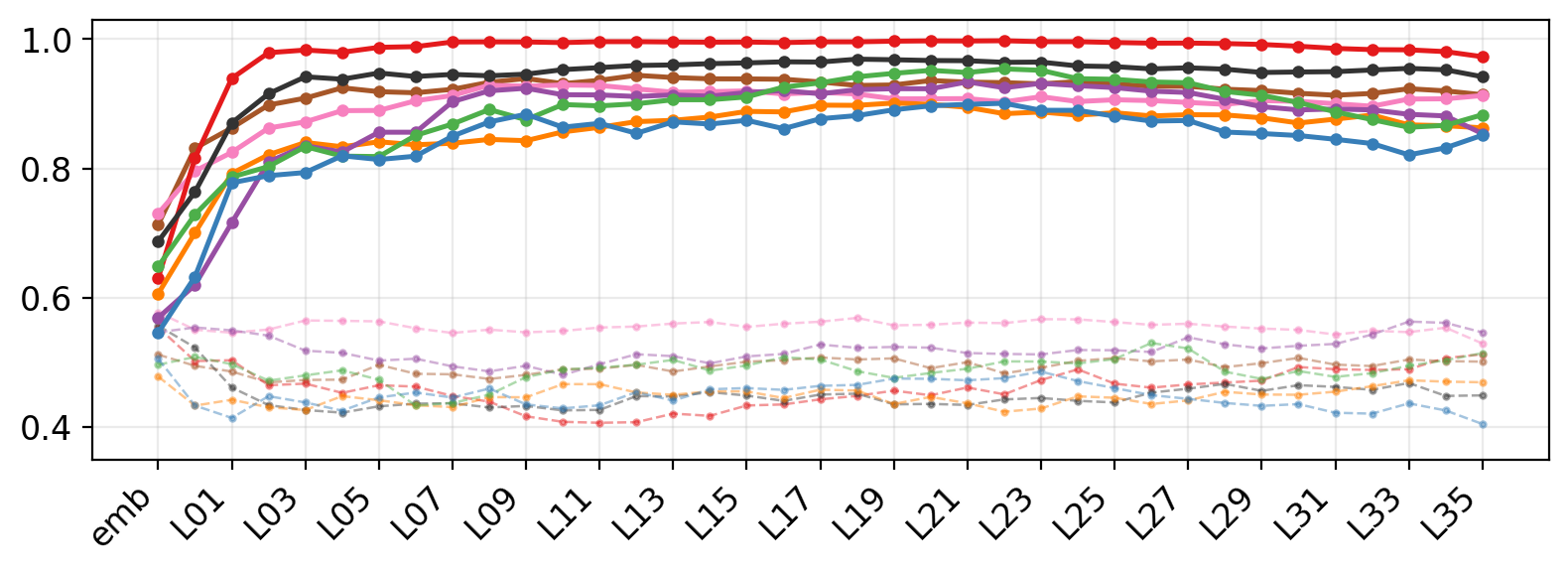}
        \caption{AUROC, Qwen3-8B}
    \end{subfigure} 
    \hfill
    \begin{subfigure}[b]{0.48\textwidth}
        \centering
        \includegraphics[width=\textwidth,height=2.5cm]{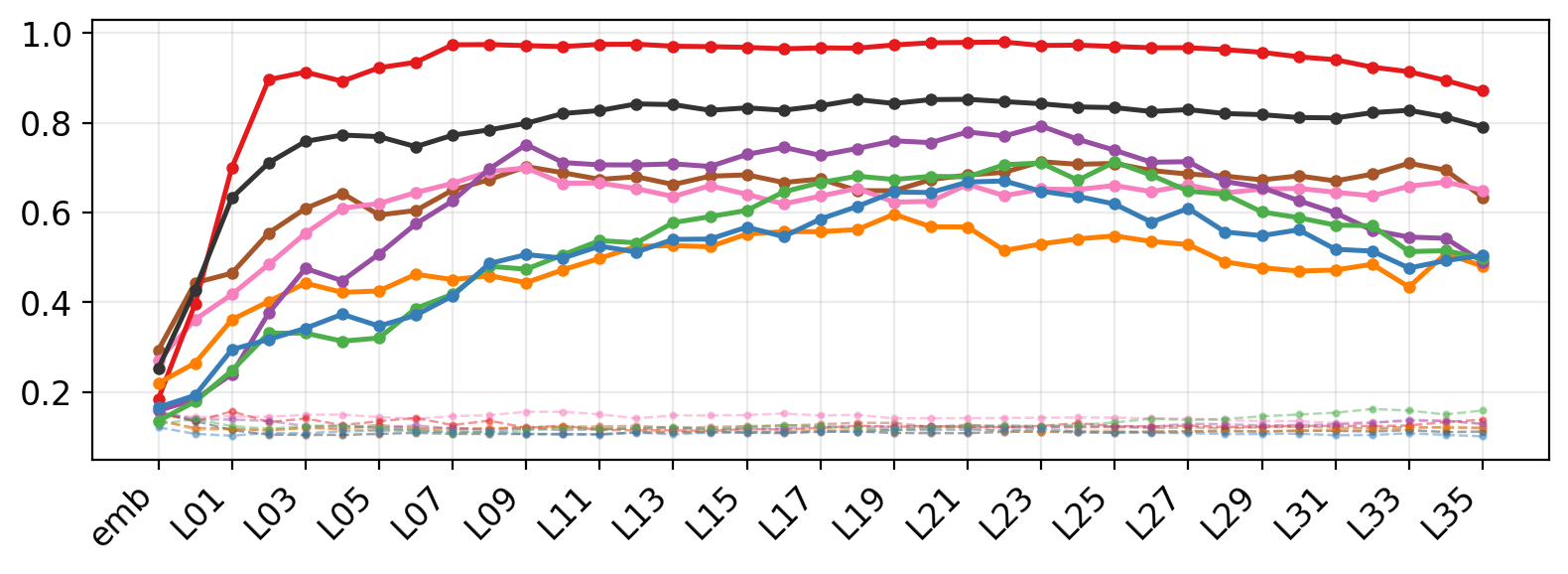}
        \caption{AUPRC, Qwen3-8B}
    \end{subfigure}

    \begin{subfigure}[b]{0.48\textwidth}
        \centering
        \includegraphics[width=\textwidth,height=2.5cm]{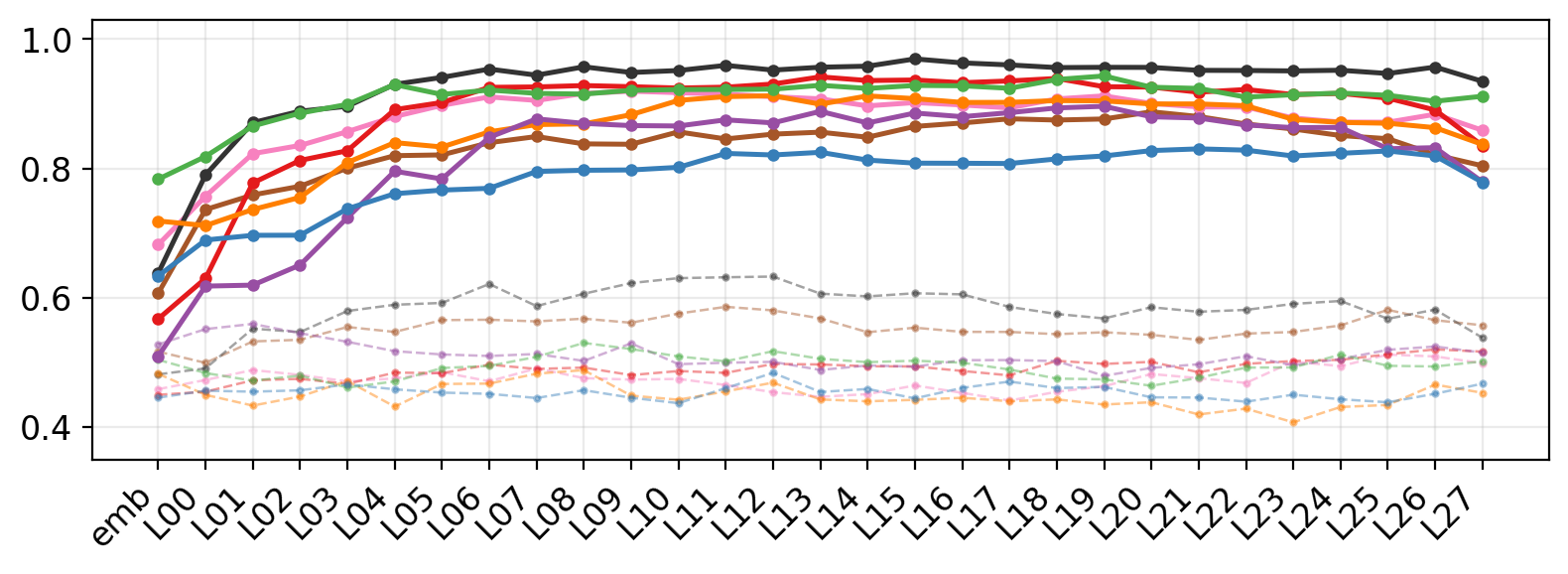}
        \caption{AUROC, Qwen2.5-7B}
    \end{subfigure} 
    \hfill
    \begin{subfigure}[b]{0.48\textwidth}
        \centering
        \includegraphics[width=\textwidth,height=2.5cm]{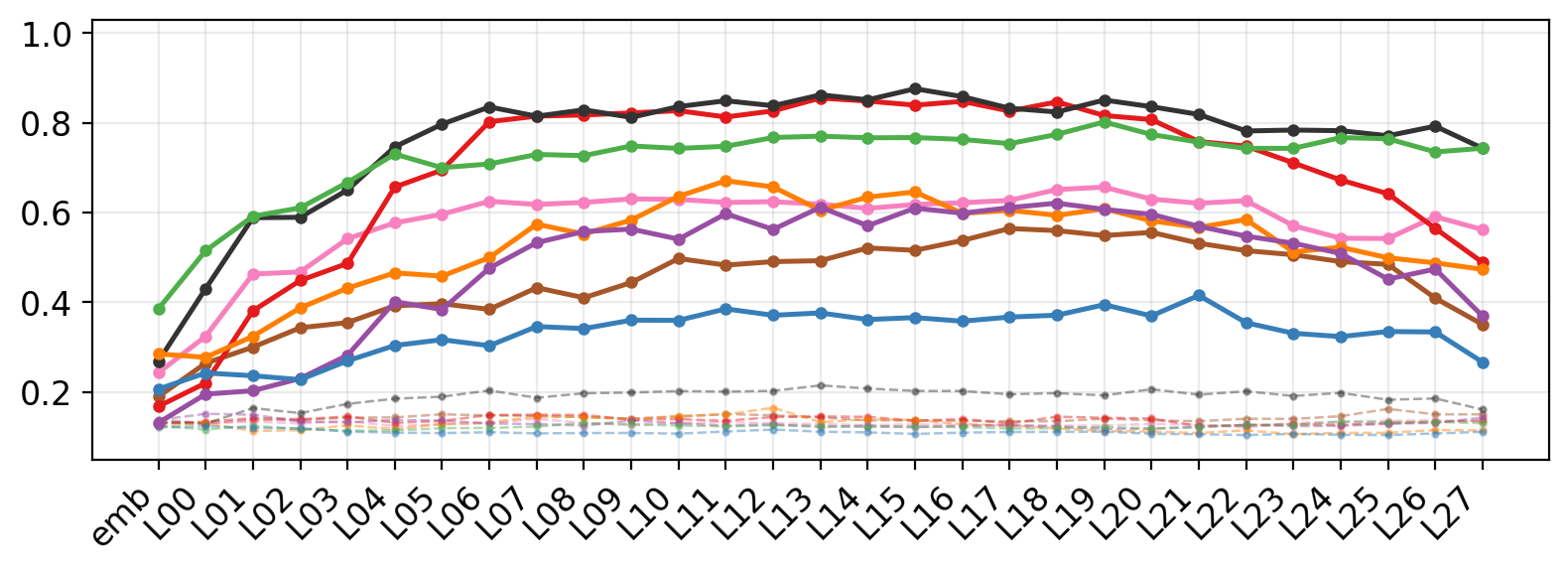}
        \caption{AUPRC, Qwen2.5-7B}
    \end{subfigure}

    \begin{subfigure}[b]{0.48\textwidth}
        \centering
        \includegraphics[width=\textwidth,height=2.5cm]{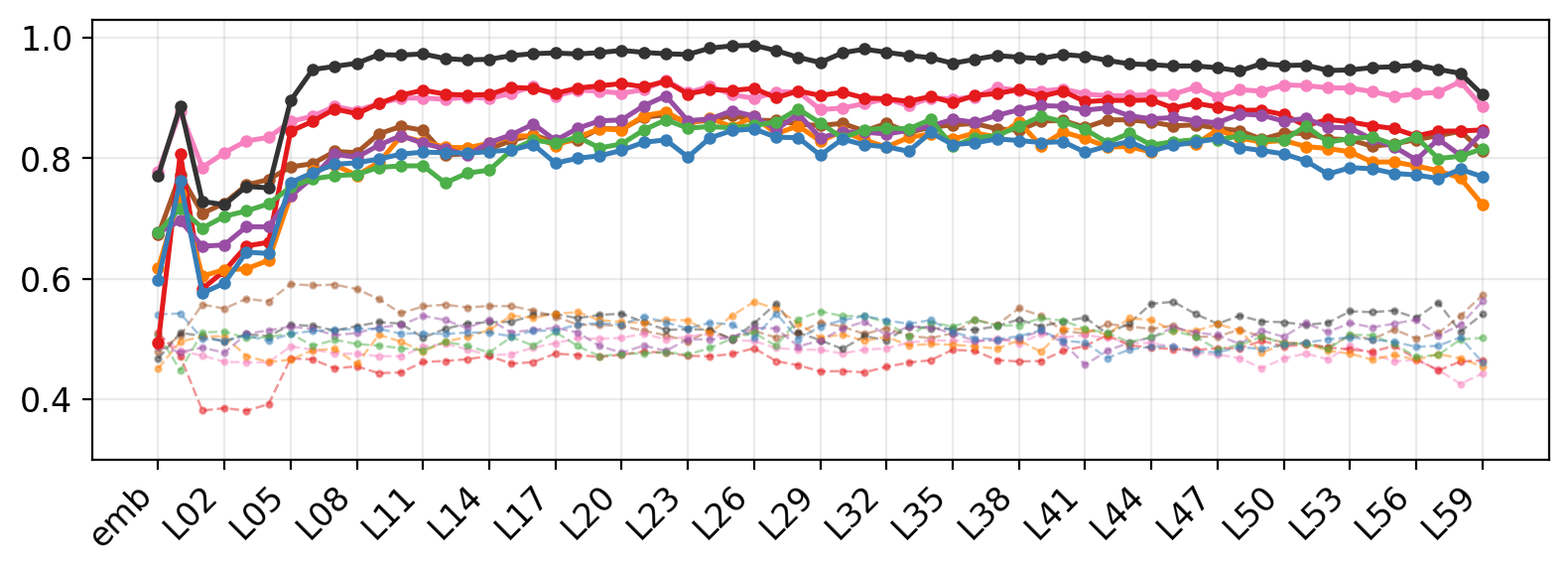}
        \caption{AUROC, Gemma4-31B}
    \end{subfigure} 
    \hfill
    \begin{subfigure}[b]{0.48\textwidth}
        \centering
        \includegraphics[width=\textwidth,height=2.5cm]{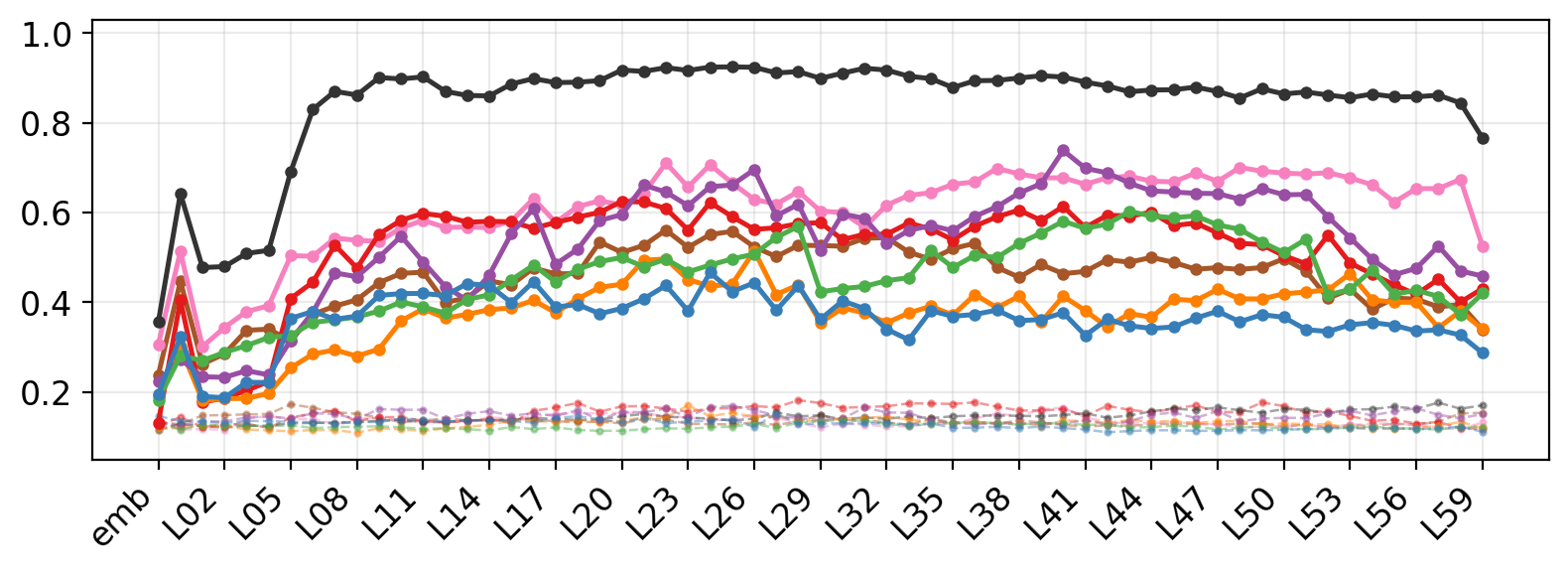}
        \caption{AUPRC, Gemma4-31B}
    \end{subfigure}
    \vspace{-.5em}
    \caption{
    \textbf{quantitative analyses on separability of reasoning operations in representation spaces at position middel.}
    AUROC and AUPRC of one-vs-rest operation classifiers are shown across layers for \texttt{Qwen3-8B}, \texttt{Qwen2.5-7B}, and \texttt{Gemma4-31B} using middle-token span representations. Solid lines denote true reasoning-operation labels, while dashed lines denote random-label and random-position baselines. Across models, reasoning operation-alignment scores remain substantially above the baselines and peak in middle layers.
    }
    \label{fig:AUROC-AUPRC-middle-res}
    \vspace{-.5em}
\end{figure*}

\begin{figure*}[t]
    \centering
    % \def\mode{res}
    % \def\position{mean}
    
    % legend
    {
    \begin{subfigure}[b]{\textwidth}
    \centering
    \footnotesize
    \renewcommand{\arraystretch}{0.7}
    \begin{tabular}{@{}ll@{\hspace{0.9em}}ll@{\hspace{0.9em}}ll@{\hspace{0.9em}}ll@{}}
        \tikz{\draw[line width=0.8mm, color=extraction] (0,0) -- (0.35,0);} &
        \texttt{Extraction} &
        \tikz{\draw[line width=0.8mm, color=symbolization] (0,0) -- (0.35,0);} &
        \texttt{Direct mapping} &
        \tikz{\draw[line width=0.8mm, color=structural] (0,0) -- (0.35,0);} &
        \texttt{Decomposition} &
        \tikz{\draw[line width=0.8mm, color=retrieval] (0,0) -- (0.35,0);} &
        \texttt{Recall} \\
    
        \tikz{\draw[line width=0.8mm, color=deduction] (0,0) -- (0.35,0);} &
        \texttt{Deduction} &
        \tikz{\draw[line width=0.8mm, color=algebraic] (0,0) -- (0.35,0);} &
        \texttt{Algebraic manipulation} &
        \tikz{\draw[line width=0.8mm, color=arithmetic] (0,0) -- (0.35,0);} &
        \texttt{Arithmetic computation} &
        \tikz{\draw[line width=0.8mm, color=finalanswer] (0,0) -- (0.35,0);} &
        \texttt{Final answer}
    \end{tabular}
    \end{subfigure}
    }

    \begin{subfigure}[b]{0.48\textwidth}
        \centering
        \includegraphics[width=\textwidth,height=3cm]{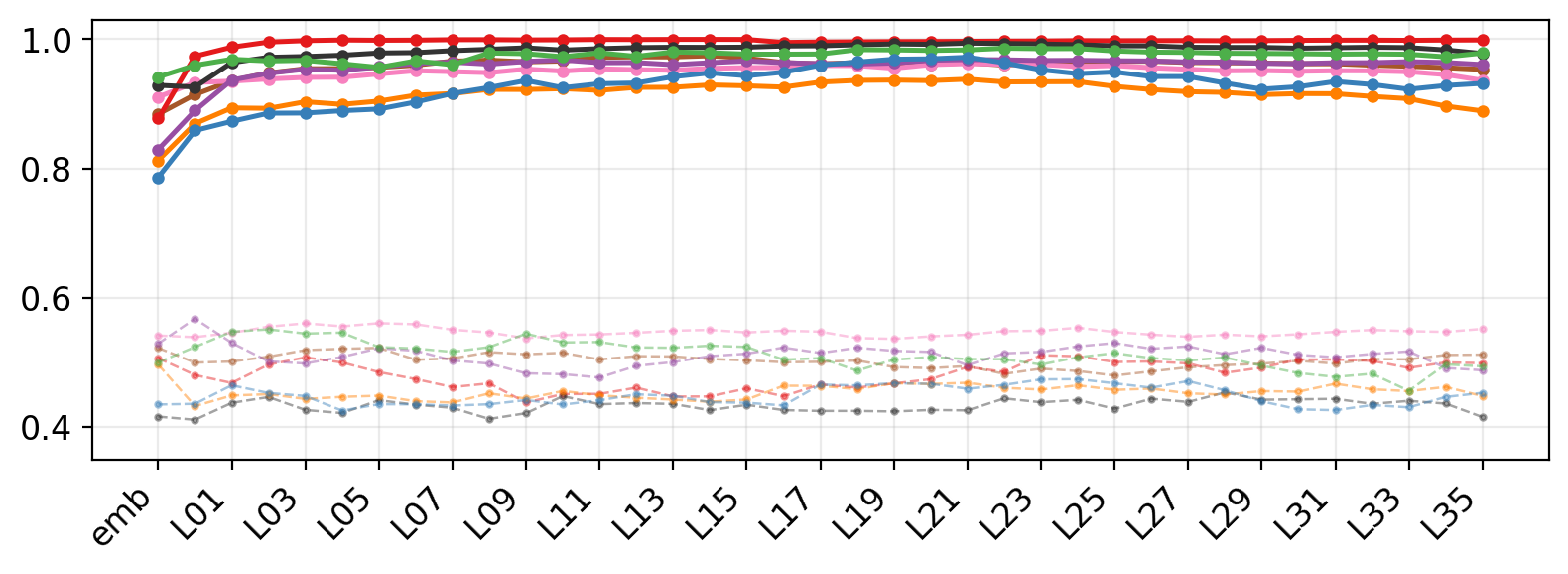}
        \caption{Qwen3-8B}
    \end{subfigure}
    \hfill
    \begin{subfigure}[b]{0.48\textwidth}
        \centering
        \includegraphics[width=\textwidth,height=3cm]{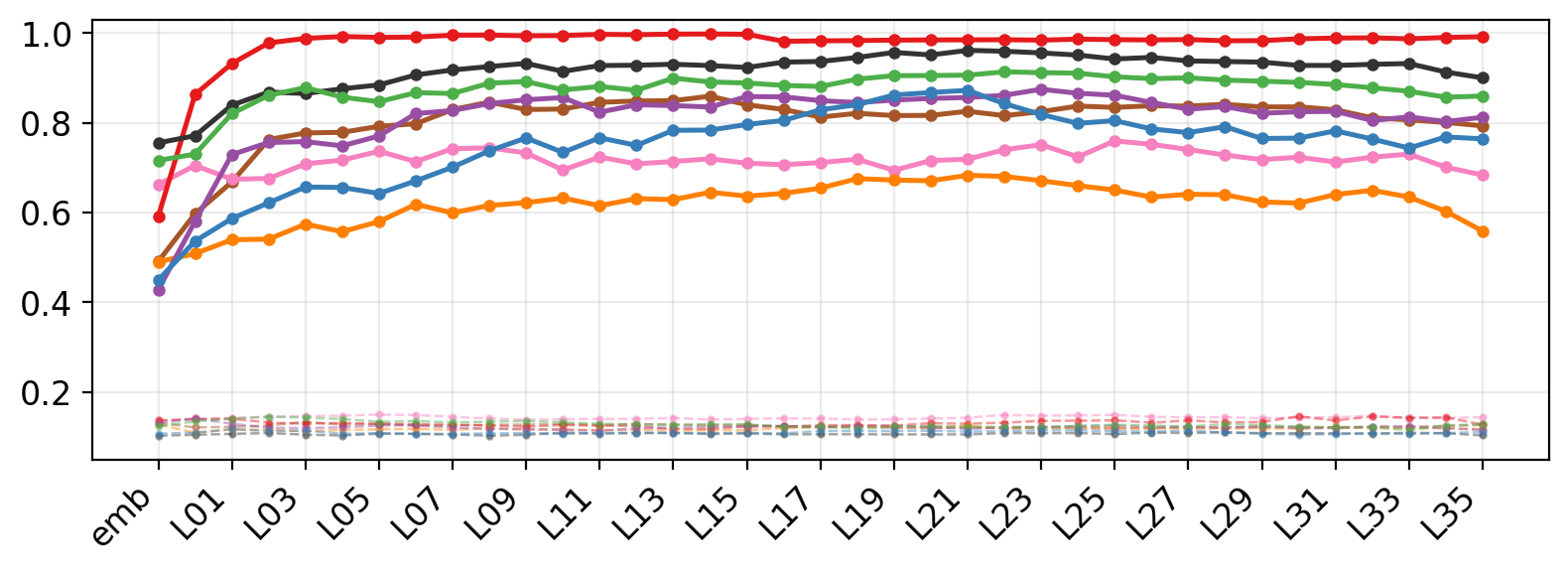}
        \caption{Qwen3-8B}
    \end{subfigure}

    \begin{subfigure}[b]{0.48\textwidth}
        \centering
        \includegraphics[width=\textwidth,height=3cm]{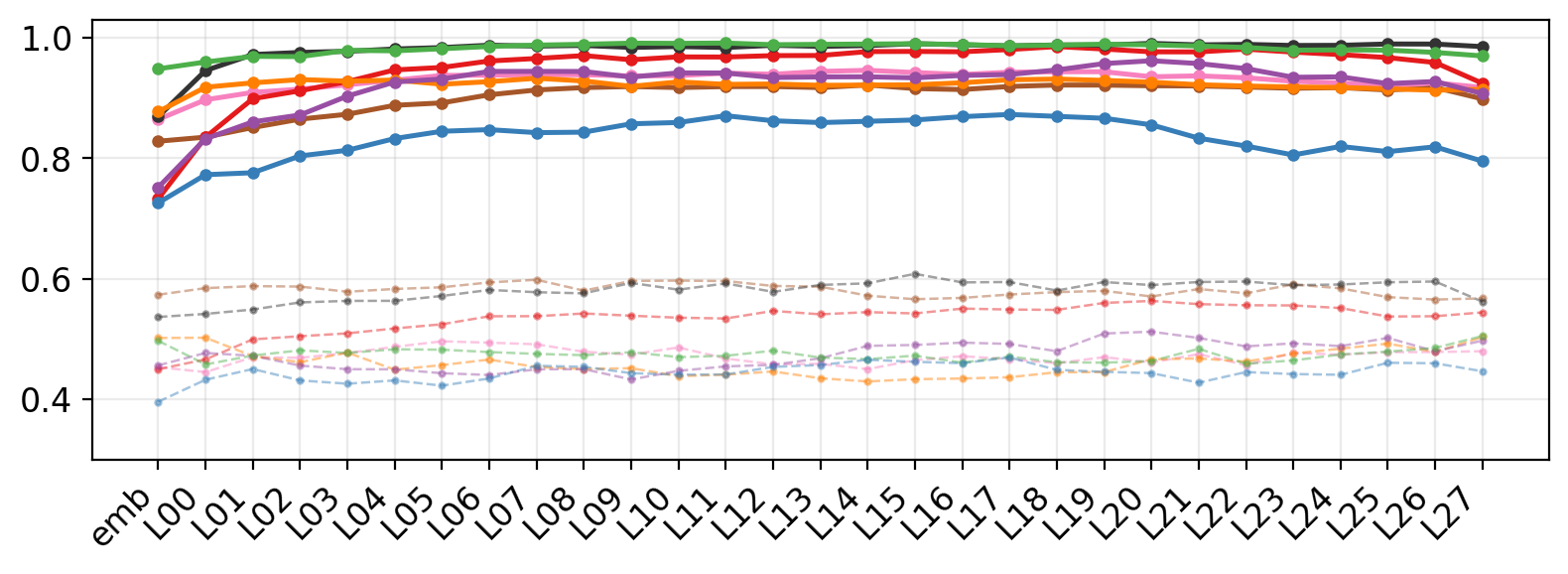}
        \caption{Qwen2.5-7B}
    \end{subfigure}
    \hfill
    \begin{subfigure}[b]{0.48\textwidth}
        \centering
        \includegraphics[width=\textwidth,height=3cm]{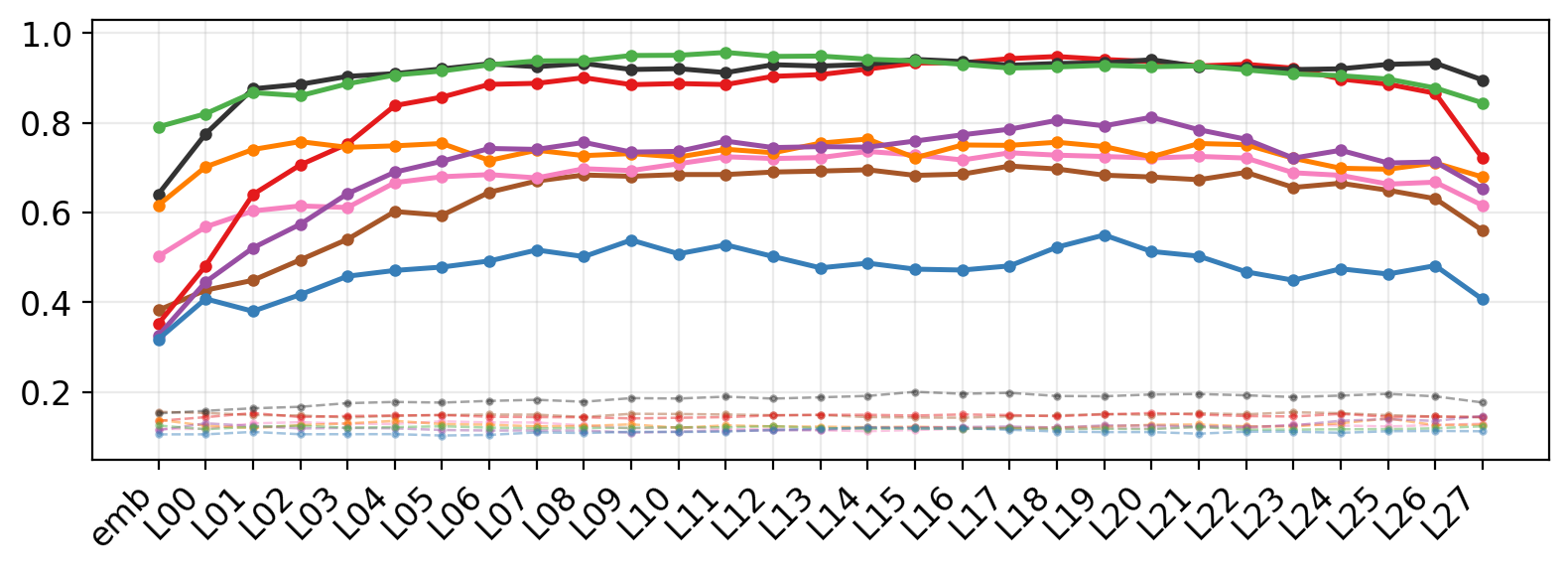}
        \caption{Qwen2.5-7B}
    \end{subfigure}

    \begin{subfigure}[b]{0.48\textwidth}
        \centering
        \includegraphics[width=\textwidth,height=3cm]{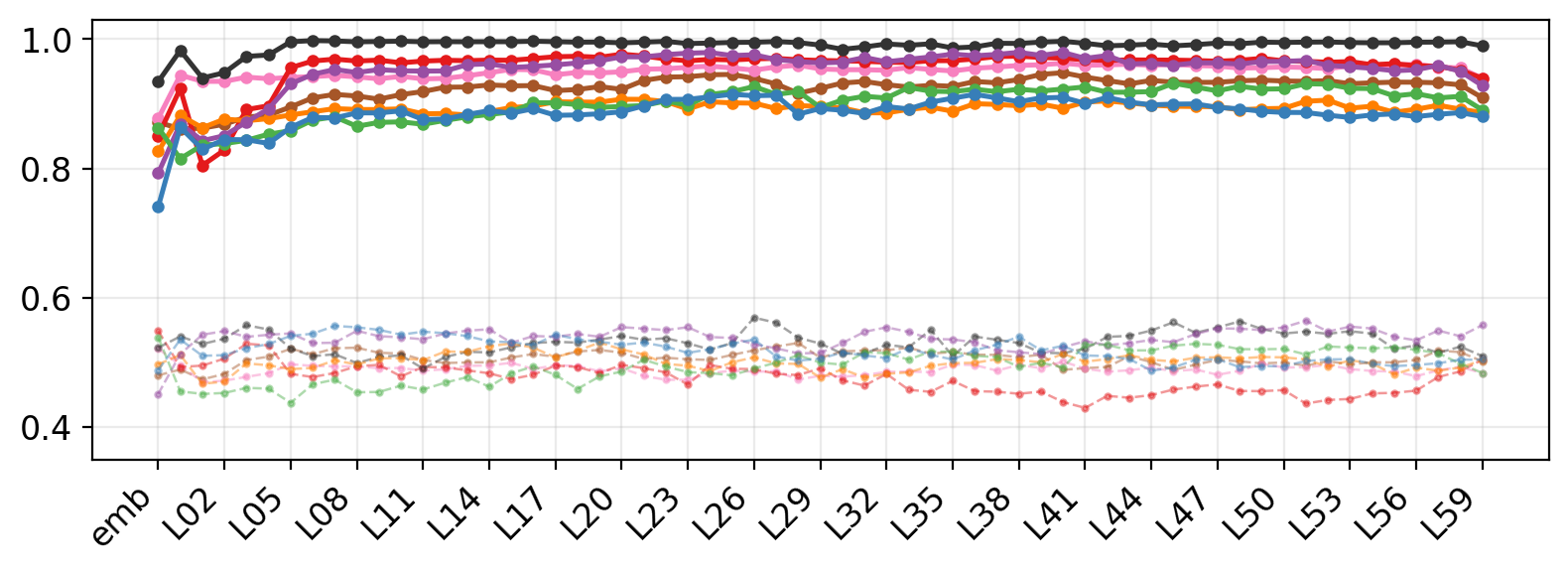}
        \caption{Gemma4-31B}
    \end{subfigure}
    \hfill
    \begin{subfigure}[b]{0.48\textwidth}
        \centering
        \includegraphics[width=\textwidth,height=3cm]{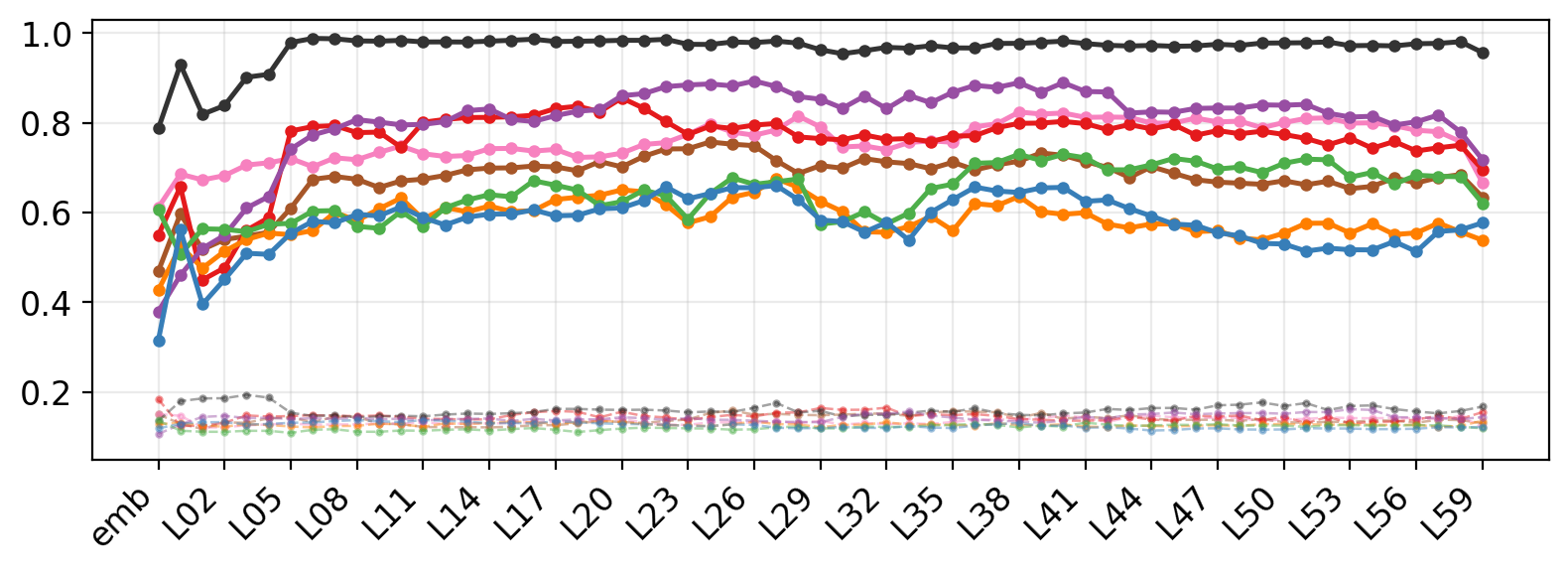}
        \caption{Gemma4-31B}
    \end{subfigure}
    
    \caption{\textbf{Quantitative analyses on separability of reasoning operations in representation spaces at position mean.} AUROC and AUPRC of one-vs-rest operation classifiers are shown across layers for \texttt{Qwen3-8B}, \texttt{Qwen2.5-7B}, and \texttt{Gemma4-31B}. Solid lines denote true reasoning-operation labels, while dashed lines denote random-label and random-position baselines. Across models, performance remains above the baselines and typically peaks in middle layers.}
    \label{fig:AUROC-AUPRC-mean-res}
    
\end{figure*}

\begin{figure*}[t]
    \centering
    % \def\mode{res}
    % \def\position{first}
    
    % Local scope for \dataset, \target, and \modelpath
    {
        % legend
        \begin{subfigure}[b]{\textwidth}
        \centering
        \footnotesize
        \renewcommand{\arraystretch}{0.6}
        \begin{tabular}{@{}ll@{\hspace{1.2em}}ll@{}}
            \tikz{\draw[line width=1mm, color=extraction] (0,0) -- (0.55,0);} &
            \texttt{[EXTRACTION]} &
            \tikz{\draw[line width=1mm, color=symbolization] (0,0) -- (0.55,0);} &
            \texttt{[SYMBOLIZATION.DIRECT-MAPPING]} \\
    
            \tikz{\draw[line width=1mm, color=structural] (0,0) -- (0.55,0);} &
            \texttt{[STRUCTURAL-ANALYSIS.DECOMPOSITION]} &
            \tikz{\draw[line width=1mm, color=retrieval] (0,0) -- (0.55,0);} &
            \texttt{[RETRIEVAL.RECALL]} \\
    
            \tikz{\draw[line width=1mm, color=deduction] (0,0) -- (0.55,0);} &
            \texttt{[INFERENCE-FLOW.CHAINING.DEDUCTION]} &
            \tikz{\draw[line width=1mm, color=algebraic] (0,0) -- (0.55,0);} &
            \texttt{[EXECUTION.OPERATION.ALGEBRAIC-MANIPULATION]} \\
    
            \tikz{\draw[line width=1mm, color=arithmetic] (0,0) -- (0.55,0);} &
            \texttt{[EXECUTION.OPERATION.ARITHMETIC-COMPUTATION]} &
            \tikz{\draw[line width=1mm, color=finalanswer] (0,0) -- (0.55,0);} &
            \texttt{[FINAL-ANSWER]}
        \end{tabular}
    \end{subfigure}
    
        % \def\model{Qwen3-8B}
        % graph plots
        \begin{subfigure}[b]{0.48\textwidth}
            \centering

            \includegraphics[width=\textwidth,height=3cm]{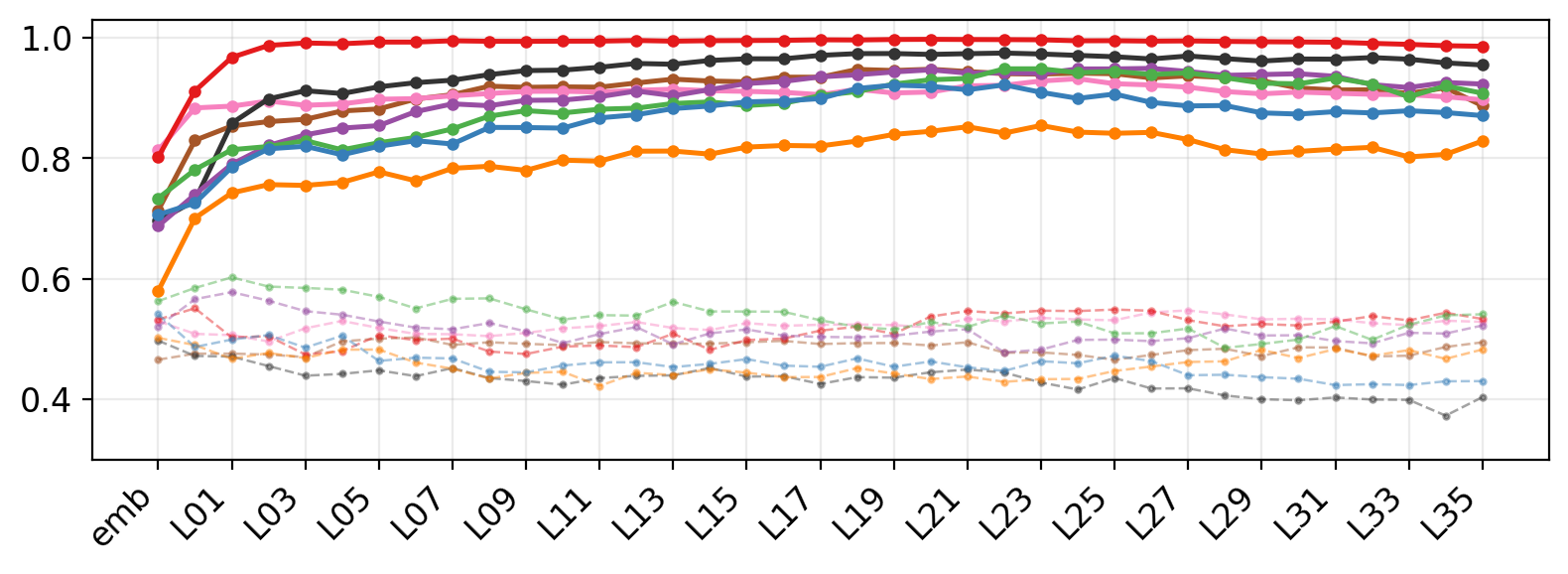}
            \caption{Qwen3-8B}
        \end{subfigure} 
        \hfill
        \begin{subfigure}[b]{0.48\textwidth}
            \centering
            \includegraphics[width=\textwidth,height=3cm]{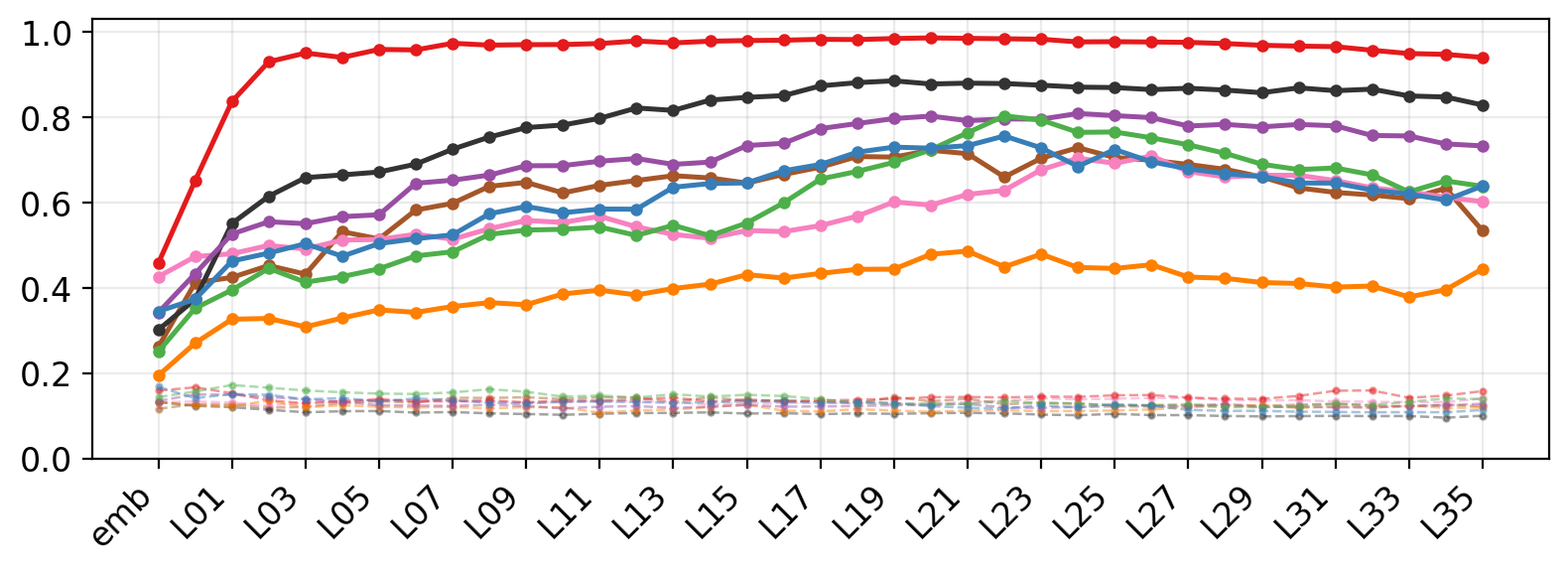}
            \caption{Qwen3-8B}
            % \rule{8cm}{3cm} 
            % \caption{.}
        \end{subfigure}
        
        % \def\model{Qwen2.5-7B}
        % graph plots
        \begin{subfigure}[b]{0.48\textwidth}
            \centering
            \includegraphics[width=\textwidth,height=3cm]{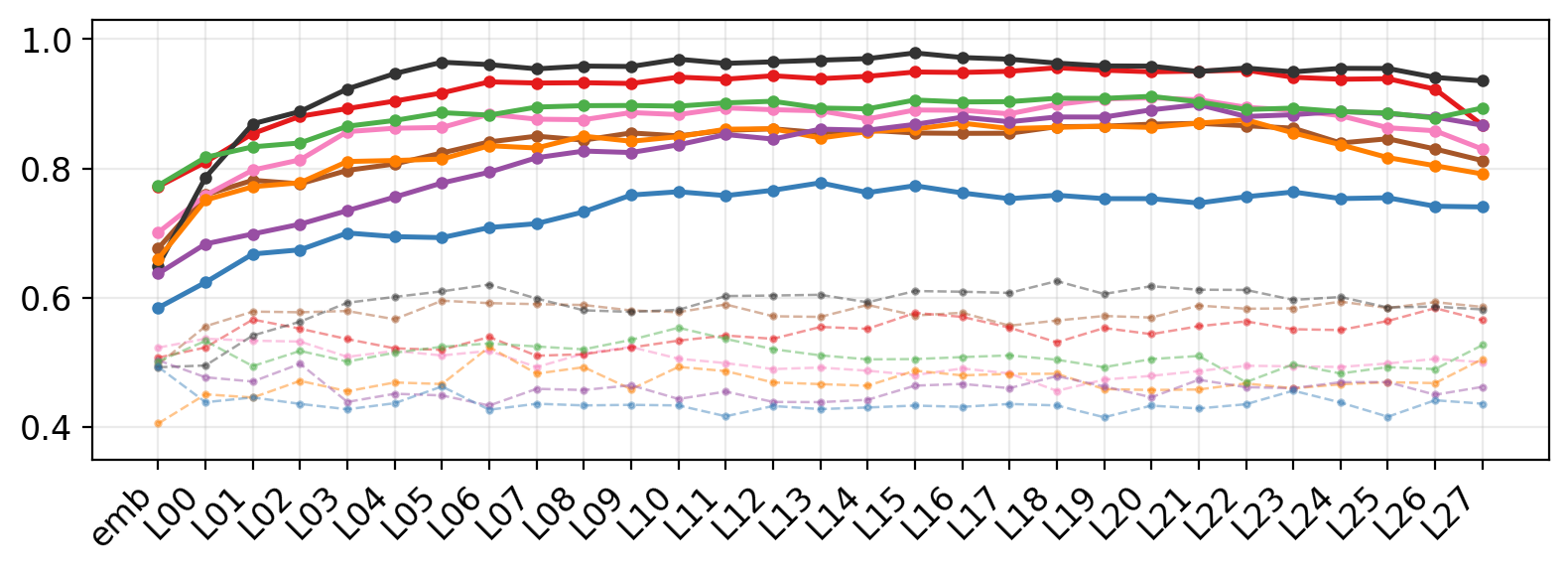}
            \caption{Qwen2.5-7B}
        \end{subfigure} 
        \hfill
        \begin{subfigure}[b]{0.48\textwidth}
            \centering
            \includegraphics[width=\textwidth,height=3cm]{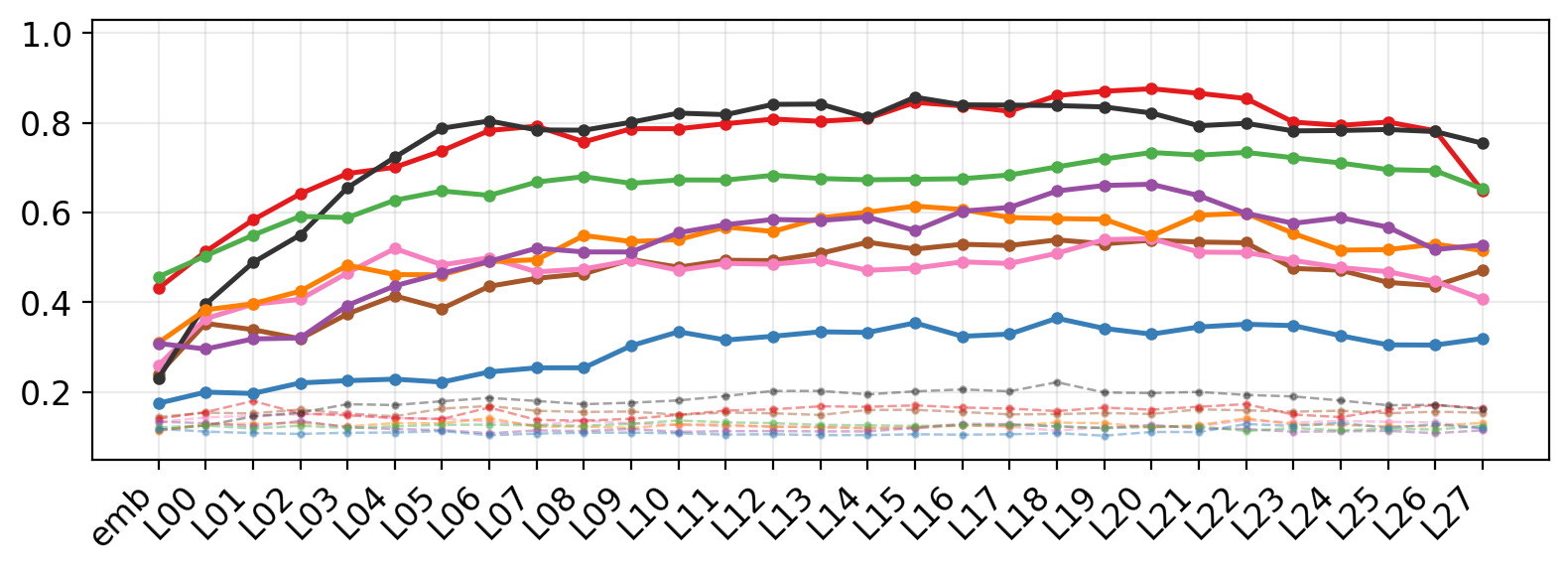}
            \caption{Qwen2.5-7B}
            % \rule{8cm}{3cm} 
            % \caption{.}
        \end{subfigure}
        
        % \def\model{Gemma4-31B}
        % graph plots
        \begin{subfigure}[b]{0.48\textwidth}
            \centering
            \includegraphics[width=\textwidth,height=3cm]{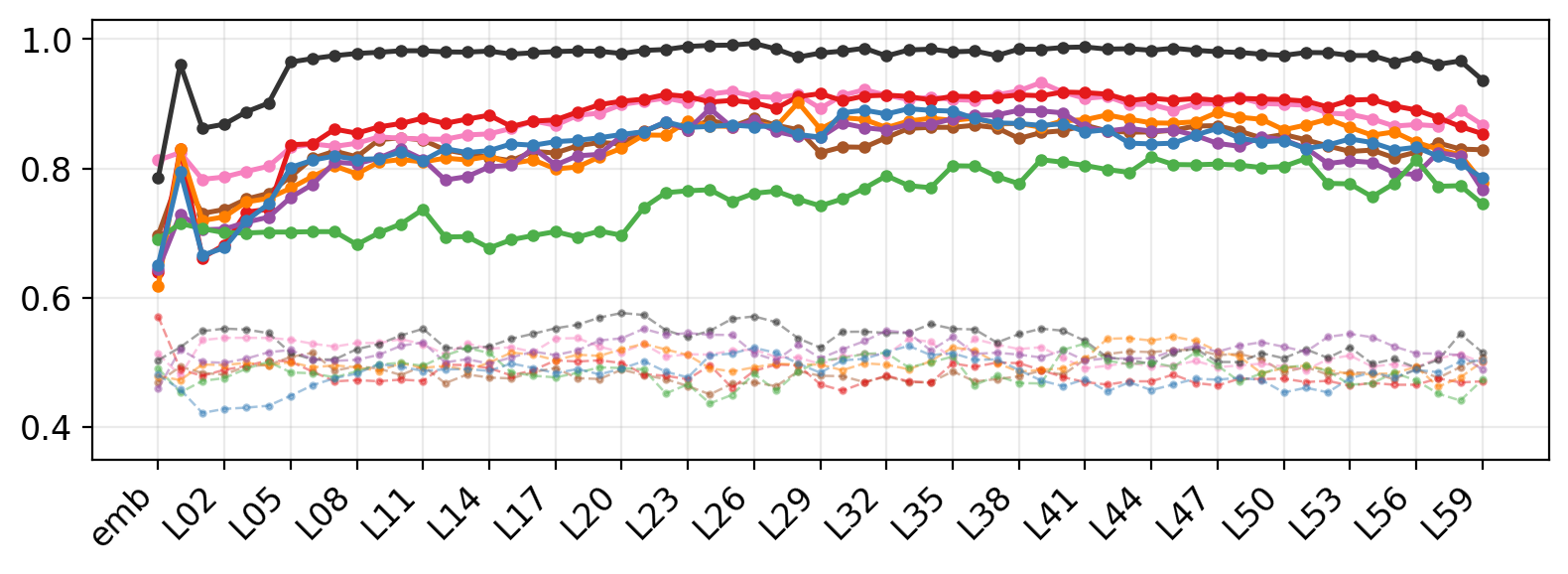}
            \caption{Gemma4-31B}
        \end{subfigure} 
        \hfill
        \begin{subfigure}[b]{0.48\textwidth}
            \centering
            \includegraphics[width=\textwidth,height=3cm]{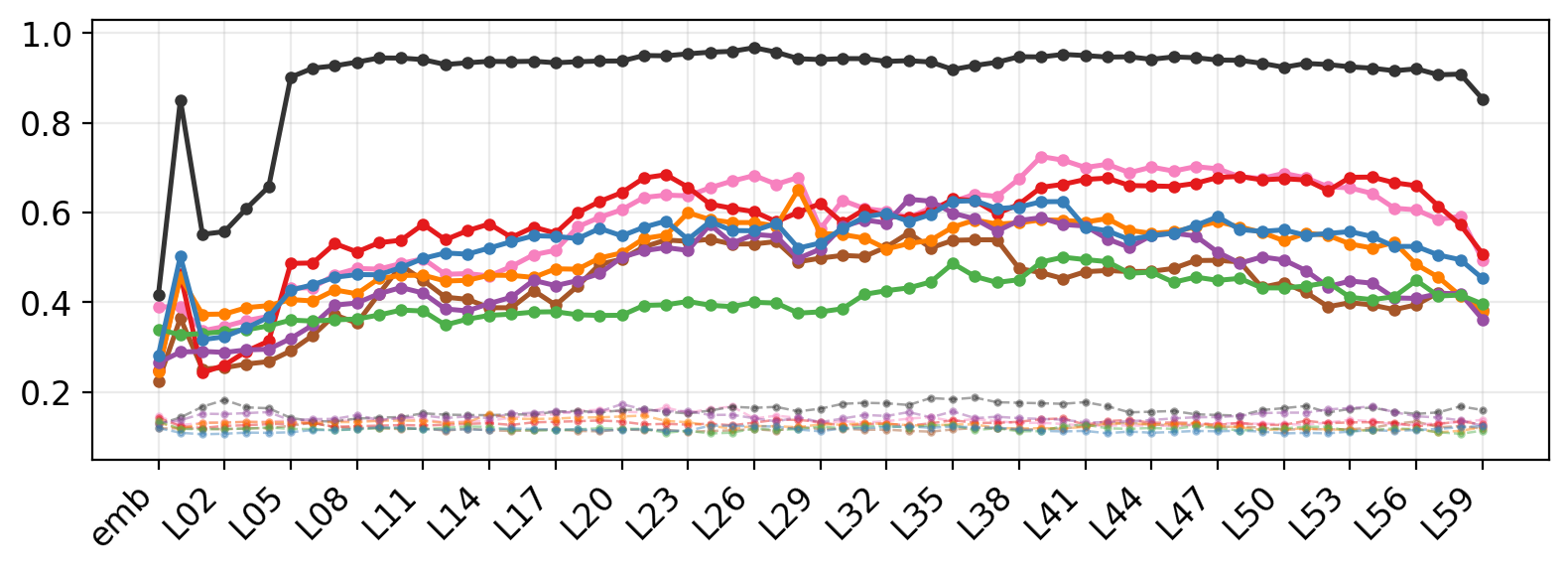}
            \caption{Gemma4-31B}
            % \rule{8cm}{3cm} 
            % \caption{.}
        \end{subfigure}
    }

    \caption{\textbf{Quantitative analyses on separability of reasoning operations in representation spaces at position first.} AUROC and AUPRC of one-vs-rest operation classifiers are shown across layers for \texttt{Qwen3-8B}, \texttt{Qwen2.5-7B}, and \texttt{Gemma4-31B}. Solid lines denote true reasoning-operation labels, while dashed lines denote random-label and random-position baselines. Across models, performance remains above the baselines and typically peaks in middle layers.
    }
    \label{fig:AUROC-AUPRC-first-res}
    
\end{figure*}

\begin{figure*}[t]
    \centering
    % \def\mode{res}
    
    % Local scope for \dataset, \target, and \modelpath
    {
        % legend
        \begin{subfigure}[b]{\textwidth}
        \centering
        \footnotesize
        \renewcommand{\arraystretch}{0.6}
        \begin{tabular}{@{}ll@{\hspace{1.2em}}ll@{}}
            \tikz{\draw[line width=1mm, color=extraction] (0,0) -- (0.55,0);} &
            \texttt{[EXTRACTION]} &
            \tikz{\draw[line width=1mm, color=symbolization] (0,0) -- (0.55,0);} &
            \texttt{[SYMBOLIZATION.DIRECT-MAPPING]} \\
    
            \tikz{\draw[line width=1mm, color=structural] (0,0) -- (0.55,0);} &
            \texttt{[STRUCTURAL-ANALYSIS.DECOMPOSITION]} &
            \tikz{\draw[line width=1mm, color=retrieval] (0,0) -- (0.55,0);} &
            \texttt{[RETRIEVAL.RECALL]} \\
    
            \tikz{\draw[line width=1mm, color=deduction] (0,0) -- (0.55,0);} &
            \texttt{[INFERENCE-FLOW.CHAINING.DEDUCTION]} &
            \tikz{\draw[line width=1mm, color=algebraic] (0,0) -- (0.55,0);} &
            \texttt{[EXECUTION.OPERATION.ALGEBRAIC-MANIPULATION]} \\
    
            \tikz{\draw[line width=1mm, color=arithmetic] (0,0) -- (0.55,0);} &
            \texttt{[EXECUTION.OPERATION.ARITHMETIC-COMPUTATION]} &
            \tikz{\draw[line width=1mm, color=finalanswer] (0,0) -- (0.55,0);} &
            \texttt{[FINAL-ANSWER]}
        \end{tabular}
    \end{subfigure}
    
        % \def\model{Qwen3-8B}
        % graph plots
        \begin{subfigure}[b]{0.48\textwidth}
            \centering
            \includegraphics[width=\textwidth,height=3cm]{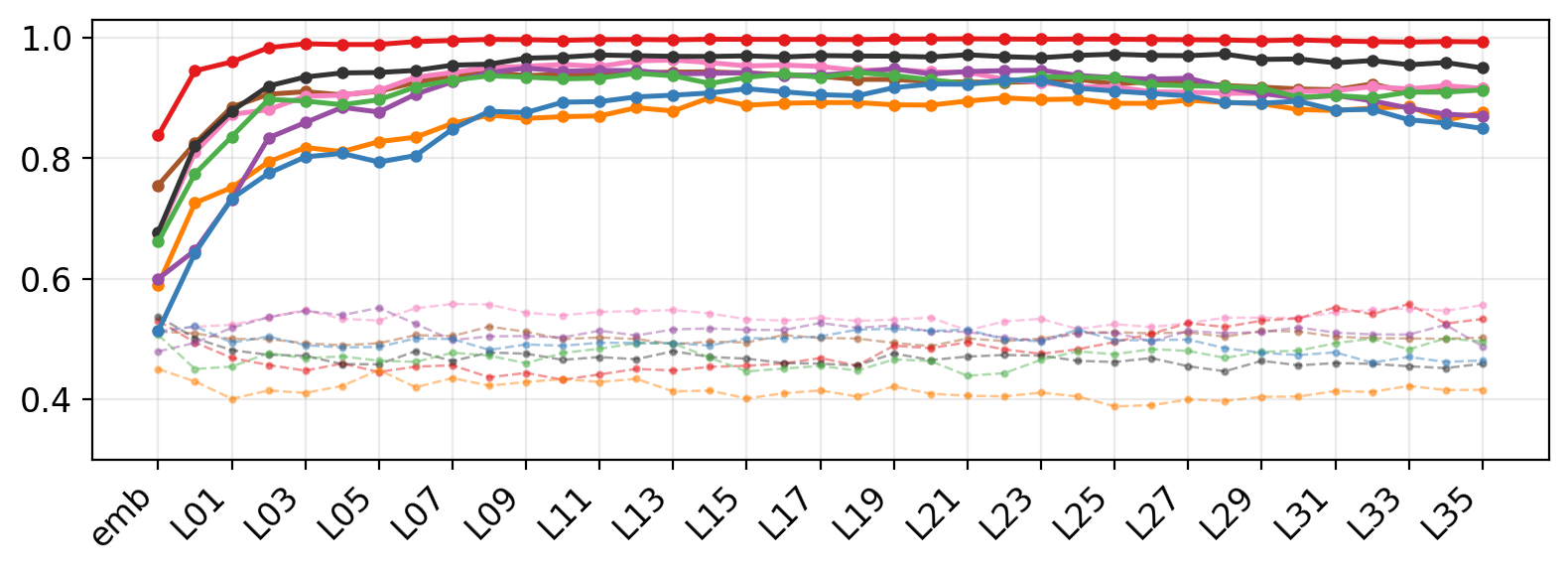}
            \caption{Qwen3-8B}
        \end{subfigure} 
        \hfill
        \begin{subfigure}[b]{0.48\textwidth}
            \centering
            \includegraphics[width=\textwidth,height=3cm]{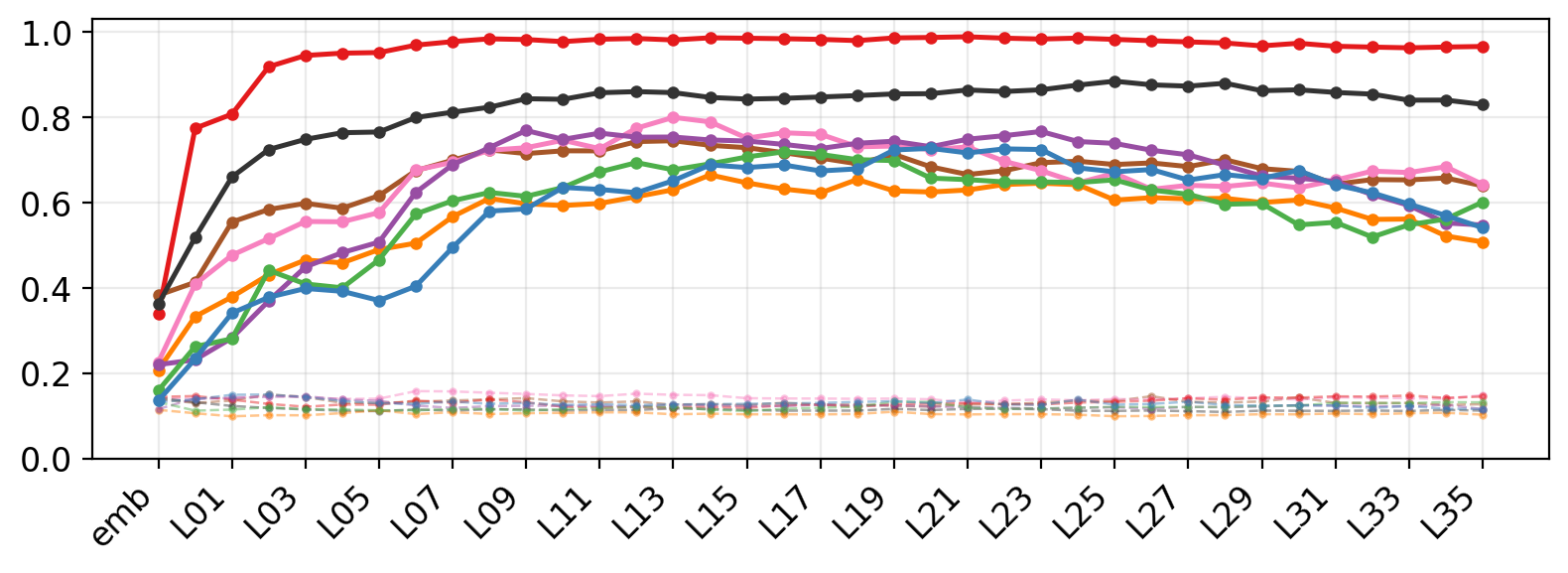}
            \caption{Qwen3-8B}
            % \rule{8cm}{3cm} 
            % \caption{.}
        \end{subfigure}
        
        % \def\model{Qwen2.5-7B}
        % graph plots
        \begin{subfigure}[b]{0.48\textwidth}
            \centering
            \includegraphics[width=\textwidth,height=3cm]{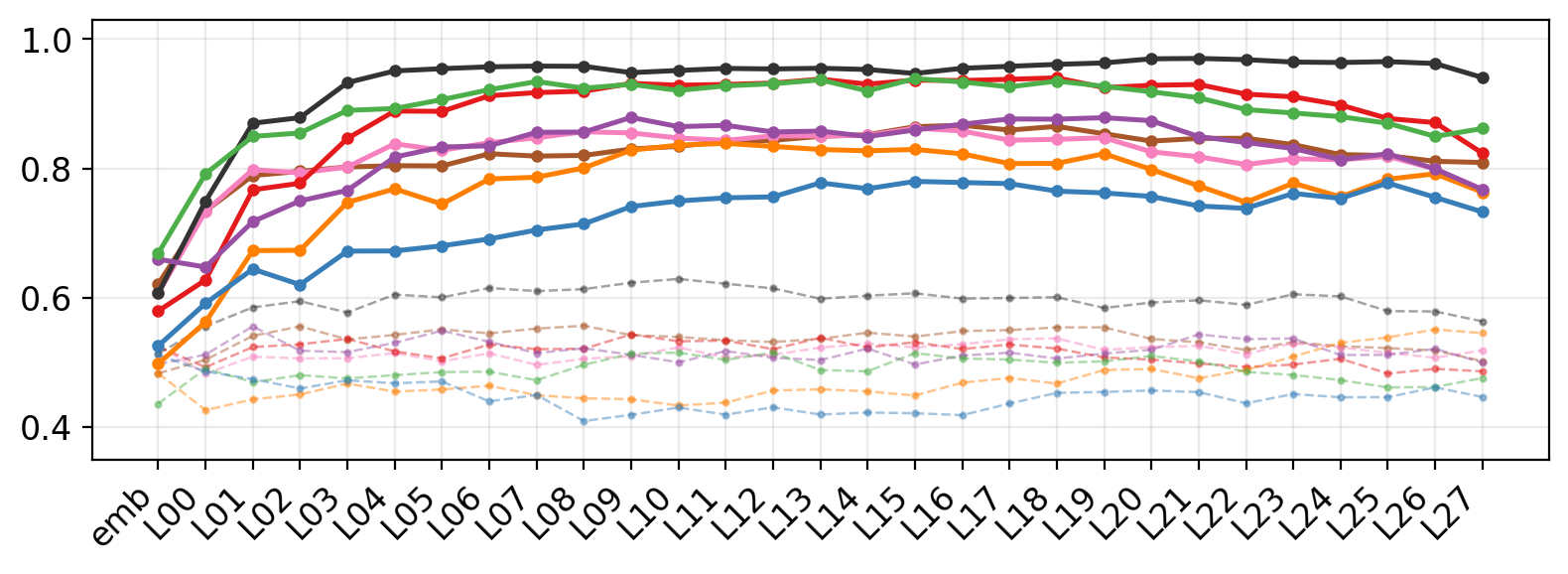}
            \caption{Qwen2.5-7B}
        \end{subfigure} 
        \hfill
        \begin{subfigure}[b]{0.48\textwidth}
            \centering
            \includegraphics[width=\textwidth,height=3cm]{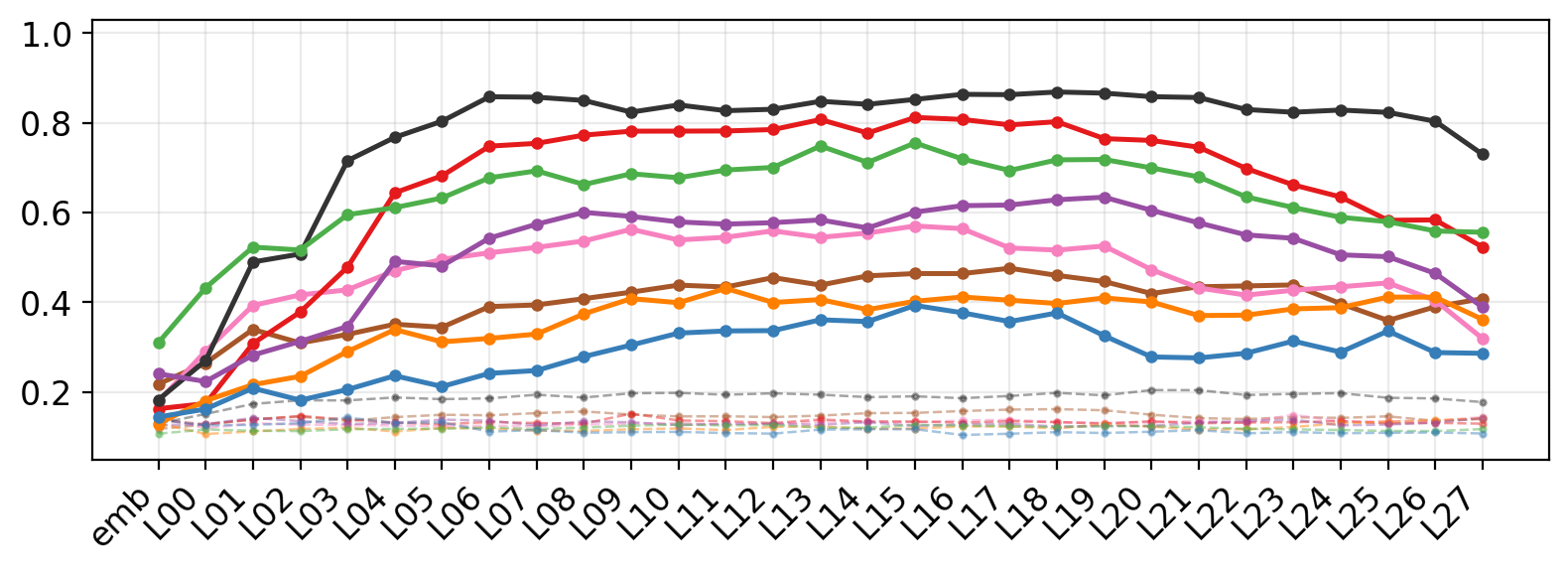}
            \caption{Qwen2.5-7B}
            % \rule{8cm}{3cm} 
            % \caption{.}
        \end{subfigure}
        
        % \def\model{Gemma4-31B}
        % graph plots
        \begin{subfigure}[b]{0.48\textwidth}
            \centering
            \includegraphics[width=\textwidth,height=3cm]{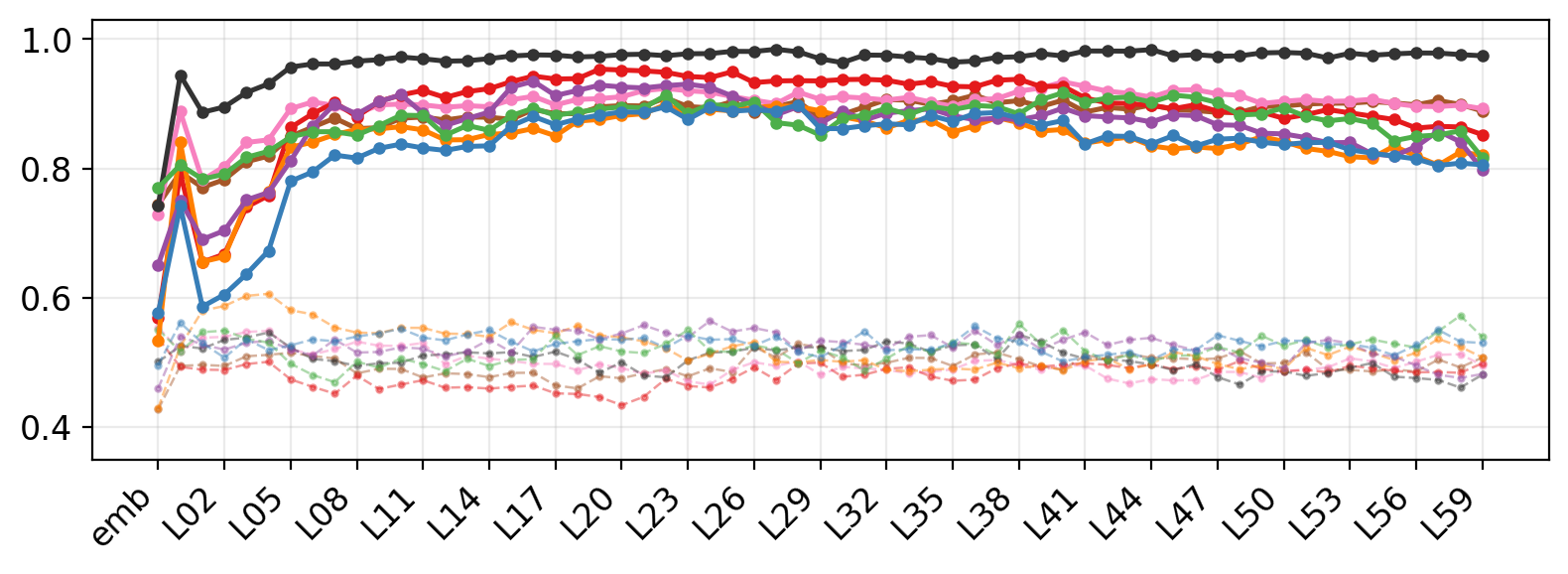}
            \caption{Gemma4-31B}
        \end{subfigure} 
        \hfill
        \begin{subfigure}[b]{0.48\textwidth}
            \centering
            \includegraphics[width=\textwidth,height=3cm]{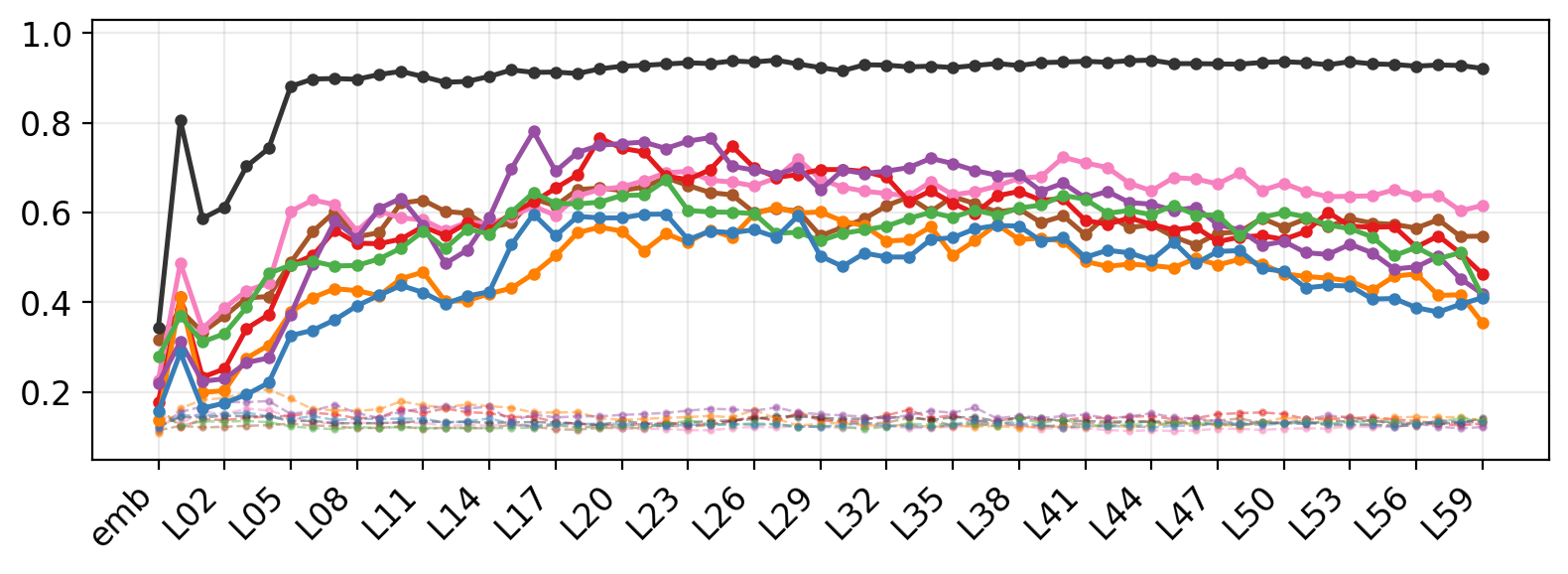}
            \caption{Gemma4-31B}
            % \rule{8cm}{3cm} 
            % \caption{.}
        \end{subfigure}
    }

    \caption{\textbf{Quantitative analyses on separability of reasoning operations in representation spaces at position last.} AUROC and AUPRC of one-vs-rest operation classifiers are shown across layers for \texttt{Qwen3-8B}, \texttt{Qwen2.5-7B}, and \texttt{Gemma4-31B}. Solid lines denote true reasoning-operation labels, while dashed lines denote random-label and random-position baselines. Across models, performance remains above the baselines and typically peaks in middle layers.
    }
    \label{fig:AUROC-AUPRC-last-res}
    
\end{figure*}

Figures~\ref{fig:AUROC-AUPRC-middle-res}, \ref{fig:AUROC-AUPRC-mean-res}, \ref{fig:AUROC-AUPRC-first-res}, and \ref{fig:AUROC-AUPRC-last-res} report the full layer-wise separability results for all models, reasoning operations, and span-representation choices, including middle-token, mean-pooled, first-token, and last-token representations. 
The first-token and last-token settings show small quantitative differences, but they largely preserve the same layer-wise trend observed in the main results.
Separability generally increases from early layers to middle layers and remains above the random baseline across most operation types.
This suggests that operation-level information is available at multiple positions within a reasoning operation chunk, although single-token representations can be more sensitive to local lexical content or boundary effects.

The mean-pooled representation shows a somewhat different pattern.
Compared with single-token representations, mean pooling often yields stronger separability from earlier layers.
This is expected because mean pooling aggregates information across multiple tokens in the span, which can make operation-level signals more stable and reduce the effect of isolated token-level noise.
As discussed in Section~\ref{sec:sep_evolve}, early-layer reasoning operation-alignment scores can be sparse and token-local, whereas middle-layer scores become more distributed across the span.
The stronger early-layer performance of mean pooling is therefore consistent with the intra-span variance analysis: aggregating over tokens can capture multiple local cues even before the operation signal becomes fully distributed in middle layers.

AUPRC is particularly informative in our one-vs-rest setting, where each target operation constitutes only a fraction of the evaluation examples and thus requires maintaining high precision while recovering the positive class. 
The consistently strong AUPRC results therefore provide complementary evidence that reasoning-operation separability is robust across models, layers, and representation positions.

Taken together, these position-variant results support the robustness of our main finding.
Reasoning operations are geometrically separable not only at the middle token, but also under alternative choices of span representation.
At the same time, the differences between single-token and mean-pooled representations indicate that the measured separability depends partly on how much within-span context is aggregated.

\begin{table*}[t]
\centering
\resizebox{\textwidth}{!}{%
\begin{tabular}{llrrr}
\toprule
\textbf{Stage} & \textbf{Reasoning operation} & \textbf{Qwen2.5-7B} & \textbf{Qwen3-8B} & \textbf{Gemma4-31B} \\
\midrule
\multirow{5}{*}{\textbf{Stage 1: Understanding the Problem}}
& Extraction & 590 & 1,035 & 682 \\
& Direct-mapping & 624 & 913 & 1,289 \\
& Reframing & 146 & 390 & 393 \\
& Abstraction & 8 & 35 & 36 \\
& State-space-definition & 190 & 315 & 259 \\
\midrule
\multirow{6}{*}{\textbf{Stage 2: Planning the Solution}}
& Pattern-matching & 225 & 322 & 389 \\
& Symmetry & 22 & 113 & 89 \\
& Invariance & 15 & 46 & 55 \\
& Decomposition & 419 & 285 & 280 \\
& Idealization & 14 & 26 & 30 \\
& Hypothesis-formation & 59 & 162 & 72 \\
\midrule
\multirow{12}{*}{\textbf{Stage 3: Carrying Out the Plan}}
& Recall & 808 & 1,903 & 1,725 \\
& Matching & 99 & 155 & 154 \\
& Deduction & 405 & 1,301 & 1,276 \\
& Induction & 2 & 2 & 6 \\
& Analogy & 2 & 1 & 1 \\
& Branching & 78 & 183 & 216 \\
& Instantiation & 366 & 307 & 660 \\
& Algebraic-manipulation & 842 & 2,172 & 2,669 \\
& Arithmetic-computation & 726 & 3,331 & 3,742 \\
& Logical-evaluation & 251 & 603 & 715 \\
& Representation-construction & 86 & 352 & 321 \\
& Pattern-extraction & 6 & 40 & 43 \\
\midrule
\multirow{5}{*}{\textbf{Stage 4: Looking Back and Final Answer}}
& Strategy-validation & 61 & 490 & 138 \\
& Error-detection & 88 & 161 & 63 \\
& Dimensional-analysis & 2 & 52 & 3 \\
& Extreme-case-testing & 3 & 39 & 10 \\
& Final-answer & 406 & 619 & 854 \\
\bottomrule
\end{tabular}%
}
\caption{Train + Test occurrence counts for each reasoning operation and model.}
\label{tab:train-test-reasoning-occurrence}
\end{table*}

\subsection{Position Controls}
\label{app:position-controls}
\begin{table}[t]
\centering
\small
\resizebox{\columnwidth}{!}{
\begin{tabular}{lccc}
\toprule
\textbf{Model}
& \textbf{Position}
& \textbf{Hidden}
& \textbf{$\Delta$} \\
\midrule
Qwen3-8B
& $0.718/0.279$
& $0.937/0.742$
& $+0.218/+0.463$ \\
Qwen2.5-7B
& $0.708/0.269$
& $0.895/0.646$
& $+0.187/+0.377$ \\
Gemma4-31B
& $0.751/0.311$
& $0.899/0.641$
& $+0.148/+0.330$ \\
\bottomrule
\end{tabular}
}
\caption{
Macro AUROC/AUPRC of position-only and mean-pooled hidden-state
predictors. $\Delta$ denotes hidden-state minus position-only
performance.
}
\label{tab:position-only}
\end{table}

\subsubsection{Position-Only Classification}

We divide each reasoning trace into 50 equal-width intervals according to relative token position. 
Each span is represented by a 50-dimensional binary vector indicating the intervals that it occupies. 
Using the same question-level train--test split, we train one-vs-rest logistic-regression classifiers to predict the operation label from this position vector alone.

Table~\ref{tab:position-only} shows that Position alone is informative, particularly for operations with strong trace-order regularities such as Extraction and Final Answer.
Nevertheless, the hidden-state probe outperforms the position-only predictor for every operation in all three models.

\subsubsection{Position-Stratified Evaluation}
\begin{table}[t]
\centering
\small
\resizebox{\columnwidth}{!}{
\begin{tabular}{lccc}
\toprule
\textbf{Position}
& \textbf{Labels}
& \textbf{AUROC/AUPRC}
& \textbf{Chance AUPRC} \\
\midrule
$[0.0,0.2)$ & 7 & $0.973/0.898$ & $0.143$ \\
$[0.2,0.4)$ & 5 & $0.930/0.808$ & $0.200$ \\
$[0.4,0.6)$ & 5 & $0.947/0.836$ & $0.200$ \\
$[0.6,0.8)$ & 5 & $0.916/0.769$ & $0.200$ \\
$[0.8,1.0]$ & 5 & $0.927/0.798$ & $0.200$ \\
\bottomrule
\end{tabular}
}
\caption{
Position-stratified Qwen3-8B evaluation using mean-pooled
representations. Macro averages include operations with sufficient
examples in each interval.
}
\label{tab:position-stratified}
\end{table}

We divide held-out spans into five intervals according to normalized start position, defined as the span's start-token index divided by the number of reasoning tokens in the trace. 
Within each interval, we evaluate the original frozen hidden-state probe using only spans in that interval. 
We sample up to 20 examples per operation and retain operations with at least 10 examples.

Table~\ref{tab:position-stratified} shows that Operation labels remain distinguishable when evaluated among spans from similar trace positions. Position therefore contributes to the prediction of some labels but does not fully explain the operation-level hidden-state structure.

\subsection{Lexical Controls}
\label{app:lexical-controls}

\subsubsection{Text-Only Classification}
\begin{table}[t]
\centering
\small
\resizebox{\columnwidth}{!}{
\begin{tabular}{lccc}
\toprule
\textbf{Model}
& \textbf{Text}
& \textbf{Hidden}
& \textbf{$\Delta$} \\
\midrule
Qwen3-8B
& $0.849/0.549$
& $0.937/0.742$
& $+0.088/+0.193$ \\
Qwen2.5-7B
& $0.854/0.562$
& $0.895/0.646$
& $+0.041/+0.084$ \\
Gemma4-31B
& $0.802/0.466$
& $0.899/0.641$
& $+0.097/+0.175$ \\
\bottomrule
\end{tabular}
}
\caption{
Macro AUROC/AUPRC of the stronger text-only baseline among TF-IDF and BoW and the
mean-pooled hidden-state probe.
}
\label{tab:text-only}
\end{table}

We train logistic-regression classifiers using bag-of-words and TF--IDF features extracted from each span. 
These classifiers use the same question-level split, class balancing, and operation labels as the hidden-state analysis. 
We compare the stronger text-only baseline with mean-pooled residual-stream probes.

Table~\ref{tab:text-only} shows that Lexical content is clearly informative, but the hidden-state probe outperforms the strongest text-only baseline in aggregate for all three models. 
Decomposition is the main operation-level exception: its stereotyped expressions make it particularly predictable from sparse lexical features, and its text-only AUPRC slightly exceeds the hidden-state AUPRC. 
We therefore do not claim that lexical content is irrelevant; rather, it does not account for the full cross-model hidden-state performance.

\subsubsection{Lexically Matched Pair Analysis}
\begin{table*}[t]
\centering
\small
\begin{tabular}{lrrl}
\toprule
\textbf{Operation}
& \textbf{Median $\Delta$}
& \textbf{$\Delta>0$}
& \textbf{95\% CI} \\
\midrule
Extraction
& 4.68 & 87.5\% & $[3.13,6.45]$ \\
Direct Mapping
& 0.57 & 75.0\% & $[0.03,1.53]$ \\
Decomposition
& 3.91 & 100.0\% & $[3.63,4.41]$ \\
Recall
& 1.05 & 70.0\% & $[-0.46,2.79]$ \\
Deduction
& 0.61 & 76.9\% & $[0.13,1.41]$ \\
Algebraic Manipulation
& 0.68 & 60.7\% & $[-0.26,1.27]$ \\
Arithmetic Computation
& 0.45 & 61.5\% & $[-0.19,0.88]$ \\
Final Answer
& 1.75 & 92.9\% & $[0.78,2.59]$ \\
\bottomrule
\end{tabular}
\caption{
Lexically matched pair analysis. Positive effects indicate that the
frozen probe is more strongly aligned with operation identity than
with broad lexical similarity. Confidence intervals are computed
across anchor-level effects.
}
\label{tab:lexically-matched}
\end{table*}

We test whether spans with similar lexical content remain distinguishable when they express different operations. 
We use the original held-out test set and extract lexical and hidden-state features from the same fixed 50-token windows. Lexical similarity is computed as cosine similarity between model-token bag-of-words count vectors.

For each anchor span $A$ with operation label $c$, we select a lexically similar span $B$ with a different operation and a lexically dissimilar span $C$ with the same operation. 
Neither hidden representations nor probe scores are used in the matching procedure, and the probe is not retrained.

Let $s_c(x)$ denote the frozen probe score of span $x$ along the anchor operation direction $c$. 
We define
\[
d_c(A,X)=|s_c(A)-s_c(X)|
\]
and the anchor-level effect
\[
\Delta_A=d_c(A,B)-d_c(A,C).
\]
A positive value indicates that the lexically matched but operation-mismatched span is farther from the anchor than the lexically dissimilar same-operation span. 
When an anchor has multiple valid comparisons, we first aggregate effects within the anchor and perform inference across anchors.

Table~\ref{tab:lexically-matched} shows that the median effect is positive for all eight operations. 
Confidence intervals are above zero for Extraction, Direct Mapping, Decomposition, Deduction, and Final Answer, while Recall, Algebraic Manipulation, and Arithmetic Computation show positive but weaker operation-level evidence. 
The results indicate that broad lexical similarity alone does not account for the frozen probe geometry.

\subsubsection{Competing-Operation Vocabulary Subsets}
\begin{table*}[t]
\centering
\small
\begin{tabular}{lcccc}
\toprule
& \multicolumn{2}{c}{\textbf{Qwen3-8B}}
& \multicolumn{2}{c}{\textbf{Qwen2.5-7B}} \\
\cmidrule(lr){2-3}\cmidrule(lr){4-5}
\textbf{Operation}
& \textbf{AUROC}
& \textbf{AUPRC}
& \textbf{AUROC}
& \textbf{AUPRC} \\
\midrule
Extraction             & 0.997 & 0.992 & 0.957 & 0.904 \\
Direct Mapping         & 0.888 & 0.811 & 0.753 & 0.499 \\
Decomposition          & 0.899 & 0.806 & 0.873 & 0.838 \\
Recall                 & 0.923 & 0.893 & 0.878 & 0.664 \\
Deduction              & 0.835 & 0.653 & 0.862 & 0.744 \\
Algebraic Manipulation & 0.926 & 0.908 & 0.887 & 0.768 \\
Arithmetic Computation & 0.921 & 0.766 & 0.904 & 0.917 \\
Final Answer           & 0.961 & 0.957 & 0.961 & 0.978 \\
\midrule
Macro                   & 0.919 & 0.848 & 0.884 & 0.789 \\
\bottomrule
\end{tabular}
\caption{
Frozen-probe performance on held-out spans containing vocabulary
associated with competing operation labels.
}
\label{tab:competing-vocabulary}
\end{table*}

The preceding analysis controls for broad lexical similarity, but the probe could still depend on a small set of highly operation-associated cue words. 
We therefore compute class-based TF--IDF using only the training spans and extract the top 5\% of words associated with each operation. 
We then evaluate the frozen probe on held-out spans that contain vocabulary associated with a competing operation. 
A span may belong to more than one subset when it contains cues associated with multiple competing labels.

Table~\ref{tab:competing-vocabulary} shows that reasoning probes remain predictive when spans contain vocabulary associated with competing operations. 
Such cue words are therefore generally insufficient to override the operation label expressed by the span, although lexical cues may still contribute to individual
predictions.

\subsubsection{Digit and Formula Density}
\begin{table*}[t]
\centering
\small
\begin{tabular}{lcccc}
\toprule
\textbf{Operation}
& \textbf{Chance AUPRC}
& \multicolumn{2}{c}{\textbf{Mean pooling}}
& \multicolumn{1}{c}{\textbf{Middle token}} \\
\cmidrule(lr){3-4}
& & \textbf{AUROC} & \textbf{AUPRC}
& \textbf{AUROC/AUPRC} \\
\midrule
Direct Mapping
& 0.204 & 0.994 & 0.979 & $0.925/0.808$ \\
Deduction
& 0.184 & 0.866 & 0.701 & $0.833/0.574$ \\
Algebraic Manipulation
& 0.204 & 0.912 & 0.755 & $0.811/0.498$ \\
Arithmetic Computation
& 0.204 & 0.816 & 0.510 & $0.780/0.458$ \\
Final Answer
& 0.204 & 0.999 & 0.998 & $0.969/0.924$ \\
\midrule
Macro
& 0.200 & 0.917 & 0.789 & $0.864/0.652$ \\
\bottomrule
\end{tabular}
\caption{
Evaluation on spans with 50--75\% digit or mathematical-token density.
The theoretical chance AUROC is $0.500$ for all rows.
}
\label{tab:density-control}
\end{table*}

We directly test whether Arithmetic Computation is separable merely because its spans contain more digits and mathematical notation. 
For each test span, we calculate the proportion of tokens classified as digits or mathematical tokens and restrict evaluation to spans with density between 50\% and 75\%. 
The resulting subset contains 49 test spans across five operations: Direct Mapping (10), Deduction (9), Algebraic Manipulation (10), Arithmetic Computation (10), and Final Answer (10).

Table~\ref{tab:density-control} shows that all five operations remain above chance within the numerically dense subset. 
Arithmetic Computation itself remains distinguishable, showing that digit and formula density alone is insufficient to explain its separability. 
This is a targeted density control rather than a complete removal of lexical variation, since spans may still differ in their specific symbols and numerical expressions.

\subsection{Statistical Reliability of Representational Separability}
\label{app:statistical-tests}
\begin{table*}[t]
\centering
\scriptsize
\setlength{\tabcolsep}{5pt}
\begin{tabular}{lcccc}
\toprule
\textbf{Reasoning operation}
& \textbf{Mean AUROC [95\% CI]}
& \textbf{Mean AUPRC [95\% CI]}
& \textbf{Middle AUROC [95\% CI]}
& \textbf{Middle AUPRC [95\% CI]} \\
\midrule

\multicolumn{5}{l}{\textbf{Qwen3-8B (layers 10--30)}} \\
Extraction
& 0.998 [0.995, 1.000] & 0.988 [0.975, 0.998]
& 0.995 [0.991, 0.998] & 0.970 [0.943, 0.989] \\
Direct mapping
& 0.947 [0.921, 0.969] & 0.802 [0.724, 0.873]
& 0.875 [0.835, 0.911] & 0.587 [0.490, 0.683] \\
Decomposition
& 0.980 [0.961, 0.994] & 0.895 [0.824, 0.954]
& 0.926 [0.890, 0.956] & 0.636 [0.514, 0.750] \\
Recall
& 0.965 [0.939, 0.986] & 0.847 [0.761, 0.925]
& 0.917 [0.875, 0.952] & 0.725 [0.640, 0.809] \\
Deduction
& 0.927 [0.898, 0.953] & 0.648 [0.545, 0.753]
& 0.883 [0.848, 0.916] & 0.531 [0.437, 0.640] \\
Algebraic manipulation
& 0.965 [0.946, 0.981] & 0.833 [0.751, 0.908]
& 0.932 [0.905, 0.955] & 0.681 [0.584, 0.780] \\
Arithmetic computation
& 0.956 [0.934, 0.974] & 0.723 [0.617, 0.841]
& 0.912 [0.873, 0.946] & 0.648 [0.544, 0.762] \\
Final answer
& 0.989 [0.981, 0.995] & 0.940 [0.897, 0.974]
& 0.960 [0.936, 0.979] & 0.835 [0.754, 0.905] \\

\midrule
\multicolumn{5}{l}{\textbf{Qwen2.5-7B (layers 8--23)}} \\
Extraction
& 0.975 [0.950, 0.993] & 0.919 [0.861, 0.965]
& 0.929 [0.885, 0.965] & 0.813 [0.735, 0.883] \\
Direct mapping
& 0.854 [0.804, 0.898] & 0.499 [0.394, 0.610]
& 0.815 [0.768, 0.857] & 0.368 [0.291, 0.468] \\
Decomposition
& 0.988 [0.979, 0.995] & 0.935 [0.885, 0.973]
& 0.924 [0.888, 0.956] & 0.759 [0.673, 0.838] \\
Recall
& 0.943 [0.912, 0.967] & 0.764 [0.669, 0.852]
& 0.877 [0.830, 0.916] & 0.581 [0.474, 0.693] \\
Deduction
& 0.924 [0.873, 0.966] & 0.741 [0.637, 0.847]
& 0.899 [0.852, 0.939] & 0.602 [0.504, 0.717] \\
Algebraic manipulation
& 0.919 [0.885, 0.948] & 0.685 [0.588, 0.781]
& 0.862 [0.819, 0.901] & 0.511 [0.409, 0.629] \\
Arithmetic computation
& 0.939 [0.912, 0.963] & 0.718 [0.611, 0.826]
& 0.904 [0.872, 0.933] & 0.624 [0.521, 0.728] \\
Final answer
& 0.987 [0.978, 0.995] & 0.928 [0.881, 0.967]
& 0.956 [0.928, 0.979] & 0.834 [0.755, 0.906] \\

\midrule
\multicolumn{5}{l}{\textbf{Gemma4-31B (layers 13--40)}} \\
Extraction
& 0.969 [0.955, 0.981] & 0.796 [0.718, 0.871]
& 0.909 [0.879, 0.936] & 0.581 [0.490, 0.695] \\
Direct mapping
& 0.898 [0.862, 0.930] & 0.617 [0.514, 0.720]
& 0.822 [0.777, 0.863] & 0.395 [0.320, 0.497] \\
Decomposition
& 0.908 [0.860, 0.947] & 0.649 [0.532, 0.765]
& 0.841 [0.784, 0.892] & 0.485 [0.368, 0.613] \\
Recall
& 0.970 [0.955, 0.983] & 0.857 [0.791, 0.916]
& 0.858 [0.814, 0.898] & 0.591 [0.505, 0.679] \\
Deduction
& 0.896 [0.857, 0.930] & 0.613 [0.515, 0.719]
& 0.842 [0.803, 0.878] & 0.411 [0.342, 0.502] \\
Algebraic manipulation
& 0.932 [0.904, 0.956] & 0.715 [0.623, 0.799]
& 0.850 [0.807, 0.888] & 0.504 [0.417, 0.597] \\
Arithmetic computation
& 0.954 [0.926, 0.976] & 0.766 [0.674, 0.861]
& 0.906 [0.874, 0.933] & 0.636 [0.547, 0.723] \\
Final answer
& 0.993 [0.984, 0.999] & 0.976 [0.950, 0.995]
& 0.972 [0.951, 0.989] & 0.903 [0.848, 0.949] \\

\bottomrule
\end{tabular}
\caption{
Bootstrap confidence intervals for reasoning-operation separability.
We report layer-averaged AUROC and AUPRC with 95\% confidence intervals
for mean-pooled and middle-token representations.
Intervals are computed from 5,000 stratified bootstrap replicates.
}
\label{tab:bootstrap-confidence-intervals}
\end{table*}

We quantify sampling uncertainty using 5,000 stratified bootstrap replicates. 
For each one-vs-rest operation, we separately resample positive and negative held-out examples with replacement, preserving the class composition of the evaluation set. 
Within each replicate, AUROC and AUPRC are computed at each layer and then averaged over the model-specific intermediate-layer ranges: layers 10--30 for \texttt{Qwen3-8B}, 8--23 for \texttt{Qwen2.5-7B}, and 13--40 for \texttt{Gemma4-31B}. 
We report results for both mean-pooled and middle-token representations, with 95\% confidence intervals defined by the 2.5th and 97.5th percentiles of the bootstrap distribution.

Table~\ref{tab:bootstrap-confidence-intervals} reports the resulting confidence intervals. 
Across all three models, eight operations, and both span representations, all 48 AUROC confidence intervals remain entirely above the chance level of 0.5, indicating that the observed separability is stable under resampling of the held-out examples.

We further assess whether the observed separability could arise under random training-label assignments using 1,000 one-sided permutation tests. 
For each permutation, we shuffle the training labels once and use the same shuffled assignment across layers, refitting the probe separately at each layer. AUROC and AUPRC are then averaged over the same model-specific intermediate-layer ranges used above.

Across three models, two span representations, eight operation labels, and two evaluation metrics, all 96 permutation tests are significant at $p<0.001$. 
Together, the bootstrap confidence intervals and permutation tests show that operation-level separability is stable under held-out resampling and unlikely to arise from random training-label assignments.

\section{Reasoning-Operation Taxonomy and Annotation Process}
\label{app:full_taxonomy_annotation}
\subsection{Full Taxonomy of Reasoning Operations}
\label{app:full_taxonomy}

The main analysis focuses on eight recurring reasoning-operation types that appear frequently enough for representation-level analysis.
At Table~\ref{tab:reasoning_type_schema_1}--\ref{tab:reasoning_type_schema_3}, we provide the full hierarchical taxonomy used for annotation, including additional operation types and subtypes that were used to organize the annotation schema but were not included in the main separability experiments.
The taxonomy follows the four problem-solving stages introduced in the main text and specifies each operation by its functional role in the generated reasoning trace.

\subsection{Annotation Prompt and Span-Selection Protocol}
\label{app:annotation_prompt}
We provide the full prompt used to annotate generated reasoning traces into reasoning-operation spans.
The prompt provides the full taxonomy, span-selection rules, output JSON format, and examples for standardizing the annotation procedure.

\subsection{Human Validation of Operation-Span Annotations}
\label{app:human-validation}
\begin{table}[t]
\centering
\small
\begin{tabular}{llr}
\toprule
\textbf{Category} & \textbf{Metric} & \textbf{Result} \\
\midrule
Samples
& Validation spans
& 84 \\
\midrule
Human agreement
& Unanimous
& 50/84 (59.5\%) \\
& Majority label
& 81/84 (96.4\%) \\
& Fleiss' $\kappa$
& 0.666 \\
\midrule
Reference labels
& Determined
& 81/84 (96.4\%) \\
& Not determined
& 3/84 (3.6\%) \\
\midrule
Human--GPT-5
& Exact agreement
& 64/84 (76.2\%) \\
& Cohen's $\kappa$
& 0.715 \\
& Soft agreement
& 67.5/84 (80.4\%) \\
\bottomrule
\end{tabular}
\caption{
Human validation of GPT-5 operation-span annotations. Each span was
independently labeled by three of seven human annotators.
}
\label{tab:human-validation}
\end{table}

We sampled up to 12 spans per operation label, for a total of 84 validation examples. 
The context provided with each target span was limited to 400 tokens. 
Seven human annotators participated, including three authors, and each sample was independently assigned to three annotators.

Annotators received definitions and examples for the eight canonical operation labels and could additionally select \textsc{Not Determined}. 
They did not need to reproduce the original span boundaries; their task was to select the functional operation expressed by the target span. 
We defined the human reference label by majority vote whenever at least two annotators selected the same canonical label.

Table~\ref{tab:human-validation} summarizes the agreement results.
The three samples without a human-majority label were conservatively counted as mismatches when computing exact human--GPT-5 agreement.
We additionally report a soft-agreement measure that assigns half credit when GPT-5 matches a minority-vote human label in a non-unanimous example. 
We treat this soft score only as a supplementary description of annotation ambiguity.

The annotation interface and core task setup are summarized in Table~\ref{tab:human-annotation-interface}. 
Annotators received the instructions in Korean. Here, we provide an English translation of the complete labeling guidelines provided to them.

\section{Experimental Setup and Implementation Details}
\label{app:setup_details}
\subsection{Detailed Setup and Results for Representational Separability}
\label{app:separability_details}

\subsubsection{Train/Test Splits and Class Balancing}
\label{app:data-statistics}

We split retained reasoning-trace instances into training and test sets after shuffling. Since multiple generation attempts may correspond to the same question, the split is not strictly question-disjoint; for Qwen3-8B, 14 of 608 questions occur in both splits.

After annotation, we sample up to at most 300 training spans and at most 60 test spans per operation without replacement. 
We fit all normalization, projection, and operation-direction parameters using only the training spans. 
Test spans are used only for held-out evaluation. 
When an operation contains fewer than the target number of spans, we retain all available examples and do not oversample.

Decomposition is the only operation that does not reach the target counts in all main-model settings. 
Qwen3-8B contains 233 training and 52 test Decomposition spans, while Gemma4-31B contains 233 training and 47 test spans. 
All other operations reach the target sample budgets.

\subsubsection{Setup Description}
This section provides additional implementation details for the representational separability analysis in Section~\ref{sec:separability}.

\paragraph{Representation extraction.}
For each model and dataset, we generate reasoning traces and retain only responses that lead to correct final answers. 
This allows us to analyze operation-level representation geometry under successful reasoning trajectories.
For every retained response, we save the generated token sequence and extract hidden representations from each layer.
For token position $t$, layer $l$, and representation type $m$, we denote the hidden representation as
\[
h_t^{(l,m)} \in \mathbb{R}^{d},
\]
where
\[
m \in \{\mathrm{attn}, \mathrm{mlp}, \mathrm{res}\}
\]
indicates the attention output, MLP output, and residual stream output after the layer, respectively.
We additionally include the embedding layer as $l=0$.

\paragraph{Reasoning-span annotation.}
Each generated reasoning trace is segmented into non-overlapping reasoning operation chunks.
A span is represented as
\[
s_i = \left(t_i^{\mathrm{start}}, t_i^{\mathrm{end}}, y_i\right),
\]
where $t_i^{\mathrm{start}}$ and $t_i^{\mathrm{end}}$ denote the inclusive token boundaries of the span, and
\[
y_i \in \mathcal{Y}
\]
is the assigned reasoning-operation label.
In the main analysis, $\mathcal{Y}$ contains the eight recurring operation labels:
Extraction, Direct mapping, Decomposition, Recall, Deduction, Algebraic manipulation, Arithmetic computation, and Final answer.
Since these annotations are obtained from generated text, they should be interpreted as labels of textually expressed reasoning functions rather than direct labels of latent cognitive states.
The number of spans generated are in Table~\ref{tab:train-test-reasoning-occurrence}.

\paragraph{Span-level representation.}
Because a reasoning operation chunk may contain multiple tokens, we construct a span-level representation from the token representations inside the span.
We consider four pooling strategies: first-token, middle-token, last-token, and mean pooling.
For a span $s_i$, layer $l$, and representation type $m$, the span representation is
\[
x_i^{(l,m)}
=
\operatorname{Pool}
\left(
\left\{
h_t^{(l,m)}
:
t \in [t_i^{\mathrm{start}}, t_i^{\mathrm{end}}]
\right\}
\right),
\]
where $\operatorname{Pool}(\cdot)$ is one of the four pooling strategies above.
The main results use the middle-token representation, while the other pooling variants are reported separately.

\paragraph{Entropy-centered span window.}
For the span-level analysis, we use a fixed-window representation to avoid very long spans dominating the pooled representation.
For each annotated span shorter than 300 tokens, we first identify the token position with the highest generation entropy:
\[
u_i
=
\arg\max_{t \in [t_i^{\mathrm{start}}, t_i^{\mathrm{end}}]}
H_t,
\]
where $H_t$ is the token-level generation entropy at position $t$.
We then select a 50-token window centered at $u_i$.
If the window exceeds the span boundary, we shift it so that the full window remains inside the annotated span.
Spans longer than 300 tokens are excluded from this analysis.
This windowing strategy is intended to capture the region where the model is relatively uncertain or computationally active within the reasoning operation chunk.

Let the resulting token window be denoted by
\[
W_i \subseteq [t_i^{\mathrm{start}}, t_i^{\mathrm{end}}].
\]
The pooled span representation is then computed as
\[
x_i^{(l,m)}
=
\operatorname{Pool}
\left(
\left\{
h_t^{(l,m)}
:
t \in W_i
\right\}
\right).
\]

\paragraph{Projection into a discriminative subspace.}
For each layer $l$ and representation type $m$, we split annotated spans into train and test sets.
All normalization and projection parameters are fitted only on the train split and then applied to the held-out test split.

We first apply $L_2$ normalization:
\[
\tilde{x}_i^{(l,m)}
=
\frac{x_i^{(l,m)}}{\left\|x_i^{(l,m)}\right\|_2}.
\]
We then apply PCA to reduce the representation dimension to $K=128$, followed by supervised Linear Discriminant Analysis (LDA).
Since the number of operation classes is $C=8$, the LDA subspace has at most $C-1=7$ dimensions.
The projected representation is
\[
z_i^{(l,m)}
=
W_{\mathrm{LDA}}^\top
W_{\mathrm{PCA}}^\top
\tilde{x}_i^{(l,m)}.
\]
Both $W_{\mathrm{PCA}}$ and $W_{\mathrm{LDA}}$ are fitted using only the train split.

\paragraph{Operation prototypes.}
For each reasoning operation $c \in \mathcal{Y}$, we compute the centroid of projected train representations belonging to class $c$:
\[
\mu_c^{(l,m)}
=
\frac{1}{|\mathcal{D}_c|}
\sum_{i \in \mathcal{D}_c}
z_i^{(l,m)},
\]
where
\[
\mathcal{D}_c
=
\{i : y_i = c\}
\]
denotes the set of train spans labeled as operation $c$.

We also compute the centroid of all other train spans:
\[
\mu_{\neg c}^{(l,m)}
=
\frac{1}{|\mathcal{D}_{\neg c}|}
\sum_{i \in \mathcal{D}_{\neg c}}
z_i^{(l,m)},
\]
where
\[
\mathcal{D}_{\neg c}
=
\{i : y_i \neq c\}.
\]

\paragraph{One-vs-rest operation vector.}
We define the one-vs-rest operation vector for class $c$ as the normalized difference between the class centroid and the rest centroid:
\[
d_c^{(l,m)}
=
\frac{
\mu_c^{(l,m)} - \mu_{\neg c}^{(l,m)}
}{
\left\|
\mu_c^{(l,m)} - \mu_{\neg c}^{(l,m)}
\right\|_2
}.
\]
This direction captures the axis in the LDA-projected space that separates operation $c$ from the remaining reasoning operations.

\paragraph{Alignment score.}
For each held-out test span $i$ and reasoning operation $c$, we compute the dot-product alignment between the projected span representation and the operation vector:
\[
a(i,c)
=
\left(z_i^{(l,m)}\right)^\top d_c^{(l,m)}.
\]
A larger value of $a(i,c)$ indicates that the span representation is more strongly aligned with the direction associated with operation $c$.

\paragraph{One-vs-rest evaluation.}
For each operation $c$, we formulate a one-vs-rest binary classification problem.
The binary target is
\[
g(i,c)
=
1[y_i = c],
\]
and the prediction score is the alignment score $a(i,c)$.
We compute AUROC and AUPRC by varying the decision threshold over $a(i,c)$.
A high AUROC indicates that spans labeled as operation $c$ tend to have higher alignment with the operation vector $d_c^{(l,m)}$ than spans labeled as other operations.

\paragraph{Random baseline.}
As a control, we repeat the same evaluation pipeline with randomly assigned operation labels and randomly selected token positions.
The random baseline uses the same number of spans and the same train-test protocol as the main experiment.
This baseline tests whether the observed separability arises from the reasoning-operation annotation rather than from the projection pipeline or dataset imbalance alone.

\subsection{Reproducibility Details for Trace Generation}
\label{app:reproducibility_details}

\paragraph{Model checkpoints.}
We use the Hugging Face checkpoints
\texttt{Qwen/Qwen2.5-7B},
\texttt{Qwen/Qwen3-8B}, and
\texttt{google/gemma-4-31B-it}.
\texttt{Qwen/Qwen2.5-7B} is the base Qwen2.5 model rather than an
instruction-tuned or math-specific variant.

\paragraph{Prompting and generation.}
For models with a chat template, each problem is provided as a single
user message of the form
\begin{quote}
\texttt{Solve the following problem step by step.\textbackslash n\textbackslash n\{question\}}
\end{quote}
with no system prompt or few-shot examples.
We construct the input using
\texttt{tokenizer.apply\_chat\_template(..., tokenize=False, add\_generation\_prompt=True)}.
When a chat template is unavailable, we use the fallback prompt
\begin{quote}
\small\ttfamily
Solve step by step.\\
\\
Problem:\\
\{question\}\\
\\
Reasoning:
\end{quote}

We do not explicitly set any thinking-mode flags, such as \texttt{enable\_thinking} or a thinking budget.
Consequently, thinking behavior follows the default configuration of the tokenizer checkpoint used for generation.

We sample responses with
\texttt{do\_sample=True},
$\texttt{temperature}=0.7$,
$\texttt{top\_p}=0.9$, and
$\texttt{max\_new\_tokens}=4096$.
Other decoding parameters, including
\texttt{top\_k},
\texttt{num\_beams}, and
\texttt{repetition\_penalty},
are left at their library defaults.
We use no custom stopping criterion or stop string; generation terminates upon the model's default end-of-sequence token or when the maximum generation length is reached.

\paragraph{Correctness filtering.}
We retain reasoning traces based on a zero-shot LLM correctness judgment rather than exact-match or rule-based answer parsing.
The judge is provided with the question, reference answer, and generated response, and is prompted with
\begin{quote}
\texttt{Is the following RESPONSE correct given the REFERENCE ANSWER?}
\end{quote}
followed by the instruction
\begin{quote}
\texttt{Answer only YES or NO.}
\end{quote}

A response is treated as correct if the judge output contains the case-insensitive substring \texttt{yes}; all other outputs are treated as incorrect.
Thus, semantic equivalence between generated and reference answers is
delegated to the judge rather than determined through numerical,
\LaTeX{}, or boxed-answer normalization.
The correctness judge is \texttt{Qwen3.6-27B}.

\paragraph{Randomness.}
Trace generation is stochastic and does not use a fixed random seed.
For each problem, generation may therefore produce different traces
across independent runs; when repeated generation attempts are required,
the random state is not reset between attempts.
Accordingly, the generated trace corpus is not expected to be bitwise
reproducible.

In contrast, subsequent question-level splitting, class-wise span
sampling, and random-baseline construction use a fixed random seed of
$42$.

\paragraph{Computation environment.}
For each model and dataset, we generate reasoning traces and collect hidden representations corresponding to the labeled reasoning steps.
Inference is performed using eight NVIDIA V100 GPUs, and subsequent representation analyses are conducted on a single V100 node.

\clearpage

% Add to preamble if not already present:
% \newcolumntype{L}[1]{>{\raggedright\arraybackslash}p{#1}}
% \newcolumntype{C}[1]{>{\centering\arraybackslash}p{#1}}

% =========================================================
% Table 1: Stage 1 & 2
\begin{table*}[t]
\centering
\footnotesize
\setlength{\tabcolsep}{2pt}
\renewcommand{\arraystretch}{1.08}

\caption{Reasoning operation schema (Stages 1--2).}
\label{tab:reasoning_type_schema_1}

\begin{tabular}{
>{\raggedright\arraybackslash}p{0.11\textwidth}
>{\centering\arraybackslash}p{0.13\textwidth}
>{\centering\arraybackslash}p{0.10\textwidth}
>{\centering\arraybackslash}p{0.10\textwidth}
>{\raggedright\arraybackslash}p{0.13\textwidth}
>{\raggedright\arraybackslash}p{0.30\textwidth}
}
\toprule
\textbf{Problem-solving stage}
& \textbf{Level 1}
& \textbf{Level 2}
& \textbf{Level 3}
& \textbf{Reasoning operation}
& \textbf{Description / Example} \\
\midrule

\multirow{5}{*}{\makecell[l]{\textbf{Stage 1}\\Understanding\\the Problem}}
& Extraction
& --
& --
& Extraction
& Directly retrieving given information from the problem statement without modification.
\newline \textit{Example:} ``Two numbers add up to 10, and both are non-negative.'' \\
\cline{2-6}

& \multirow{3}{*}{\centering Symbolization}
& Direct-mapping
& --
& Direct-mapping
& Directly converting a statement into a formal representation without changing its structure.
\newline \textit{Example:} The sum of two numbers is 10 $\rightarrow x+y=10$. \\
\cline{3-6}

&
& Reframing
& --
& Reframing
& Reconstructing the problem in a different representational system.
\newline \textit{Example:} Geometry problem $\rightarrow$ coordinate geometry. \\
\cline{3-6}

&
& Abstraction
& --
& Abstraction
& Removing unnecessary details and retaining only the essential structure.
\newline \textit{Example:} 3 red balls and 2 blue balls $\rightarrow (3,2)$. \\
\cline{2-6}

& State-space-definition
& --
& --
& State-space-definition
& Defining possible states, constraints, and the solution space.
\newline \textit{Example:} $x \geq 0$, $y \geq 0$, $x+y=10$. \\

\midrule

\multirow{6}{*}{\makecell[l]{\textbf{Stage 2}\\Planning\\the Solution}}
& Pattern-matching
& --
& --
& Pattern-matching
& Mapping the current problem to a known problem type, template, or strategy.
\newline \textit{Example:} Flip a coin three times $\rightarrow$ probability/combinatorics problem. \\
\cline{2-6}

& \multirow{3}{*}{\centering Structural-analysis}
& Symmetry
& --
& Symmetry
& Identifying whether the problem remains unchanged under transformations.
\newline \textit{Example:} Normal distribution is symmetric about the mean $\rightarrow$ compute one side and double it. \\
\cline{3-6}

&
& Invariance
& --
& Invariance
& Identifying quantities that remain unchanged throughout a process.
\newline \textit{Example:} Energy conservation $\rightarrow E_{\mathrm{initial}}=E_{\mathrm{final}}$. \\
\cline{3-6}

&
& Decomposition
& --
& Decomposition
& Breaking a complex problem into simpler or independent subproblems.
\newline \textit{Example:} Heptagon $\rightarrow$ divide into triangles $\rightarrow$ solve each triangle. \\
\cline{2-6}

& Idealization
& --
& --
& Idealization
& Simplifying the problem by removing irrelevant or noisy details.
\newline \textit{Example:} Motion with friction $\rightarrow$ ignore friction. \\
\cline{2-6}

& Hypothesis-formation
& --
& --
& Hypothesis-formation
& Generating a conjecture or assumption based on observed patterns or partial reasoning.
\newline \textit{Example:} Terms cancel out $\rightarrow$ the result may be 0. \\

\bottomrule
\end{tabular}
\end{table*}

% =========================================================
% Table 2: Stage 3
\begin{table*}[t]
\centering
\footnotesize
\setlength{\tabcolsep}{2pt}
\renewcommand{\arraystretch}{1.08}

\caption{Reasoning operation schema (Stage 3).}
\label{tab:reasoning_type_schema_2}

\begin{tabular}{
>{\raggedright\arraybackslash}p{0.11\textwidth}
>{\centering\arraybackslash}p{0.13\textwidth}
>{\centering\arraybackslash}p{0.10\textwidth}
>{\centering\arraybackslash}p{0.10\textwidth}
>{\raggedright\arraybackslash}p{0.13\textwidth}
>{\raggedright\arraybackslash}p{0.30\textwidth}
}
\toprule
\textbf{Problem-solving stage}
& \textbf{Level 1}
& \textbf{Level 2}
& \textbf{Level 3}
& \textbf{Reasoning operation}
& \textbf{Description / Example} \\
\midrule

\multirow{12}{*}{\makecell[l]{\textbf{Stage 3}\\Carrying Out\\the Plan}}
& \multirow{2}{*}{\centering Retrieval}
& Recall
& --
& Recall
& Retrieving relevant definitions, rules, concepts, or formulas.
\newline \textit{Example:} Need area of a circle $\rightarrow A=\pi r^2$. \\
\cline{3-6}

&
& Matching
& --
& Matching
& Determining whether retrieved knowledge applies to the current situation.
\newline \textit{Example:} Apply $A=\pi r^2$ when the radius is given. \\
\cline{2-6}

& \multirow{4}{*}{\centering Inference-flow}
& \multirow{3}{*}{\centering Chaining}
& Deduction
& Deduction
& Deriving a specific conclusion from a general rule.
\newline \textit{Example:} General theorem $\rightarrow$ apply to the given case. \\
\cline{4-6}

&
&
& Induction
& Induction
& Inferring a general rule from observed patterns.
\newline \textit{Example:} Several terms follow the same difference $\rightarrow$ infer arithmetic pattern. \\
\cline{4-6}

&
&
& Analogy
& Analogy
& Reasoning from a structurally similar case.
\newline \textit{Example:} Solve a new problem using a similar solved template. \\
\cline{3-6}

&
& Branching
& --
& Branching
& Splitting into cases and analyzing each separately.
\newline \textit{Example:} $x^2=4 \rightarrow x=2$ or $x=-2$. \\
\cline{2-6}

& \multirow{4}{*}{\centering Execution}
& Instantiation
& --
& Instantiation
& Assigning specific values or conditions to abstract variables or expressions.
\newline \textit{Example:} $x+y=10$, let $x=3$, then $y=7$. \\
\cline{3-6}

&
& \multirow{3}{*}{\centering Operation}
& Algebraic-manipulation
& Algebraic-manipulation
& Transforming expressions or equations.
\newline \textit{Example:} $x+y=10 \rightarrow y=10-x$. \\
\cline{4-6}

&
&
& Arithmetic-computation
& Arithmetic-computation
& Performing numerical calculations.
\newline \textit{Example:} $10-3=7$. \\
\cline{4-6}

&
&
& Logical-evaluation
& Logical-evaluation
& Evaluating truth values or logical implications.
\newline \textit{Example:} If $P \rightarrow Q$ and $P$ is true, then $Q$ is true. \\
\cline{2-6}

& Representation-construction
& --
& --
& Representation-construction
& Organizing computed results into structured forms such as lists, sequences, or tables.
\newline \textit{Example:} Possible values: $(1,9),(2,8),(3,7),\ldots$ \\
\cline{2-6}

& Pattern-extraction
& --
& --
& Pattern-extraction
& Identifying patterns, regularities, or trends from computed results.
\newline \textit{Example:} $2,4,6,8 \rightarrow$ increase by 2. \\

\bottomrule
\end{tabular}
\end{table*}

% =========================================================
% Table 3: Stage 4
\begin{table*}[t]
\centering
\footnotesize
\setlength{\tabcolsep}{2pt}
\renewcommand{\arraystretch}{1.08}

\caption{Reasoning operation schema (Stage 4).}
\label{tab:reasoning_type_schema_3}

\begin{tabular}{
>{\raggedright\arraybackslash}p{0.11\textwidth}
>{\centering\arraybackslash}p{0.13\textwidth}
>{\centering\arraybackslash}p{0.10\textwidth}
>{\centering\arraybackslash}p{0.10\textwidth}
>{\raggedright\arraybackslash}p{0.13\textwidth}
>{\raggedright\arraybackslash}p{0.30\textwidth}
}
\toprule
\textbf{Problem-solving stage}
& \textbf{Level 1}
& \textbf{Level 2}
& \textbf{Level 3}
& \textbf{Reasoning operation}
& \textbf{Description / Example} \\
\midrule

\multirow{5}{*}{\makecell[l]{\textbf{Stage 4}\\Looking Back\\and Final Answer}}
& Strategy-validation
& --
& --
& Strategy-validation
& Verifying whether the chosen strategy or assumptions were valid.
\newline \textit{Example:} Check whether differences are constant before assuming an arithmetic sequence. \\
\cline{2-6}

& Error-detection
& --
& --
& Error-detection
& Identifying mistakes in reasoning, calculation, or logical steps.
\newline \textit{Example:} $10-3=6$ $\rightarrow$ incorrect arithmetic; should be 7. \\
\cline{2-6}

& Dimensional-analysis
& --
& --
& Dimensional-analysis
& Checking whether units or dimensions are consistent with the required quantity.
\newline \textit{Example:} Speed = distance/time $\rightarrow$ meters/second. \\
\cline{2-6}

& Extreme-case-testing
& --
& --
& Extreme-case-testing
& Testing whether the solution holds under extreme or boundary conditions.
\newline \textit{Example:} For $f(x)=x/(x+1)$, test $x=0$ and $x\rightarrow\infty$. \\
\cline{2-6}

& Final-answer
& --
& --
& Final-answer
& Clearly presenting the final result in the required format.
\newline \textit{Example:} Therefore, $x=3$ and $y=7$. \\

\bottomrule
\end{tabular}
\end{table*}
\clearpage
\onecolumn
\begin{table}[t]
\centering
\footnotesize
\setlength{\tabcolsep}{3.5pt}
\renewcommand{\arraystretch}{1.08}
\begin{tabular}{p{0.22\columnwidth} p{0.71\columnwidth}}
\hline
\textbf{Item} & \textbf{Instructions given to annotators} \\
\hline

\textbf{Task objective} &
Determine the role played by the highlighted text in the reasoning
process. Annotators should identify the \emph{function} of the
highlighted span within the overall solution, rather than solve the
problem themselves or verify whether the final answer is correct. \\

\hline

\textbf{Question} &
The original problem to which the model-generated reasoning responds. \\

\textbf{Response with highlighted span} &
The model-generated reasoning trace. Only the highlighted portion
(\emph{full span}) is the annotation target. The surrounding response
is provided solely to understand why the highlighted span appears at
that point in the reasoning process. \\

\textbf{\texttt{shortened\_span\_text}} &
A shorter phrase extracted from the full span when the full span is
long. When the full span is already short, the shortened span is
identical to the full span. \\

\textbf{Answer columns} &
For each example, annotators provide three judgments:
(1) the reasoning-operation label that best characterizes the full
span; (2) their confidence in that label; and
(3) whether the \texttt{shortened\_span\_text} preserves the main
reasoning operation expressed by the full span. \\

\textbf{Disabled cells} &
Black cells are intentionally disabled and should be left untouched;
annotators should not type or make any selection in these cells. \\

\hline

\textbf{Label selection} &
Select the single reasoning operation that best describes the
\emph{primary function} of the highlighted full span. The label should
reflect what the span is doing in the reasoning process, rather than
its surface wording alone. \\

\textbf{Multiple operations} &
If several operations appear in a span but one clearly dominates,
select the dominant operation. Select \emph{Multiple operations, with
no clear primary} only when two or more operations are similarly
important and no single operation clearly dominates. \\

\textbf{No applicable label} &
Select \emph{None of the above} when the highlighted span does not
express any of the defined reasoning operations or when its function
cannot be determined. \\

\textbf{Correctness} &
Do not use mathematical correctness as a criterion for assigning the
operation label. An incorrect calculation can still be
\emph{Arithmetic computation}, and an incorrectly applied theorem can
still instantiate \emph{Recall} or \emph{Deduction}. \\

\textbf{Difficult mathematics} &
When the mathematical content is difficult, focus on the
\emph{action} being performed: for example, whether the span extracts
given information, translates it into symbols, recalls a formula,
derives a conclusion, rearranges an expression, performs a numerical
calculation, decomposes the problem into subproblems, or states the
final answer. \\
\hline
\end{tabular}

\caption{
Summary of the human annotation procedure and interface.
The instructions were originally provided to annotators in Korean.
}
\label{tab:human-annotation-interface}
\end{table}

\twocolumn
\clearpage
\onecolumn
\begin{promptbox}{System Prompt}
\begin{promptlisting}
You are given a numbered solution to a math/science problem.

Each word has been prefixed with §N (a zero-based global index):
  §0 word0 §1 word1 §2 word2 ...

Your task: identify reasoning spans by their §N word-index range and assign each a tag.

----------------------------------

Full description of tags: 

# Stage 1: Understanding the Problem
## What this stage does
The Understanding stage is the process of transforming a problem from its original, often informal description into a structured and analyzable representation.
At this stage, the goal is not to solve the problem, but to:
- identify the relevant information,
- convert it into formal representations, and
- define the space in which the solution exists.

## Components
1. EXTRACTION
  [Definition] The process of directly retrieving the given information from the problem statement without modification.
  [Role]
  - Ensures that all relevant data is explicitly identified
  - Prevents omission or misinterpretation of key conditions
  [Example]
  - Two numbers add up to 10, and both are non-negative.
2. SYMBOLIZATION
  [Definition] The process of converting informal information (natural language, descriptions) into formal representations such as variables, equations, or logical structures.
  [Example]
  - The sum of two numbers is 10 -> x + y = 10
  [Subcomponents] 
  2-1. SYMBOLIZATION.DIRECT-MAPPING
	  [Definition] Directly converting the given statement into a formal representation without changing its structure or interpretation.
	  [Example]
	  - The sum of two numbers is 10 → x + y = 10
	  - x is greater than 3 → x > 3
	  - A number is even → x = 2k
  2-2. SYMBOLIZATION.REFRAMING
    [Definition] Reconstructing the problem in a different representational system (e.g., algebraic, geometric, probabilistic).
    [Example]
    - Geometry problem → coordinate geometry
    - Word problem → algebraic equation
  2-3. SYMBOLIZATION.ABSTRACTION
    [Definition] Removing unnecessary details and retaining only the essential structure of the problem.
    [Example]
    - 3 red balls and 2 blue balls -> (3, 2) (Color becomes irrelevant; only quantities matter)
3. STATE-SPACE-DEFINITION
  [Definition] The process of defining the set of all possible states (solutions or configurations), and the constraints that restrict them
  [Core elements]
  - State variables (e.g., x, y)
  - Constraints (e.g., x $\leq$ 0, y $\geq$ 0)
  - Solution space (set of all valid states)
  [Role]
  - Determines where valid solutions can exist
  - Restricts the problem to a well-defined domain
  - Enables systematic reasoning and search
  [Example]
  - x x $\leq$ 0, y $\geq$ 0
  - x + y = 10
\end{promptlisting}
\end{promptbox}

\begin{promptbox}
\begin{promptlisting}
# Stage 2: Planning the Solution
## What this stage does
The Planning stage is the process of determining how to solve the problem after it has been properly understood.
At this stage, the goal is to:
- identify relevant problem-solving strategies,
- analyze structural relationships within the problem, and
- propose a direction for execution.
Unlike the Understanding stage (which builds representation), Planning focuses on strategy selection and structural insight.
## Components
1. PATTERN-MATCHING
  [Definition] The process of mapping the current problem to a known problem type, template, or strategy.
  [Role]
  - Activates prior knowledge
  - Reduces problem complexity
  - Suggests a solution approach
  [Example]
  - Flip a coin three times → **probability + combinatorics problem**
  - Find the shortest path between points → **graph / optimization problem**
2. STRUCTURAL-ANALYSIS
  [Definition] The process of analyzing the internal structure, relationships, and invariant properties of the problem.
  [Core Idea] Focus on how elements within the problem are related and how those relationships constrain the solution.
  [Subcomponents]
  2-1. STRUCTURAL-ANALYSIS.SYMMETRY
	[Definition] Identifying whether the problem remains unchanged under certain transformations (e.g., reflection, rotation, permutation)
	[Example]
	Normal distribution
	→ symmetric about the mean
	→ **compute one side and double it**
	2-2. STRUCTURAL-ANALYSIS.INVARIANCE
	[Definition] Identifying quantities that remain unchanged throughout a process.
	[Example]
	Energy conservation
	→ total energy remains constant
	→ **E_initial = E_final**
	2-3. STRUCTURAL-ANALYSIS.DECOMPOSITION
	[Definition] Breaking a complex problem into simpler or independent subproblems.
	[Example]
	Heptagon → **divide into triangles → solve each triangle** → combine results
	2-4. IDEALIZATION
	[Definition] Simplifying the problem by removing irrelevant or noisy details.
	[Example]
	Motion with friction → **ignore friction** → idealized motion
3. HYPOTHESIS-FORMATION
	[Definition] The process of generating a conjecture or assumption based on observed patterns or partial reasoning.
	[Role]
	- Guides further reasoning
	- Narrows down possibilities
	- Suggests direction for execution
	[Example]
	Observed pattern: Terms cancel out → Hypothesis: The result may be 0
    \end{promptlisting}
\end{promptbox}

\begin{promptbox}
\begin{promptlisting}
# Stage 3: Carrying Out the Plan
## What this stage does
The Carry Out stage is the process of executing the planned strategy to produce concrete results.
At this stage, the goal is to:
- apply relevant knowledge and rules,
- perform logical and numerical operations,
- explore possible cases, and
- construct intermediate and final results.
Unlike the Planning stage (which determines what to do), this stage focuses on actually performing the reasoning and computations.
## Components
1. RETRIEVAL
  [Definition] The process of recalling relevant concepts, definitions, rules, or formulas from prior knowledge.
  [Role]
  - Provides the necessary tools for execution
  - Connects abstract knowledge to the current problem
  [Subcomponents]
  2-1. RETRIEVAL.RECALL: retrieving definitions or formulas
  2-2. RETRIEVAL.MATCHING: determining whether the knowledge applies to the current situation
  [Structure]
	- [Problem context] → [Relevant concept retrieved] → [Apply if applicable]
	[Example]
	- Area of a circle is needed -> **A = $\pi$r²** -> apply to problem
2. INFERENCE-FLOW
  [Definition] The process of constructing a sequence of logical steps (chaining) or dividing the problem into multiple cases (branching).
  [Role]
  - Enables step-by-step reasoning
  - Handles conditional or multiple-case scenarios
  [Subcomponents]
  2-1. CHAINING
    2-1-1. CHAINING.DEDUCTION (general → specific)
    2-1-2. CHAINING.INDUCTION (pattern → general rule)
    2-1-3. CHAINING.ANALOGY (similar structure)
  2-2. BRANCHING: Splitting into cases and analyzing each separately
  [Structure]
  - Chaining: [A] → [B] → [C]
  - Branching: [Condition] → Case 1 / → Case 2 / → Case 3
  [Example]
  - x² = 4 → Branching: **x = 2** or **x = -2**
3. EXECUTION
  [Subcomponents]
  3-1. EXECUTION.INSTANTIATION
  [Definition] Assigning specific values or conditions to abstract variables or expressions.
  [Role]
  - Moves from abstract representation to concrete evaluation
  [Example]
  - x + y = 10 → Let x = 3 → Then y = 7
  3-2. EXECUTION.OPERATION: Performing actual computations or transformations on expressions.
    3-2-1. EXECUTION.OPERATION.ALGEBRAIC-MANIPULATION: Transforming the structure of expressions or equations.
   [Example]
   - x + y = 10 → y = 10 - x
    3-2-2. EXECUTION.OPERATION.ARITHMETIC-COMPUTATION: Performing numerical calculations.
   [Example]
   - 10 - 3 = 7
    3-2-3. EXECUTION.OPERATION.LOGICAL-EVALUATION: Evaluating truth values or logical implications.
    [Example]
    - If P → Q, and P is true → Q is true
4. REPRESENTAION-CONSTRUCTION
  [Definition] Organizing computed results into structured forms such as lists, sequences, or tables.
  [Role]
	- Makes results interpretable
	- Prepares for further reasoning or comparison
	[Example]
	- Possible values: (1,9), (2,8), (3,7), ...
\end{promptlisting}
\end{promptbox}

\begin{promptbox}
\begin{promptlisting}
5. PATTERN-EXTRACTION
  [Definition] Identifying patterns, regularities, or trends from computed results.
  [Role]
  - Enables generalization
  - Supports further reasoning or simplification
  [Example]
  - Sequence: 2, 4, 6, 8 → Pattern: **Increase by 2**

# Stage 4: Looking Back and Final Answer
## What this stage does
The Look Back stage is the process of evaluating and validating the solution after it has been obtained.
At this stage, the goal is to:
- verify correctness,
- detect possible errors,
- test robustness under different conditions, and
- improve or refine the solution if necessary.
Unlike the Carry Out stage (which produces results), this stage focuses on checking, validating, and reflecting on those results.
## Components
1. STRATEGY-VALIDATION
	[Definition] The process of verifying whether the chosen strategy or assumptions were valid.
	[Role]
	- Confirms that the reasoning approach is appropriate
	- Ensures that conclusions follow logically from assumptions
	[Example]
	- Assumption: The sequence is arithmetic → Check differences: 2, 4, 6, 8 → Validation: **Differences are constant → correct assumption**
2. ERROR-DETECTION
	[Definition] Identifying mistakes in reasoning, calculation, or logical steps.
	[Role]
	- Prevents incorrect conclusions
	- Ensures internal consistency
	[Example]
	- Computed: 10 - 3 = 6 → **Error detected: Incorrect arithmetic → should be 7**
3. DIMENSIONAL-ANALYSIS
  [Definition] Checking whether the units or dimensions of the result are consistent with the required quantity.
  [Role]
  - Ensures physical or logical correctness
  - Detects hidden inconsistencies
  [Example]
	- Speed = distance / time → **Unit check: meters / seconds → valid**
4. EXTREME-CASE-TESTING
	[Definition] Testing whether the solution holds under extreme or boundary conditions. 
	[Role]	
	- Reveals hidden flaws
	- Tests robustness of the solution
	[Example]
	- f(x) = x / (x + 1) → Test x = 0 → f(0) = 0 → valid → Test x → $\infty$ → f(x) → 1 → consistent
5. FINAL-ANSWER
	[Definition] The process of clearly presenting the final result derived from the problem-solving process in the required format.
	[Role]
	- Communicates the solution
	- Ensures clarity and completeness
	- Matches the required output format
	[Example]
	- Therefore, the values are **x = 3 and y = 7**.
\end{promptlisting}
\end{promptbox}

\begin{promptbox}
\begin{promptlisting}
----------------------------------

Tagging Schema (use EXACT labels below):

EXTRACTION
SYMBOLIZATION.DIRECT-MAPPING
SYMBOLIZATION.REFRAMING
SYMBOLIZATION.ABSTRACTION
STATE-SPACE-DEFINITION
PATTERN-MATCHING
STRUCTURAL-ANALYSIS.SYMMETRY
STRUCTURAL-ANALYSIS.INVARIANCE
STRUCTURAL-ANALYSIS.DECOMPOSITION
IDEALIZATION
HYPOTHESIS-FORMATION
RETRIEVAL.RECALL
RETRIEVAL.MATCHING
INFERENCE-FLOW.CHAINING.DEDUCTION
INFERENCE-FLOW.CHAINING.INDUCTION
INFERENCE-FLOW.CHAINING.ANALOGY
INFERENCE-FLOW.BRANCHING
EXECUTION.INSTANTIATION
EXECUTION.OPERATION.ALGEBRAIC-MANIPULATION
EXECUTION.OPERATION.ARITHMETIC-COMPUTATION
EXECUTION.OPERATION.LOGICAL-EVALUATION
REPRESENTATION-CONSTRUCTION
PATTERN-EXTRACTION
STRATEGY-VALIDATION
ERROR-DETECTION
DIMENSIONAL-ANALYSIS
EXTREME-CASE-TESTING
FINAL-ANSWER
----------------------------------

Output ONLY a valid JSON array. No explanation, no markdown, no other text.

Format:
[
  {"start": 3, "end": 7, "tag": "TAG_NAME"},
  {"start": 12, "end": 15, "tag": "TAG_NAME"}
]

Where start and end are inclusive word indices (the §N numbers).
A span can cross multiple lines if the reasoning continues.

When end - start > 20 (span covers more than 20 words), you MUST add a "reason" field
explaining in one sentence why it is one continuous reasoning act:
  {"start": 45, "end": 120, "tag": "TAG_NAME", "reason": "..."}

Rules:
- start and end must be §N indices that actually appear in the text
- Do not overlap spans
- Do not tag filler words or transitions unless they are part of a reasoning span
- A single span may cover multiple lines if it is one continuous reasoning act
- If end - start > 20, include a "reason" field (one sentence, concrete)
- If unsure, leave untagged rather than guess
- Tag ONLY the key reasoning content, not filler words.
- Do NOT wrap entire sentences unless absolutely necessary.
- Focus on:
  - problem type (pattern)
  - structural insights
  - key equations or transformations
  - branching conditions
  - final conclusions
- Output only the JSON array, nothing else
\end{promptlisting}
\end{promptbox}

\begin{promptbox}
\begin{promptlisting}
----------------------------------

Example 1:

[Question]
What is 2 + 3?

[Solution]
§0 We §1 add §2 2 §3 and §4 3 §5 to §6 get §7 5.

[Output]
[
  {"start": 2, "end": 7, "tag": "EXECUTION.OPERATION.ARITHMETIC-COMPUTATION"},
  {"start": 7, "end": 7, "tag": "FINAL-ANSWER"}
]

----------------------------------

Example 2:

[Question]
A store has 3 types of products. Type A costs $5 each, Type B costs $8 each, Type C costs $12 each. If a customer buys 4 of type A, 3 of type B, and 2 of type C, what is the total cost?

[Solution]
§0 To §1 find §2 the §3 total §4 cost, §5 I §6 need §7 to §8 calculate §9 each §10 type §11 separately §12 and §13 add §14 them §15 together.
§16 Type §17 A: §18 4 §19 items §20 × §21 $5 §22 = §23 $20.
§24 Type §25 B: §26 3 §27 items §28 × §29 $8 §30 = §31 $24.
§32 Type §33 C: §34 2 §35 items §36 × §37 $12 §38 = §39 $24.
§40 Now §41 I §42 combine §43 these §44 results: §45 $20 §46 + §47 $24 §48 + §49 $24 §50 = §51 $68.
§52 Therefore, §53 the §54 total §55 cost §56 is §57 $68.

[Output]
[
  {"start": 0, "end": 15, "tag": "STRUCTURAL-ANALYSIS.DECOMPOSITION"},
  {"start": 18, "end": 23, "tag": "EXECUTION.OPERATION.ARITHMETIC-COMPUTATION"},
  {"start": 26, "end": 31, "tag": "EXECUTION.OPERATION.ARITHMETIC-COMPUTATION"},
  {"start": 34, "end": 39, "tag": "EXECUTION.OPERATION.ARITHMETIC-COMPUTATION"},
  {"start": 45, "end": 51, "tag": "EXECUTION.OPERATION.ARITHMETIC-COMPUTATION"},
  {"start": 52, "end": 57, "tag": "FINAL-ANSWER"}
]

----------------------------------

Example 3 (long span requiring "reason"):

[Question]
For each integer n from 1 to 9, check whether n is divisible by 3 and sum the digit counts of those that are.

[Solution]
§0 I §1 will §2 check §3 each §4 integer §5 from §6 1 §7 to §8 9. §9 n=1: §10 digit §11 sum §12 1, §13 not §14 divisible §15 by §16 3. §17 n=2: §18 digit §19 sum §20 2, §21 not §22 divisible §23 by §24 3. §25 n=3: §26 digit §27 sum §28 3, §29 divisible §30 by §31 3, §32 count §33 it. §34 n=4: §35 digit §36 sum §37 4, §38 not §39 divisible §40 by §41 3. §42 n=5: §43 digit §44 sum §45 5, §46 not §47 divisible §48 by §49 3. §50 n=6: §51 digit §52 sum §53 6, §54 divisible §55 by §56 3, §57 count §58 it. §59 n=7: §60 digit §61 sum §62 7, §63 not §64 divisible §65 by §66 3. §67 n=8: §68 digit §69 sum §70 8, §71 not §72 divisible §73 by §74 3. §75 n=9: §76 digit §77 sum §78 9, §79 divisible §80 by §81 3, §82 count §83 it. §84 Total §85 qualifying: §86 3.

[Output]
[
  {"start": 0, "end": 83, "tag": "EXECUTION.OPERATION.LOGICAL-EVALUATION", "reason": "repeated application of the same digit-sum divisibility check across all nine consecutive integers"},
  {"start": 84, "end": 86, "tag": "FINAL-ANSWER"}
]
\end{promptlisting}
\end{promptbox}

\begin{promptbox}
\begin{promptlisting}
----------------------------------
Example 4(Other examples)
§0 Flip §1 a §2 coin §3 three §4 times §5 → §6 This §7 is §8 a §9 probability §10 + §11 combinatorics §12 problem
[{"start": 6, "end": 12, "tag": "PATTERN-MATCHING"}]

§0 Find §1 the §2 shortest §3 path §4 between §5 points §6 → §7 This §8 becomes §9 a §10 graph §11 / §12 optimization §13 problem
[{"start": 7, "end": 13, "tag": "PATTERN-MATCHING"}]

§0 The §1 sum §2 of §3 two §4 numbers §5 is §6 10 §7 → §8 x §9 + §10 y §11 = §12 10
[{"start": 8, "end": 12, "tag": "SYMBOLIZATION.DIRECT-MAPPING"}]

§0 x² §1 = §2 4 §3 → §4 x §5 = §6 2 §7 or §8 x §9 = §10 -2
[{"start": 4, "end": 10, "tag": "INFERENCE-FLOW.BRANCHING"}]
\end{promptlisting}
\end{promptbox}

\begin{promptbox}{User Prompt}
\begin{promptlisting}
[Question]
{question}

---
[Solution]
{reasoning}
\end{promptlisting}

\end{promptbox}
\twocolumn

\end{document}